\documentclass[lettersize,journal]{IEEEtran}
\usepackage{amsmath,amsfonts,amssymb}
\usepackage{mathtools}
\usepackage{amsthm}
\usepackage{algorithm}
\usepackage{array}
\usepackage[caption=false,font=normalsize,labelfont=sf,textfont=sf]{subfig}
\usepackage{textcomp}
\usepackage{stfloats}
\usepackage{url}
\usepackage{verbatim}
\usepackage{graphicx}
\usepackage{cite}
\usepackage{wrapfig}
\usepackage[dvipsnames]{xcolor}

\newcommand{\PLN}{\sct{SiLLy-N}}
\newcommand{\PLF}{\sct{SiLLy-F}}

\newcommand{\ProgNN}{\sct{ProgNN}}
\newcommand{\DFCNN}{\sct{DF-CNN}}
\newcommand{\Dn}{\mathbf{S}}

\newcommand{\DN}{\sct{DN}}
\newcommand{\zoo}{\sct{Model}\ \sct{Zoo}}
\floatname{algorithm}{Algorithm}

\usepackage{amsmath}
\DeclareMathOperator*{\argmax}{arg\,max}

\usepackage{bbm}
\usepackage[numbers]{natbib}
\usepackage{multirow} 
\usepackage{algorithm}
\usepackage{algpseudocode}

\providecommand{\sct}[1]{{\sc \texttt{#1}}}

\newcommand{\T}{^{\ensuremath{\mathsf{T}}}} % transpose

\providecommand{\mc}[1]{\mathcal{#1}}
\providecommand{\tmc}[1]{\tilde{\mathcal{#1}}}
\providecommand{\mb}[1]{\boldsymbol{#1}}

\providecommand{\mbb}[1]{\mathbb{#1}}

\newcommand{\Real}{\mathbb{R}}

\newtheorem{Def}{Definition}

\begin{document}

\title{Simple Lifelong Learning Machines}

\author{$\sigma$(Joshua T. Vogelstein,
Jayanta Dey$^*$)$^{1}$,
Hayden S. Helm,$^{1}$
Will LeVine,$^1$
Ronak D. Mehta,$^1$
Tyler M. Tomita,$^1$
Haoyin Xu,$^1$
Ali Geisa,$^1$
Qingyang Wang,$^1$
Gido M. van de Ven,$^{2,3}$
%Emily Chang,$^1$
Chenyu Gao,$^1$
Weiwei Yang,$^4$
Bryan Tower,$^4$
Jonathan Larson,$^4$
Christopher M. White,$^4$ and
Carey E. Priebe$^{1}$
        % <-this % stops a space
\thanks{$^1$Johns Hopkins University,
% $^2$Progressive Learning,
$^2$Baylor College of Medicine,
$^3$University of Cambridge,
$^4$Microsoft Research}% <-this % stops a space
\thanks{${\sigma}$ \textbf{denotes equal contribution (author order was decided using a coin flip and both authors have the right to list their name first in their CV)},
$*$ corresponding author: jdey4@jhmi.edu}}

\markboth{© 2025 IEEE. Accepted in IEEE TPAMI. DOI: 10.1109/TPAMI.2025.3595364}
{Shell \MakeLowercase{\textit{et al.}}: A Sample Article Using IEEEtran.cls for IEEE Journals}


\maketitle

 \begin{abstract}
In lifelong learning, data are used to improve performance not only on the present task, but also on  past and future (unencountered) tasks. 
While typical transfer learning algorithms can improve performance on future tasks, their performance on prior tasks degrades upon learning new tasks (called forgetting). Many recent approaches for continual or lifelong learning have attempted to \textit{maintain} performance on old tasks given new tasks. But striving to avoid forgetting sets the goal unnecessarily low. The goal of lifelong learning should be to use data to improve performance on both  future tasks (forward transfer) and past tasks (backward transfer). In this paper, we show that a simple approach---representation ensembling---demonstrates both forward and backward transfer in a variety of simulated and benchmark data scenarios, including tabular, vision (CIFAR-100, 5-dataset, Split Mini-Imagenet, Food1k, and CORe50), and speech (spoken digit), in contrast to various reference algorithms, which typically failed to transfer either forward or backward, or both. Moreover, our proposed approach can flexibly operate with or without a computational budget.
\end{abstract}


\section{Introduction}
\label{sec:introduction}

\IEEEPARstart{L}{earning} is a process by which an intelligent system improves performance on a given task by leveraging data~\citep{mitchell1999machine}. In classical machine learning,  the system is often optimized for a single task~\citep{Vapnik1971-um,Valiant1984-dx}. While it is relatively easy to \textit{simultaneously} optimize for multiple tasks (multi-task learning)~\citep{caruana1997multitask}, it has proven much more difficult to \textit{sequentially} optimize for multiple tasks~\citep{thrun1996learning, Thrun2012-sj}.  Specifically, classical machine learning systems, and natural extensions thereof, exhibit ``catastrophic forgetting'' when trained sequentially, meaning their performance on the prior tasks drops precipitously upon training on new tasks~\citep{mccloskey1989catastrophic,mcclelland1995there, doan2021theoretical}. 
However, learning could be lifelong, with agents continually building on past knowledge and experiences, improving on many tasks given data associated with any task.   
For example, in humans, learning a second language often improves performance in an individual's native language~\citep{Zhao2016-vo}.

% In the past 30 years, a number of  sequential task learning algorithms have attempted to overcome catastrophic forgetting \citep{de2021continual}. These approaches naturally fall into one of two camps. In one camp, the algorithm has fixed resources, and so must reallocate resources (essentially compressing representations) in order to incorporate new knowledge~\citep{kirkpatrick2017overcoming,zenke2017continual,li2017learning,schwarz2018progress,Finn2019-yv}. 

% Biologically, this corresponds to adulthood, where brains have a nearly fixed or decreasing number of cells and synapses. In the other camp, the algorithm adds (or builds) resources as new data arrive (essentially ensembling representations)~\citep{Ruvolo2013-hk, rusu2016progressive,Lee2019-eg, sodhani2020toward}.  
% Biologically, this corresponds to development, where brains grow by adding cells, synapses, etc. 

% Approaches from both camps demonstrate some degree of continual (or lifelong) learning~\citep{parisi2019continual}. In particular, they can sometimes learn new tasks faster due to prior learning on related tasks, while not catastrophically forgetting old tasks (see Appendix \ref{app:related_work} for a detailed discussion on the relevant algorithms). 

In lifelong learning, where tasks arrive sequentially, the ability to transfer knowledge across tasks is characterized by two complementary objectives: \emph{forward transfer} and \emph{backward transfer}. \emph{Forward transfer} facilitates accelerated learning in new tasks using previous knowledge. In contrast, \emph{backward transfer} evaluates the impact of new learning on previously encountered tasks. Achieving both positive forward and backward transfer is crucial for an effective lifelong learner. However, as we will demonstrate, many existing lifelong learning algorithms do not enable forward transfer to future tasks, and most do not exhibit positive backward transfer to previously learned tasks.

%With high enough sample sizes, some of these algorithms can transfer forward or backward, but transfer is more important in low sample size regimes \citep{Chen2016-dk, Lee2019-eg}.  This inability to effectively transfer in low-sample-size regimes has been identified as one of the key obstacles limiting the capabilities of artificial intelligence~\citep{Pearl2019-bp,Marcus2019-dj}.
% In this paper, we propose an algorithm that can flexibly operate in both resource constrained and resource growing modes.
 
\begin{figure*}[!ht]
    \centering
    \includegraphics[width=.7\linewidth]{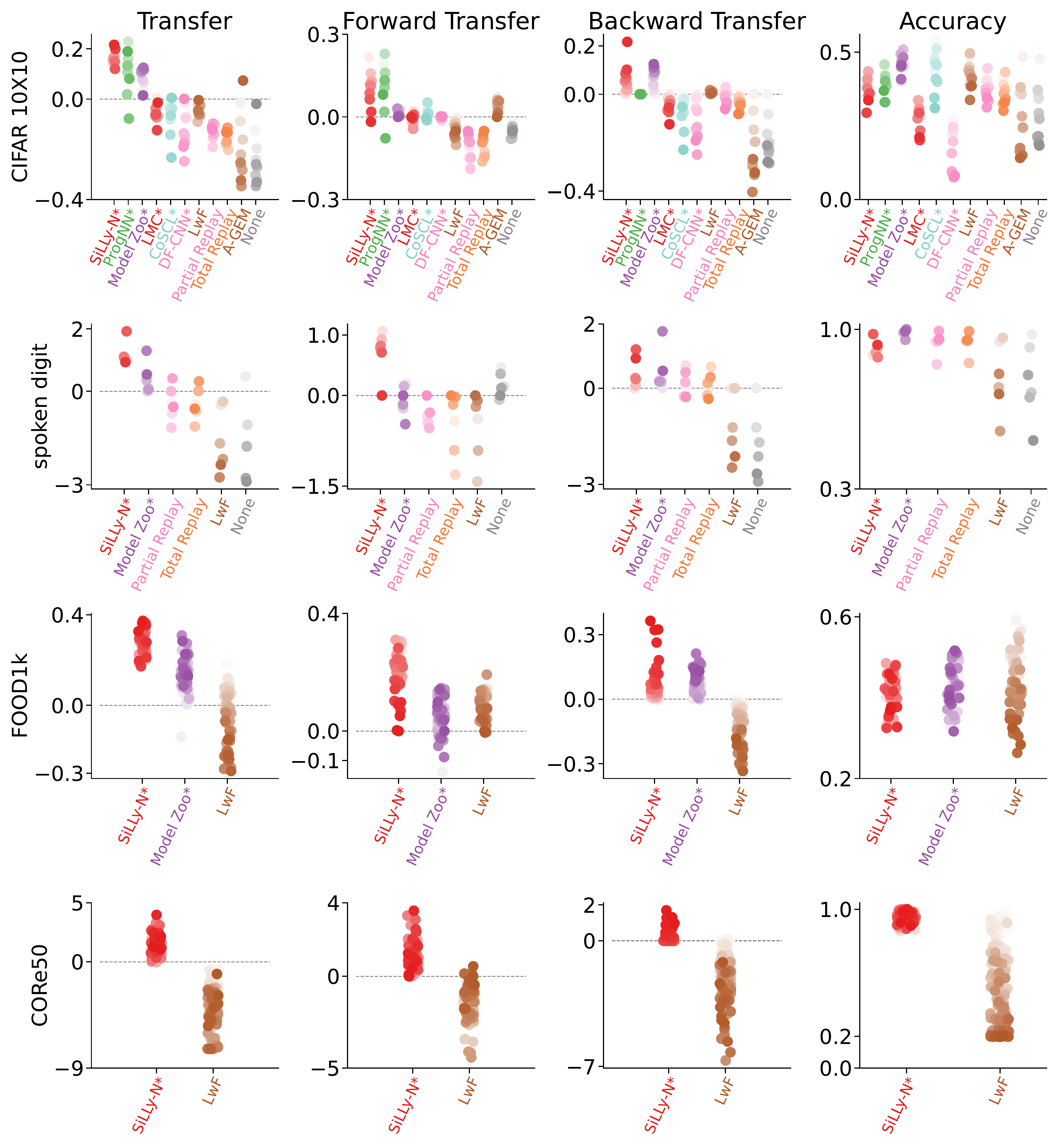}
    \caption{\textbf{Performance summary on different vision and speech benchmark datasets.} Columns are different evaluation criteria (see Section~\ref{sec:methods} for definitions, and Section~\ref{sec:real} for experimental details), each strip of colored dots corresponds to an algorithm (we introduce \PLN\ here) and each dot represents a task. 
    % In all the figure, 
    Older tasks have darker colors. Resource growing algorithms have a `*'. \sct{EWC}, \sct{O-EWC}, \sct{SI}, \sct{TAG}\ and \sct{ER} always perform worse than \sct{LwF}, and hence we do not show them in the plot. 
    \PLN\ (red) outperforms all reference algorithms in terms of overall (first column), forward (second column), and backward (third column). Importantly, such better transfer is achieved at high overall accuracy (fourth column). For CORe50 dataset, we are unable to run \zoo\ in the resource constrained environment of the experiment and \sct{LwF}\ completely forgets the first several tasks with the task-accuracies going down to the chance level ($20\%$). See Appendix Figure \ref{fig:strip_IQR} for the error bars for each dot in the figure.
    %\textit{Top:} Interquartile range for each dot in the top row.
    %More datasets are evaluated in Figure \ref{fig:all}. 
    % show better performance than \sct{Model Zoo} on CIFAR 10X10 (column 1), Speech (column 2), FOOD1k (column 3) which have relatively smaller sample size per task compared to those of Imagenet(column 4) and Five Dataset (column 5).
    }
    \label{fig:strip}
\end{figure*}

In this paper, we propose a general and simple approach for lifelong learning which can be used with many existing encoder models. Specifically, we focus  our approach  on ensembling deep  networks (Simple Lifelong Learning Networks, \PLN). 
Additionally, we demonstrate how the same approach can be generalized for lifelong learning based on  ensembling decision forests (Simple Lifelong Learning Forests, \PLF). Table~\ref{tab:tax} and Figure~\ref{fig:size} show our proposed approaches grow linearly with the task number and can flexibly operate with both growing and constant resources depending on the available computation budget. Moreover, we explore our proposed algorithm as compared to a number of reference algorithms on an extensive suite of numerical experiments that span simulation, vision datasets including CIFAR-100, 5-dataset, Split Mini-Imagenet, and Food1k, as well as the spoken digit dataset. 
Figure~\ref{fig:strip} illustrates that our algorithm outperforms all the reference algorithms in terms of forward, backward, and overall transfer on different vision and speech datasets.  
% This is the case for our proposed resource growing algorithms, using either previously proposed (appendix Figure \ref{fig:strip_overall}) or our novel evaluation criteria (Figure \ref{fig:strip}).
Ablation studies indicate the degree to which the amount of representation or storage capacity and replaying old task data impact performance of our algorithms.  
All our code and experiments are open source to facilitate reproducibility. 

The subsequent organization of the paper can be summarized as:
\begin{enumerate}
    \item Section \ref{app:related_work} presents a discussion of relevant algorithms, highlighting their architecture and computational complexity in comparison to \PLN\ and \PLF.
    \item Section \ref{sec:methods} introduces the learning environment and proposes three  transfer statistics.
    \item Section \ref{sec:algorithms} details the design of \PLN\ and \PLF, incorporating insights from various ensembling approaches.
    \item Sections \ref{sec:simulations} and \ref{sec:real} provide empirical evaluations of \PLN\ and \PLF\ on both simulated and benchmark datasets. 
    \item Section \ref{sec:discussion} concludes with a discussion of strengths, limitations, and directions for future research.
\end{enumerate}

\section{Related Works} \label{app:related_work}
In this work, we propose a lifelong learning approach based on representation ensembling which can flexibly operate in both resource growing and resource constrained modes. Existing resource constrained algorithms, such as \sct{EWC} \citep{kirkpatrick2017overcoming}, \sct{Online EWC} \citep{schwarz2018progress}, \sct{EWC++} \citep{chaudhry2018riemannian}, \sct{SI} \citep{zenke2017continual}, and  \sct{LwF} \citep{li2017learning}, use regularization techniques to exploit the stability-plasticity trade-off to mitigate forgetting. 

%Prior work in the literature illustrates that ensembling learners can yield huge advantages in a wide range of applications. For example, in classical machine learning, ensembling trees leads to state-of-the-art random forest \citep{breiman2001random} and gradient boosting tree algorithms \citep{Chen2016-fx}.  Similarly, ensembling networks shows promising results in various real-world applications \citep{qiu2014ensemble, potes2016ensemble}. 
On the contrary, the resource growing approaches add resources as they face more tasks. Authors from \citep{wang2003mining} used a weighted ensemble of learners in a streaming setting with distribution shift. \sct{TrAdaBoost} \citep{dai2007boosting} boosts ensemble of learners to enable transfer learning. In continual learning scenarios, many algorithms have been built on these ideas by ensembling dependent representations. For example, \sct{Learn++} \citep{polikar2001learn++} boosts ensembles of weak learners learned over different data sequences  in class incremental lifelong learning settings \citep{van2022three}.  \sct{Model Zoo} \citep{ramesh2021model} uses the same boosting approach in task incremental lifelong learning scenarios. 
Another group of algorithms, \ProgNN~\citep{rusu2016progressive} and \DFCNN~\citep{Lee2019-eg} learn a new ``column'' of nodes and edges with each new task, and ensembles the columns for inference (such approaches are commonly called `modular' now). The primary difference between \ProgNN~and \DFCNN~is that \ProgNN~has forward connections to the current column from all the past columns. This creates the possibility of forward transfer while freezing backward transfer. However, the forward connections in \ProgNN~render it computationally inefficient for a large number of tasks. \DFCNN~gets around this problem by learning a common knowledge base and thereby, creating the possibility of backward transfer.

% JOVO reproduce citet functionality
Recently, many other modular approaches have been proposed in the literature that improve on  \ProgNN's  capacity growth. These methods consider the capacity for each task being composed of modules that can be shared across tasks and grown as necessary. For example, \sct{PackNet} \cite{mallya2018packnet} starts with a fixed capacity network and trains for additional tasks by freeing up portion of the network capacity using iterative pruning. Veniat et al. \cite{veniat2020efficient} trains additional modules with each new task, and the old modules are only used selectively. \cite{ostapenko2021continual} improved the memory efficiency of the modular methods by adding new modules according to the complexity of the new tasks. Authors in \citep{mehta2021continual} proposed non-parametric factorization of the layer weights that promotes sharing of the weights between tasks. However, all of modular methods described above lack backward transfer because the old modules are not updated with the new tasks.  Dynamically Expandable Representation (DER) \citep{yan2021dynamically} proposed an improvement over the modular approaches where the model capacity is dynamically expanded and the model is fine-tuned by replaying a portion of the old task data along with the new task data. This approach achieves backward transfer between tasks as reported by the authors in the experiments. Another modular approach, proposed by Kang et al. \citep{kang2022forget}, uses sub-networks inspired by the lottery ticket hypothesis \citep{frankle2018lottery} and dynamically expands the network capacity with the number of tasks.

Another strategy for building lifelong learning machines is to use total or partial replay~\citep{tolias_architecture, robins1995catastrophic,shin2017continual, chaudhry2019tiny, buzzega2020dark}. Replay approaches
keep the old data and replay them when faced with new tasks to   mitigate catastrophic forgetting. Many recent approaches combine replay with other transfer learning strategies like knowledge distillation \citep{chen2021human} to improve storage efficiency of replay data. Recently, hypernetwork based old network weight generation accompanied by synthetic replay \citep{von2019continual} demonstrated potential of memory-efficient operation in a learning environment with a huge number of tasks. Another replay algorithm, \sct{MER} \citep{riemer2018learning}, aligns the gradient update between task datasets while replaying old task data. However, as we illustrated in the main text, previously proposed replay algorithms do not demonstrate positive backward transfer in our experiments, though they often do not forget as much as other approaches.

Our approach builds directly on previously proposed modular and replay approaches with one key distinction: in our approach, representations are learned independently.  %The conceptual advantage of growing statistically independent representations per task is similar to the conceptual advantage of growing statistically independent representations per tree in random forest: Brieman \citep{breiman2001random} showed that doing so asymptotically yields optimal performance.  
% Empirically, for low sample sizes random forests (which learn independent trees) typically outperform gradient boosted trees (which learn dependent trees)~\citep{caruana2004ensemble, Caruana2008-tb, Fernandez-Delgado2014-qu}. Because our approach of representation ensembling is similar to that of random forest, we expect learning independent representations to outperform learning dependent representations in these scenarios as well. This phenomenon is empirically shown in the main text Figure \ref{fig:strip}.
 Independent representations also have computational advantages, as doing so merely requires quasilinear time and space, and can be learned in parallel.

\section{Mathematical Framework} \label{sec:methods}

\subsection{The lifelong learning objective}

In a lifelong learning setting, we consider a sequence of tasks \(\mathcal{T} = \{1, 2, \dots, T\}\), where the tasks are known during training and testing. Each task \( t \in \mathcal{T} \) shares the same input space, \(\mathcal{X} \subset \mathbb{R}^D\), with class labels \(\mathcal{Y} = \{1, \dots, K_t\}\). The tasks arrive sequentially, and within each task \(t\), the dataset \(\mathbf{S}^t = \{(X_i, Y_i)\}_{i=1}^{n_t}\) is sampled $iid$ from a fixed distribution \(\mathcal{D}_t\), with \(\sum_{t=1}^T n_t = n\).  

 Given access to all observed datasets \( \bigcup_{t=1}^T \Dn^t \), the goal is to find a learner \( f \) that minimizes the overall generalization error across all tasks:  

\begin{equation} \label{eq:l2_def}
    \begin{array}{ll}         
         \textnormal{minimize}  &  \sum_{t=1}^T\mc{E}_f^t(\bigcup_{t'=1}^T\Dn^{t'}) \\ 
        \textnormal{subject to }\qquad  &f \in \mathcal{F}
    \end{array} \enspace ,
\end{equation} 

where $\mc{E}_f^t$ is the generalization error or expected risk on task $t$ and the learner will have access to a total of $T$ datasets after $T$ tasks, $\bigcup_{t=1}^T\Dn^t$ \footnote{More generally, we may have $J$ datasets, where $J \neq T$ and each dataset may be associated with the target distributions of multiple tasks. For simplicity, we do not consider such scenarios further at this time.}. Risk is defined as the expected task-specific loss $\ell_t: \mc{Y} \times \mc{Y} \rightarrow [0, \infty)$ in task $t$ (see \citep{dey2021towards} for a detailed formulation on different out-of-distribution learning scenarios). 
%This formulation ensures that the learner optimally selects hypotheses for each task while leveraging all available data to minimize the cumulative generalization error.

\subsection{Lifelong learning evaluation criteria}
\label{sec:evaluation-ceiterion}

%Our proposed statistics are similar to those simultaneously proposed by \citet{veniat2020efficient} (shown in Figure~\ref{fig:strip_overall}), except for the fact that we consider the ratio of errors rather than the difference of them in our definitions. As always, no single statistic can serve all purposes because it is a scalar summary of a corpus of data.  Please see Appendix~\ref{app:evaluation-ceiterion} where we illustrate scenarios in which existing metrics fail to provide the kinds of insights that we desire, therefore motivating us to clearly enumerate our desired properties.

Others have previously introduced criteria to evaluate transfer, including forward and backward transfer~\citep{LopezPaz2017GradientEM,Benavides-Prado2018-nv, diaz2018don, veniat2020efficient}. 
Pearl~\cite{judea2018gained} introduced the transfer benefit ratio, which builds directly off relative efficiency from classical statistics \citep{bickel2015mathematical}. 
% We refer to these and other such criteria as statistics that quantify lifelong learning performance (rather than metrics, which formally means a notion of distance). 
We define three notions of transfer building on relative efficiency.

\begin{Def}[Transfer] \label{def:LE}
Overall transfer of algorithm $ f $ for a given Task $ t $ is:
\begin{equation}\label{eq:transfer-efficiency}
\mathsf{Transfer}^t(f) := \log \frac{\mc{E}^t_f(\Dn^t)}{\mc{E}^t_f(\bigcup_{t'=1}^{T}\Dn^{t'})}.
\end{equation}

In words, Equation~\ref{eq:transfer-efficiency} quantifies the extent that a learner $f$, is able to improve the performance on task $t$, when using  the data on all other tasks, $(1,\dots,T)$. It does so  by computing the ratio of generalization error on task $t$ while only training on data from task $t$, relative to the generalization error on task $t$ when training on the data from \textit{all} tasks.
% 
% In Equation~\ref{eq:transfer-efficiency}, $\mathcal{E}$ carries a superscript $t$ to indicate the specific task for which the generalization error is evaluated, a subscript $f$ to represent the learner, and a dataset enclosed in parentheses `$(\cdot)$' to denote the data that the learner $f$ has access to.

%  We say that an algorithm $ f $ has  transferred to task $t$ from all the tasks up to $T$
% if and only if 
% \begin{equation}
%     \mathsf{Transfer}^t(f) > 0.
% \end{equation}
 
\end{Def}   

Forward transfer quantifies how much performance a learner transfers forward to future tasks, given prior tasks.
\begin{Def}[Forward Transfer] \label{def:FLE}
The forward transfer of $ f $ for  task $t$ is :
 \begin{equation}
 \label{eq:FLE}
    \mathsf{Forward~Transfer}^t(f) := \log \frac{\mc{E}^t_f(\Dn^t)}  {\mc{E}^t_f(\bigcup_{t'=1}^t\Dn^{t'})}.
\end{equation}
Forward transfer is identical to transfer as defined in Equation~\ref{eq:transfer-efficiency}, except that it only considers the relation between training on data from task $t$ as compared to training on the prior tasks, $(1,\dots,t)$, excluding future tasks, $(t+1,\dots,T)$. 

% We say an algorithm (positively) forward transfers for task $t$ if and only if:
% \begin{equation}
%     \mathsf{Forward~Transfer}^t(f) >0.
% \end{equation}
 
\end{Def}

% Backward learning efficiency is the ratio of 
% the generalization error on task $t$ of the learned hypothesis with
% (i)  access to the data up to and including the last observation from task $ t $, to (ii)  access to the entire dataset. 
% Thus, this quantity measures the relative effect of future task data on the performance on Task $ t $. 
Backward transfer quantifies how much a learner transfers backward to previously observed tasks, in light of new tasks. 
\begin{Def}[Backward Transfer] \label{def:BLE}
The backward transfer of $ f $ for Task $t $ is:
\begin{equation}
\label{eq:BLE}
    \mathsf{Backward~Transfer}^t(f) := \log \frac{\mc{E}^t_f(\bigcup_{t'=1}^t\Dn^{t'})}{\mc{E}^t_f(\bigcup_{t'=1}^{T}\Dn^{t'})}.
\end{equation}
Backward transfer is also like Equation~\ref{eq:transfer-efficiency}, except it compares the performance when training on all prior tasks, $(1,\dots,t)$, with training on all tasks, $(1,\dots,T)$. 

% We say an algorithm backward transfers to  Task $t$ from  all the future tasks up to $T$ if and only if: 

% \begin{equation}
%   \mathsf{Backward~Transfer}^t(f) >0.  
% \end{equation}
 
\end{Def} 
Note that  $\mathsf{Transfer}$ can be decomposed into $\mathsf{Forward~Transfer}$ and $\mathsf{Backward~Transfer}$ (see Appendix \ref{app:decomposition_tx} for details):
\begin{equation} \label{eq:decompose}
\mathsf{Transfer}^t(f) = \mathsf{Forward~Transfer}^t(f) + \mathsf{Backward~Transfer}^t(f).
\end{equation}

The above equation underscores the sequential progression of tasks in lifelong learning, indicating that the total knowledge a learner acquires is derived from both previously encountered tasks (forward transfer) and those yet to be encountered (backward transfer). While forward transfer is relatively straightforward to achieve \citep{chakraborty2022efficient}, backward transfer presents a significant challenge due to the issue of catastrophic forgetting. In particular, a learner that demonstrates strong forward transfer ($\mathsf{Forward~Transfer}^t(f) > 0$) but experiences negative backward transfer ($\mathsf{Backward~Transfer}^t(f) < 0$) from all future tasks will eventually lose the learned knowledge, resulting in $\mathsf{Transfer}^t(f) \leq 0$ for the task.
% Of note, previously proposed quantities for evaluating lifelong learning performance could likely also be decomposed as we have done, though we have not seen it in the literature (see appendix Figure~\ref{fig:strip_overall}). 

Another paper \citep{veniat2020efficient}, concomitantly introduced transfer and forgetting (backward transfer). Their statistics are the same as ours, except they do not use a $\log$.  We opted for a $\log$ to address numerical stability issues in comparing small numbers. 
Because log is a monotonic function, the order of ranking algorithms is preserved (Appendix Figure~\ref{fig:strip_overall} shows a version of Figure~\ref{fig:strip}, but using Veniat's statistics, which is nearly visually identical).
% 
% rediscovered transfer, though they define it as the difference between errors, rather than the difference between the log of errors (see appendix Figure~\ref{fig:strip_overall} for a summary of our experiments using Veniat's statistics). Because log is a monotonic function, the order of ranking algorithms is preserved. We prefer log because log is more sensitive to small changes and typically more numerically stable~\citep{press2007numerical}. 
% 
By virtue of introducing $\mathsf{Forward~Transfer}$ here, we can identify the inherent trade-off between forward and backward transfer, for a fixed amount of total transfer. Apart from the above statistics, we also report accuracy per task.

\begin{Def}[Accuracy] \label{def:acc}
The accuracy of algorithm $ f $ on task $t$ after observing total $T$ datasets is:
 \begin{equation}
 \label{eq:acc}
    \mathsf{Accuracy}^t(f) := 1 -  {\mc{E}^t_f(\bigcup_{t'=1}^T\Dn^{t'})}.
\end{equation}

\end{Def}

% However, unlike \citet{veniat2020efficient}, we consider two extensions of $\mathsf{Transfer}$ to evaluate a lifelong learning algorithm while respecting the streaming nature of the tasks. $\mathsf{Forward~Transfer}$ is the ratio of the generalization error of the learning algorithm on task $t$ with (i)  access only to Task $t$ data, (ii) access to the data up to and including the last observation from Task $ t $. This quantity measures the relative effect of previously seen out-of-task data on the performance on Task $ t $.

\subsection{Computational Taxonomy of Lifelong Learners}
\label{sec:taxonomy}
\begin{table*}[!htb]
% \centering
\caption{\textbf{A computational taxonomy of lifelong learners.} We show  soft-O notation ($\tmc{O}(\cdot, \cdot)$) as a function of total training samples, $n = \sum_t^T n_t$, where $n_t$ is the number of training samples for the $t^{th}$ task and total task, $T$, as well as the common setting where $n$ is proportional to $T$. Parametric, semi-parametric, and non-parametric algorithms have parameters that remain fixed, grow slowly, and scale proportionally to $n$, respectively.}
    \label{tab:tax}
    \begin{center}
\resizebox{0.9\textwidth}{!}{ 
\begin{tabular}{|l|l|l|l|l|l|l|}
    \hline
    \textbf{Parametric}
        &  \textbf{Capacity}  
        & \multicolumn{2}{c|}{\textbf{Space}}
        & \multicolumn{2}{c|}{\textbf{Time}}
        & \textbf{Examples}\\
    \cline{2-6}
         &  ($n,T$)  &  ($n,T$) & ($n \propto T$) & $(n,T)$ & $(n \propto T)$ & 
        %  \textbf{Examples}
         \\
            \hline parametric & $1$ & $1$ & $1$ & $1$ & $1$ & \PLN-M, \PLF-M \\
    \hline parametric     & $1$    & $1$   & $1$   & $n $ & $n$ & \sct{O-EWC} \citep{schwarz2018progress}, \sct{SI} \citep{zenke2017continual}, \sct{LwF} \citep{li2017learning}\\ 
    \hline parametric     & $1$  & $T$ & $n$ & $nT$ & $n^2$ & \sct{EWC} \citep{kirkpatrick2017overcoming}\\
    \hline parametric     & $1$    & $n$   & $n$   & $nT$ & $n^2$ & \sct{Total Replay} \\
    % \hline parametric     & $1$    & $n$   & $n$   & $nT $ & $n$ & Partial Replay \\
    \hline semi-parametric & $T$ & $T^2$ & $n^2$ & $n T$ & $n^2$ & \ProgNN \citep{rusu2016progressive}\\
    \hline semi-parametric & $T$ & $T$   & $n$   & $n$ & $n$ & \sct{DF-CNN} \citep{Lee2019-eg}\\
    % 
    % \hline 
    \hline semi-parametric & $T$ & $T+n$ & $n$ & $n$ & $n$ & \PLN, \zoo \citep{ramesh2021model}, \sct{DER} \citep{yan2021dynamically}, \sct{LMC}\citep{ostapenko2021continual} \\
    \hline
         non-parametric & $n$ & $n$ & $ n$ & $n$ & $n$ & \PLF, \sct{IBP-WF} \citep{mehta2021continual}\\
    \hline
\end{tabular}
}
\end{center}
\end{table*}
   
In the past 30 years, a number of  lifelong learning algorithms have attempted to overcome catastrophic forgetting \citep{de2021continual, yuan2023survey}. These algorithms can be broadly classified into three categories: parametric, semi-parametric, and non-parametric approaches, based on their representational capacity and how it scales with the number of tasks and data samples. Table \ref{tab:tax} shows a computational taxonomy of lifelong learners as a function of the sample size $n$ and the total task $T$, as well as the common scenario where the sample size is fixed per task and therefore proportional to the number of tasks, $n \propto T$. For time complexity, we only consider the training time complexity. The space complexity of the learner refers to the amount of memory space needed to store the learner \citep{kuo2003optimal}. We also study the representation capacity of these algorithms. Capacity is defined as the size of the subset of hypotheses that is achievable by the learning algorithm \citep{zhang2021understanding}. For a detailed discussion and assumptions behind the complexity analysis, see Appendix~\ref{app:complexity}.

% JOVO makes these more clearly definitions

\section{Representation ensembling algorithms}\label{sec:algorithms}

\begin{figure*}[!htb] 
    \centering
    \includegraphics[width=0.8\textwidth]{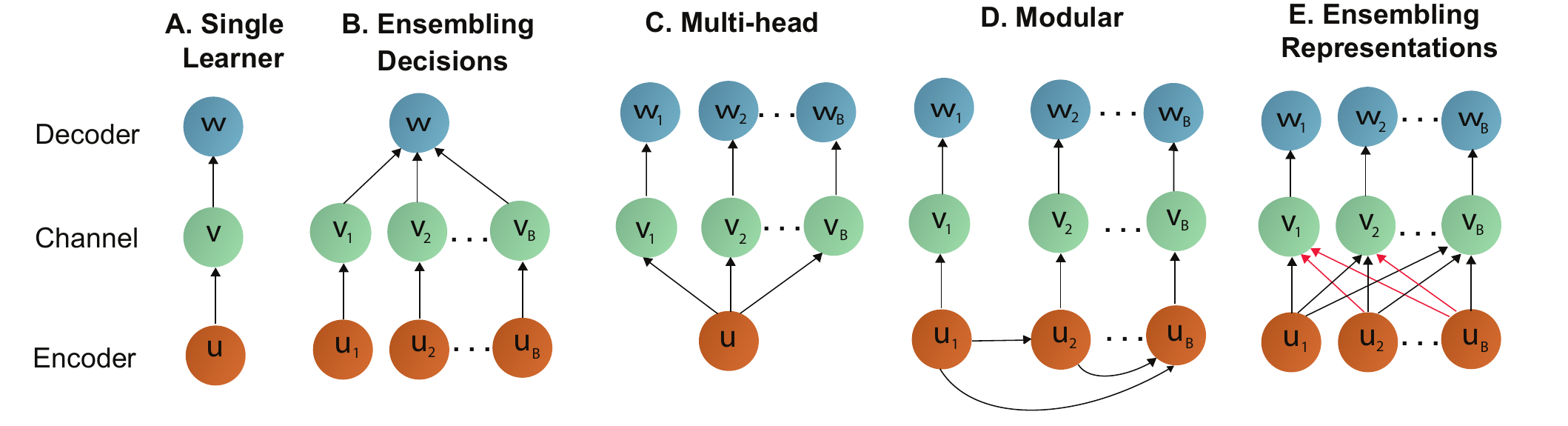}
    \caption{Schemas of composable hypotheses. A. Single task learner. B. Ensembling decisions (as output by the channels) is a well-established practice, including random forests and gradient boosted trees. C. Learning a joint representation or D. learning future representations depending on the past encoders was previously used in lifelong learning scenarios, but encoders were not trained independently as in E. Note that the new encoders in E interact with the previous encoders through the channel layer (indicated by red arrows), thereby, enabling backward transfer. Again the old encoders interact with the future encoders (indicated by black arrows), thereby, enabling forward transfer.
    }
    \label{fig:schematic}
\end{figure*}
% TODO it is worth looking at the paper from Guido, and seeing if their schematics improved upon ours, and then consider updating ours.  for example, are these the schematics for training, or inference, or what?

Shannon proposed that a learned hypothesis can be decomposed into three components: an encoder, a channel, and a decoder~\citep{Cover2012-sl,Cho2014-ew}: $ h(\cdot) = w \circ v \circ u(\cdot) $. Figure \ref{fig:schematic} shows these three components as the building blocks of different learning schemas. 
The encoder, $u: \mc{X} \mapsto \tilde{\mc{X}}$,  maps an $\mc{X}$-valued input into an internal representation space $\tilde{\mc{X}}$~\citep{Vaswani2017-lq,Devlin2018-lk}.
The channel  $v: \tilde{\mc{X}} \mapsto \Delta_{\mc{Y}}$ maps the transformed data into a posterior distribution (or, more generally, a score). 
Finally, a decoder   $w: \Delta_{\mc{Y}} \mapsto \mc{Y}$,  produces a predicted label.

A canonical example of a single learner depicted in Figure \ref{fig:schematic}A is a decision tree. Importantly, one can subsample the training data to learn different components of the tree~\citep{breiman1984classification, denil14, Athey19}.  For example, one can use a portion of data to learn the tree structure (which is the encoder). Then, by pushing the remaining data (sometimes called the `out-of-bag' data) through the tree, one can learn posteriors in each leaf node (which are the channel). The channel thus gives scores for each data point denoting the probability of that data point belonging to a specific class. Using separate sets of data to learn the encoder and the channel results in less bias in the estimated posterior in the channels as in `honest trees'~\citep{breiman1984classification, denil14, Athey19}. Finally, the decoder provides the predicted class label using $\argmax$ over the posteriors from the channel.
% \footnote{In coding theory, these three functions are frequently  called the encoder, channel, and decoder, respectively}
% See Appendix

One can generalize the above decomposition by allowing for multiple encoders, as shown in  Figure~\ref{fig:schematic}B.
Given $B$ different encoders, one can attach a single channel to each encoder, yielding $B$ different channels.  
Doing so requires generalizing the definition of a decoder so that it would operate on multiple channels. 
Such a decoder ensembles the \textit{decisions}, because here each channel provides the final output based on the encoder. This is the learning paradigm behind 
bagging~\citep{Breiman1996-yz} and
boosting~\citep{Freund1995-md}; indeed, decision forests are a canonical example of a decision function operating on an ensemble of $B$ outputs~\citep{breiman2001random}.
% A decision forest learns $B$ different decision trees, each of which has a tree structure corresponding to an encoder.  Each tree is assigned a channel that outputs each tree's vote that an observation is in any class. The decoder outputs the most likely class averaged over the trees.

Although the task specific structure in Figure~\ref{fig:schematic}B can provide useful decision on the corresponding task, they cannot, in general, provide meaningful decisions on other tasks, because those tasks might have completely different class labels. Therefore, in the multi-head structure (Figure~\ref{fig:schematic}C) a single encoder is used to learn a joint representation from all the tasks, and a separate channel is learned for each task to get the score or class conditional posteriors for each task, which is followed by each task specific decider~\citep{kirkpatrick2017overcoming, schwarz2018progress, zenke2017continual}. 

Modular approaches, such as \ProgNN\ and \sct{LMC} (Figure~\ref{fig:schematic}D), have both multiple encoders and decoders. Connections from past to future encoders enables forward transfer. However, they freeze backward transfer.

% Further modification of the multi-head structure allows ProgNN or other modular approaches to learn separate encoder for each task with forward connections from the past encoders to the current one (Figure~\ref{fig:schematic}D). This creates the possibility of having forward transfer while freezing backward transfer. 
%Note that if the encoders are learned independently across different tasks, they may have learned useful \textit{representations} that the tasks can mutually leverage. 

Our approach also uses multiple encoders and decoders (Figure \ref{fig:schematic}E). Unlike modular approaches, we allow interaction among encoders through the channels, including both forward and backward interactions.
% This scheme requires generalizing the definition of a channel so that it can operate on multiple encoders. 
The result is that the channels \textbf{ensemble representations} (learned by the encoders), rather than decisions (learned by the channels as in Figure \ref{fig:schematic} B). 
In our algorithms, we push all the data through each encoder, and each channel learns and ensembles across all encoders.
% In this scenario, generalizing with bagging and boosting, the ensemble of channels then feeds into \textit{each}  task specific decoder. 
When each encoder has learned complementary representations, the channels can leverage that information to improve over single task performance. 
This approach has applications in few-shot and multiple task scenarios, as well as lifelong learning. 

\subsection{Our representation ensembling algorithms}
\begin{figure}%{.2}{0.5\textwidth}
    \centering
    \includegraphics[width=1.0\linewidth]{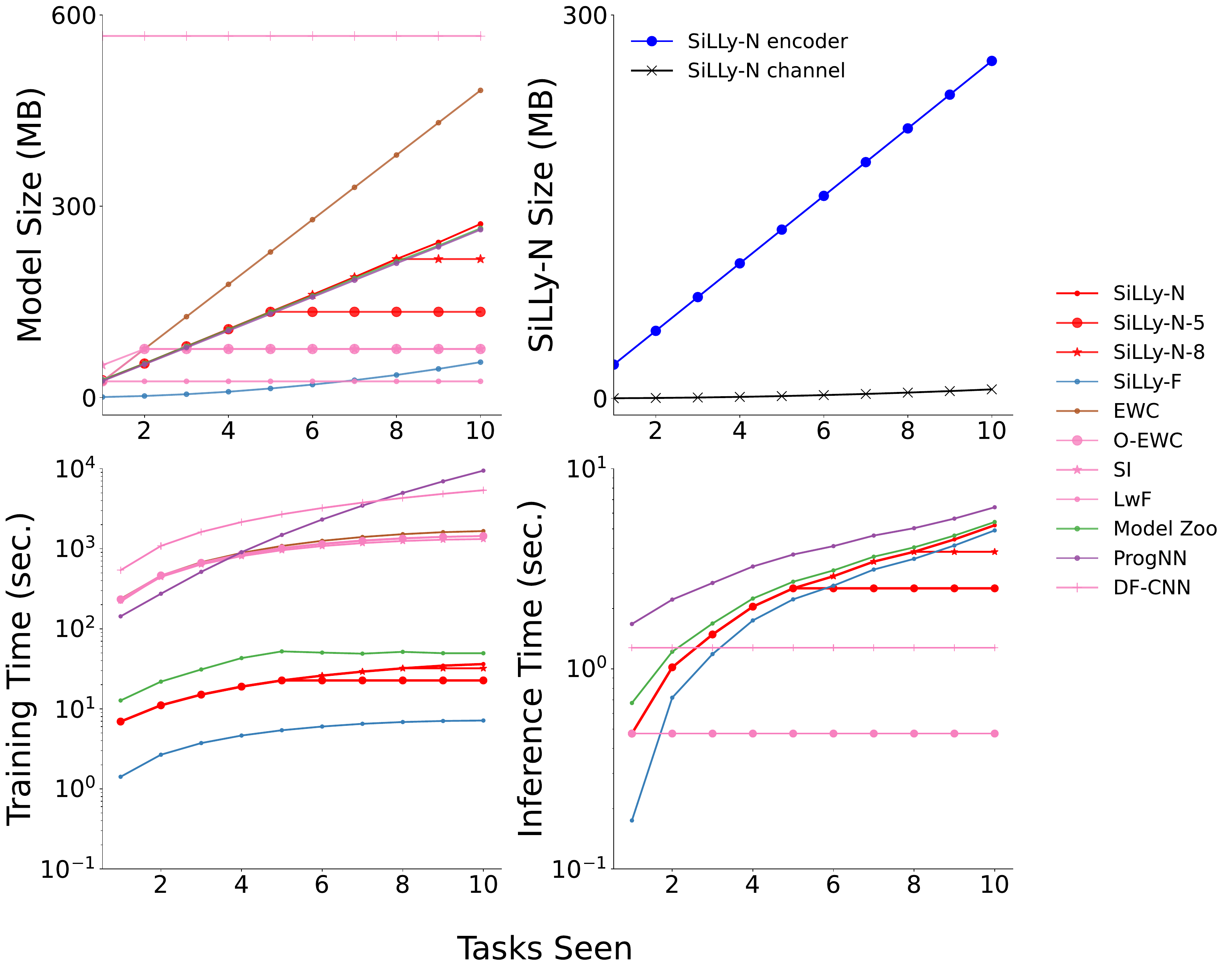}
    \caption{\textbf{Space and time complexity as a function on number of tasks in CIFAR 10X10.} Section \ref{sec:real} describes the experimental setup and architecture of the convolutional network used.
    \textit{Top left:} Model size, 
    \sct{SiLLy}\ can flexibly operate between resource constrained and growing modes. \textit{Top right:} Memory consumed by \PLN\ is dominated by the encoder size. \textit{Bottom left:} \sct{SiLLy} takes less training time compared to other baselines. \textit{Bottom right:} Inference time taken by different lifelong learners for $1000$ samples. Bottom row has log y-axis.
    % the experiment while using the codes provided by the authors. 
    }
    \label{fig:size}
\end{figure}

Figure \ref{fig:schematic}E shows a general structure of our algorithm. As data from a new task arrives, the algorithm first builds a new encoder. Then, it builds the channel for this new task by pushing the new task data through \textit{all} existing encoders.
 Thus the channel integrates information across all existing encoders using the new task data, thereby enabling forward transfer.  At the same time, if it stores old task data (or can generate such data), it can push that data through the new encoders to update the channels from the old tasks, thereby enabling backward transfer. In either case, new test data are passed through all existing encoders and corresponding channels to make a prediction 
 % As we will show empirically, our two ensemble methods achieve both forward and backward transfer on several benchmark datasets 
 (see appendix for detailed description of this approach).

%The key to both of our algorithms is the realization that both forests and networks partition feature space into a union of polytopes~\citep{Priebe2020-jn}.  Thus, the internal representation learned by each can be considered a sparse vector encoding which polytope a given sample resides in. We can combine the discriminative information over different sets of polytopes learned over different tasks by populating the polytopes with the corresponding task data and thereby, learn a channel for that specific task.

\subsubsection{Simple Lifelong Learning Networks in resource growing mode}
\label{synn}
A Simple Lifelong Learning Network (\PLN) ensembles deep networks.
% is a  deep  network (DN) based  instance of ensembling representations. 
For each task, the encoder $ u_{t} $ in \PLN\ is the ``backbone'' of a deep network (\DN). Thus, each $u_t$ maps an element of $ \mc{X} $ to an element of $ \mbb{R}^{d} $, where $ d $ is the number of neurons in the penultimate layer of \DN. The channels are learned by averaging the outputs from decision forests~\citep{Amit1997-nd, breiman2001random} trained on the $d$ dimensional representations of $\mc{X}$. See Appendix Figure \ref{fig:forest_voter} where we conduct experiments on a CIFAR 10X10 (described later in Section \ref{sec:cifar10X10}) while varying the number of decision trees per channel. Appendix Figure \ref{fig:forest_voter} shows that decision forest-based channels are robust to hyperparameter perturbation.
%$ k $-Nearest Neighbors ($k$-NN)~\citep{Stone1977-fi} trained over the $d$ dimensional representations of $\mc{X}$. Note that the channel is trained on the $d$ dimensional outputs from the encoders which is much smaller in size than the original training data and hence, the $k$-NN channels are inexpensive storage-wise  (shown later in Figure \ref{fig:size}). 
% Being said that, decision trees and 
Other algorithms could also be used to learn the channels, though we do not pursue them here.
% However, to keep the content of the paper concise, we will not pursue other ways of learning the channels in this paper.
%Recall that a $k$-NN, with $ k $ chosen such that as the number of samples goes to infinity, $ k $ also goes to infinity, while $ \frac{k}{n} \to 0 $, is a universally consistent classifier~\citep{Stone1977-fi}. We use $k = 16\log_2{n}$, which satisfies these conditions.
% The decoder is the same as above.  
The decoder $w_t$ outputs the $\argmax$ to produce a single prediction.

\subsubsection{Simple Lifelong Learning Networks in resource constrained mode}
The above resource growing approach is ideal when the upcoming tasks become more and more complex and there is no constraint imposed by the computation and storage budget available. However, real-world scenarios often impose computational constraints.
% in practice, as we will show later, lifelong learning tasks have finite complexity and we can learn all the tasks in the environment with a finite number of encoders. 
In the constant resource mode, we stop building new encoders after we have reached the computation and the storage budget imposed by the user. As new tasks arrive, we only learn new channels associated with new tasks using the old encoders. Note that this approach completely excludes the need to save old task data after we have reached the budget.

Hereafter, we will use suffix `-M' after the algorithm name whenever we use resource constrained operation of \PLN. Here $M$ is the total number of encoder allowed by the budget.
% \PLN\ differs from ProgNN in two key ways. First, recall that ProgNN builds a new neural network ``column'' for each new task, and also builds lateral connections between the new column and all previous columns. \textit{In contrast, \PLN\ excludes those lateral connections, thereby greatly reducing the number of parameters and train time.}  Moreover, this makes each representation independent, thereby potentially avoiding interference across representations.  Second, for inference on task $j$ data, assuming we have observed tasks up to $J > j$, \ProgNN~only leverages representations learned from tasks up to $j$, thereby excluding tasks $j+1, \ldots, J$. \textit{In contrast, \PLN\ leverages representations from all $J$ tasks, a key difference which enables backward transfer.}  

\subsubsection{Additional realization of our approach using random forest as encoder}
\label{synf}
Simple Lifelong Learning Forest (\PLF) ensembles decision trees or forests. 
% is a decision forest-based instance of ensembling representations.  
% 
For each task, the encoder $u_t$ of \PLF\ is the representation learned by a decision forest. 
%The leaf nodes of each decision forest partition the input space $ \mc{X}$ into polytopes~\citep{breiman1984classification}. 
%The representation of $ x \in \mc{X} $ corresponding to a single tree can be a one-hot encoded $L_b$-dimensional vector with a 1 in the location corresponding to the leaf $x$ falls into of tree $b$. Conceptually, this corresponds to partitioning the input space into $L_b$ polytopes with $x$ falling into one of them.
%The representation of $ x $ resulting from the collection of trees simply concatenates the $B$ one-hot vectors from the $B$ trees. 
%Thus, the encoder $ u_{t} $ is the mapping from $ \mc{X}$ to a $B$-sparse vector of length $\sum_{b=1}^B L_b$. In other word, $u_t$ is a collection of polytopes learned over the input space.
The channel then learns the class-conditional posteriors by populating the forest leaves with out-of-task samples, as in ``honest trees''~\citep{breiman1984classification, denil14, Athey19}. 
Each channel outputs the posteriors averaged across the collection of forests learned over different tasks. 
The decoder $w_t$ outputs the argmax to produce a single prediction. %, as per~\eqref{eqn:ll_hypothesis}. 
% Recall that honest decision forests are universally consistent classifiers and regressors~\citep{Athey19}, 
% meaning that with sufficiently large sample sizes, under suitable though general assumptions, they will converge to minimum risk.
% Thus, the single task version of this approach simplifies to an approach called ``Uncertainty Forests''~\citep{Guo2019-xe}. %Table~\ref{tab:hyperparameter_table} in the appendix lists the hyperparameters used in the CIFAR experiments. 

Note that the amount of additional representation capacity added per task by \PLF\ is a function of the amount and complexity of the data for a new task.  Contrast this with \PLN\, and other deep net based modular or representation ensembling approaches, which \textit{a priori} choose how much additional representation to add, prior to seeing all the new task data. So, \PLF\ has capacity, space complexity, and time complexity scale with the complexity and sample size of each task. In contrast,  \ProgNN, \PLN\ (and others like it) have a fixed capacity for each task, even if the tasks have very different sample sizes and complexities. 

Figure~\ref{fig:size} top left shows the model size for our proposed approach grows linearly with the number of tasks. Moreover, the memory consumed by the new channels is negligible compared to the memory required to store the encoders (Figure~\ref{fig:size} top right). The time required for inference on $1000$ testing points (Figure~\ref{fig:size} bottom right) is an order of magnitude lower compared to the time required to train a new encoder with $500$ samples(Figure~\ref{fig:size} bottom left).
\section{Simulation Data Study}
\label{sec:simulations} 

\begin{figure*}[!htb]
    \centering
    \includegraphics[width=.7\linewidth]{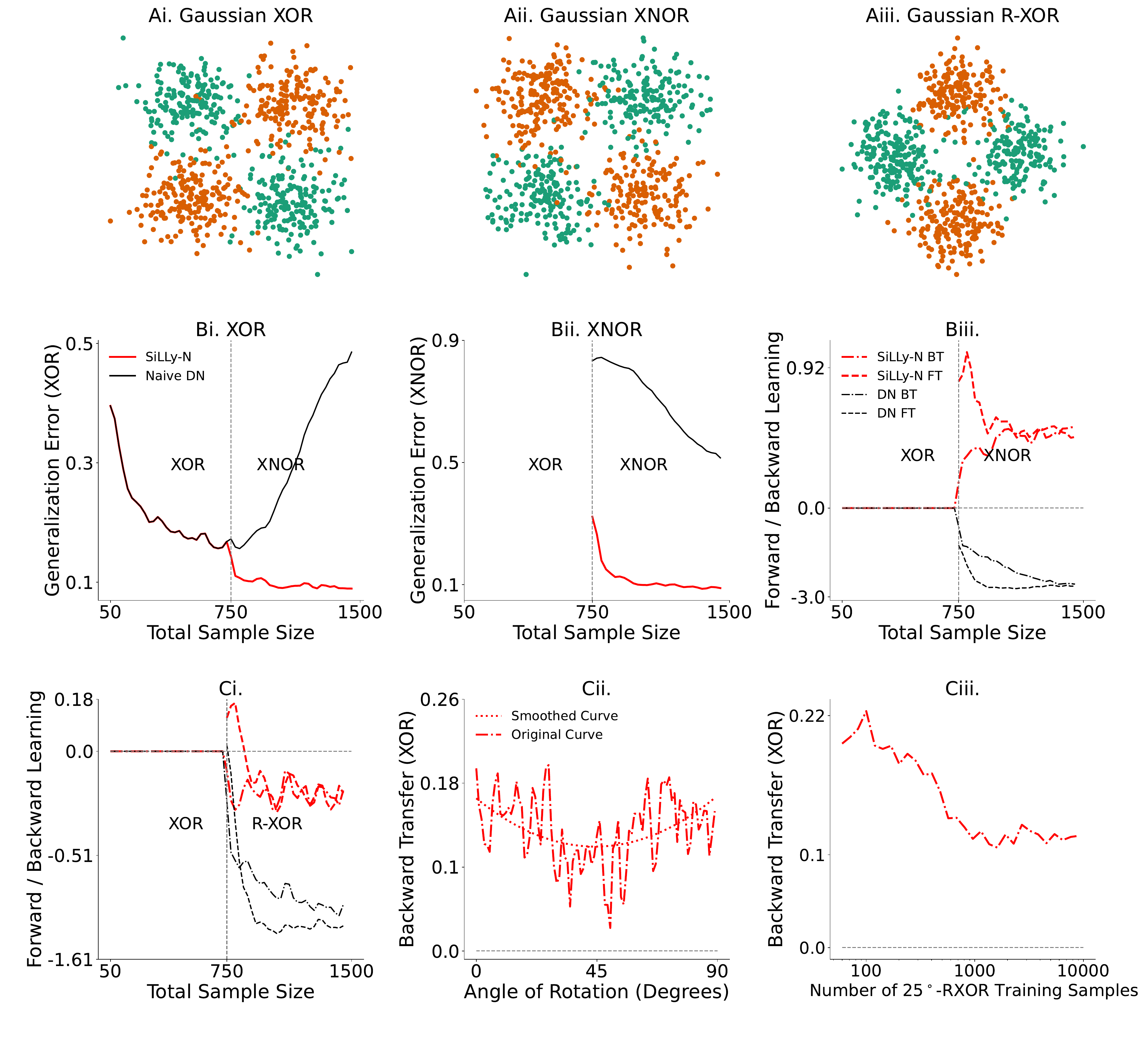}
    \caption{\textbf{\PLN\ demonstrates forward and backward transfer.} 
    % The learner is trained from scratch for each sample size so that we can observe the impact of increasing sample size on our algorithms.
    (\textit{A}) 750 samples from:   (\textit{Ai})  Gaussian XOR,  (\textit{Aii}) XNOR, which has the same optimal discriminant boundary as XOR, and (\textit{Aiii}) R-XOR, which has a discriminant boundary that is uninformative, and therefore adversarial, to XOR. (\textit{Bi}) Generalization error for XOR, and (\textit{Bii}) XNOR.  
    % of both \PLF\ (red), \RF\ (green), \PLN\ (blue), \DN\ (dark orange).  
    \PLN\ outperforms \DN\ on XOR when XNOR data is available, and on XNOR when XOR data are available. 
    % The same result is true for \PLN\ relative to \DN. 
    (\textit{Biii})
     Forward and backward transfer of \PLN\ are positive for all sample sizes. 
     %Forward and backward transfer for \PLN\ is higher than \DN\ for all sample sizes.
     % Again, FLE and BLE is higher for \PLN compared to those of \DN. 
    %  The forward (dashed) and backward (solid) curves 
    %  are the ratio of the generalization error of \PLF\ to \RF. \PLF\ 
    %  demonstrates both forward and backward transfer in this environment. 
     (\textit{Ci})
     In an adversarial task setting, 
     % ($100$ samples of XOR followed by $100$ samples of R-XOR),
     \PLN\ gracefully forgets XOR, whereas \DN\  catastrophically forget and interfere. 
    (\textit{Cii}) $\mathsf{Backward~Transfer}$ is maximum positive with respect to XOR when the optimal decision boundary of $\theta$-XOR is similar to that of XOR (e.g. angles far from $45^\circ$), and negative otherwise. The dashed line shows the regression line fitted through the original points. 
    (\textit{Ciii}) $\mathsf{Backward~Transfer}$ is a nonlinear function of the source training sample size (XOR sample size is fixed at $500$).
    }
    \label{fig:xor-xnor}
\end{figure*}

% \phantomsection
\subsection{Forward and backward transfer in  a simple environment}

Consider a very simple two-task environment: Gaussian XOR and Gaussian Exclusive NOR (XNOR) (Figure~\ref{fig:xor-xnor}A, see Appendix~\ref{app:sim} for details).
The two tasks share the exact same discriminant boundaries: the coordinate axes.  
Thus, transferring from one task to the other merely requires learning a bit flip of the class labels. We sample a total 750 samples from XOR, followed by another 750 samples from XNOR. 

 \PLN\ and deep network (\DN) achieve the same generalization error on XOR when training with XOR data (Figure~\ref{fig:xor-xnor}Bi). But because \DN\ does not account for a change in task, when XNOR data appear, \DN\ performance on XOR deteriorates (it catastrophically forgets). 
In contrast,  \PLN\ continues to improve on XOR given XNOR data, demonstrating backward transfer.
Now consider the  generalization error on \textit{XNOR} (Figure~\ref{fig:xor-xnor}Bii).  Both \PLN\ and \DN\ are at chance levels for XNOR when only XOR data are available.  When XNOR data are available, \DN\ must unlearn everything it learned from the XOR data, and thus its performance on XNOR starts out nearly maximally inaccurate, and quickly improves.  On the other hand, because \PLN\ can leverage the encoder learned using the XOR data, upon getting \textit{any} XNOR data, it immediately performs quite well, and then continues to improve with further XNOR data, demonstrating forward transfer 
%  
%  In this setting, forward learning efficiency is the expected ratio of generalization errors for XNOR, and backward learning efficiency is the ratio of the generalization errors for XOR 
 (Figure~\ref{fig:xor-xnor}Biii).
\PLN\ demonstrates positive forward and backward transfer for all sample sizes, whereas \DN\ fails to demonstrate neither forward nor backward transfer, and eventually catastrophically forgets the previous tasks. 
%Results for \PLN\ and \DN\ are qualitatively similar to those of \PLF\ and \RF\, respectively.
% Qualitatively similar results are visible for \PLN ~and \DN ~in Figure~\ref{fig:xor-xnor}.

\subsection{Forward and backward transfer for adversarial tasks}

%Statistics has a rich history of \textit{robust learning}~\citep{huber1996robust, ramoni2001robust}, and machine learning has recently focused on \textit{adversarial learning}~\citep{Bruna2013-iq, zhang2018mitigating, zhang2020attacks, lowd2005adversarial}.  
%However, in both cases the focus is on adversarial \textit{examples}, rather than adversarial \textit{tasks}. 
In the context of lifelong learning, we informally define a task $t$ to be adversarial with respect to task $t'$ if the true joint distribution of  task $t$, without any domain adaptation,  impedes performance on task $t'$.  In other words, training data from task $t$ can only add noise, rather than signal, for task $t'$.  An adversarial task for Gaussian XOR is Gaussian XOR rotated by $45^\circ$ (R-XOR) (Figure~\ref{fig:xor-xnor}Aiii).  Training  on R-XOR therefore impedes the performance of \PLN\  on XOR, and thus backward transfer becomes negative, demonstrating graceful forgetting \citep{alijundi2018memory} (Figure~\ref{fig:xor-xnor}Ci). 
%Because R-XOR is  more difficult than XOR for \PLF\ (because the discriminant boundaries are oblique~\citep{Tomita2020-xe}), and because the discriminant boundaries are  learned imperfectly with finite data, data from XOR can actually improve performance on R-XOR, and thus forward transfer is positive.  In contrast, both forward and backward transfer are negative for \RF\ and \DN.

To further investigate this relationship, we design a suite of R-XOR examples, generalizing R-XOR from only $45^\circ$ to any  rotation angle  between $0^\circ$ and $90^\circ$, sampling 100 points from XOR, and another 100 from each R-XOR (Figure~\ref{fig:xor-xnor}Cii). Note that we could not run the experiment for a lot of Monte Carlo repetition to have a smooth curve and hence we have shown a regressed curve fitted to the low repetition noisy curve.
As the angle increases from $0^\circ$ to $45^\circ$, $\mathsf{Backward~Transfer}$ gradually decreases for \PLN. 
%A similar trend is also visible for \PLN.
The $45^\circ$-XOR is the maximally adversarial R-XOR.  Thus, as the angle further increases, $\mathsf{Backward~Transfer}$ increases back up to $\approx 0.18$ at $90^\circ$, which has an identical discriminant boundary to XOR. Moreover, when $\theta$ is fixed at $25^\circ$, $\mathsf{Backward~Transfer}$ increases at different rates for different sample sizes of the source task (Figure~\ref{fig:xor-xnor}Ciii). 

\textit{Together, these experiments  indicate that the amount of transfer can be a complicated function of (i) the difficulty of learning good representations for each task,  (ii) the relationship between the two tasks, and (iii) the sample size of each.} 
%Appendix~\ref{app:sim} further investigates this phenomenon in a multi-spiral environment.

\section{Benchmark data Study}
\label{sec:real}

For benchmark data, we build \PLN~encoders using the network architecture described in \cite{tolias_architecture}. We use the same network architecture for all benchmarking models. For the following experiments, we consider two modalities of real data: vision and language. 

\subsection{Reference algorithms}
\label{sec:reference}
We compared our approaches to $16$ reference lifelong learning methods. Among them five are resource growing as well as modular approach: \ProgNN~\citep{rusu2016progressive}, \sct{DF-CNN}~\citep{Lee2019-eg}, \sct{LMC}\ \citep{ostapenko2021continual}, \zoo \citep{ramesh2021model}, \sct{CoSCL} \citep{wang2022coscl}. 
Note that "\zoo" was published after our work was archived on arXiv, and the authors have built on our work (personal communications). Other reference algorithms are resource constrained: Elastic Weight Consolidation (\sct{EWC}) \citep{kirkpatrick2017overcoming}, Online-EWC (\sct{O-EWC}) \citep{schwarz2018progress}, Synaptic Intelligence (\sct{SI}) \citep{zenke2017continual}, Learning without Forgetting (\sct{LwF}) \citep{li2017learning}, \sct{RanDumb} \citep{prabhu2024randumb} and ``None''.
%These algorithms can be classified into two groups based on whether they add capacity resources per task, or not. 

%Among them, \ProgNN~\citep{rusu2016progressive}, Deconvolution-Factorized CNNs (\sct{DF-CNN})~\citep{Lee2019-eg} and \sct{LMC}\ \citep{ostapenko2021continual} learn new tasks by building new resources. 
%For \ProgNN, for each new task a new ``column'' of network is introduced.  In addition to introducing this column, lateral connections from all previous columns to the new column are added. These lateral connections are computationally costly, as explained in Subsection~\ref{sec:taxonomy}.  
%\sct{DF-CNN}~\citep{Lee2019-eg} is a lifelong learning algorithm that improves upon \ProgNN\ by introducing a knowledge base with lateral connections to each new column, thereby avoiding all pairwise connections, and dramatically reducing computational costs. \sct{LMC}\ improves further upon \ProgNN\ and \sct{DF-CNN}\ by introducing new layers or module only when a sufficient deviation from the previous tasks is detected in the locally decomposed modules.

We also compare two variants of exact replay (Total Replay and Partial Replay) using the code provided in \citep{tolias_architecture}.
% According to the code, b
Both Total and Partial Replay store all the data they have ever seen, but Total Replay replays all of it upon acquiring a new task, whereas Partial Replay replays $N$ samples, randomly sampled from the entire corpus, whenever we acquire a new task with $N$ samples. The above two replay approaches can be considered as two variants of \sct{GDumb} \citep{prabhu2020gdumb}. Additionally, we have compared our approach with more constrained ways of replaying old task data, including Averaged Gradient Episodic Memory (\sct{A-GEM}) \citep{chaudhry2018efficient}, Experience Replay (\sct{ER}) \citep{chaudhry2019tiny} and Task-based Accumulated Gradients (\sct{TAG}) \citep{malviya2021tag}. 
For the baseline ``None'', the network was incrementally trained on all tasks in the standard way while always only using the data from the current task.
The implementations for all of the algorithms are adapted from open source codes~\citep{Lee2019-eg,Van_de_Ven2019-wy};   for implementation details, see Appendix~\ref{app:archs}.

\begin{figure*}[!ht]
    \centering
    \includegraphics[width=.8\linewidth]{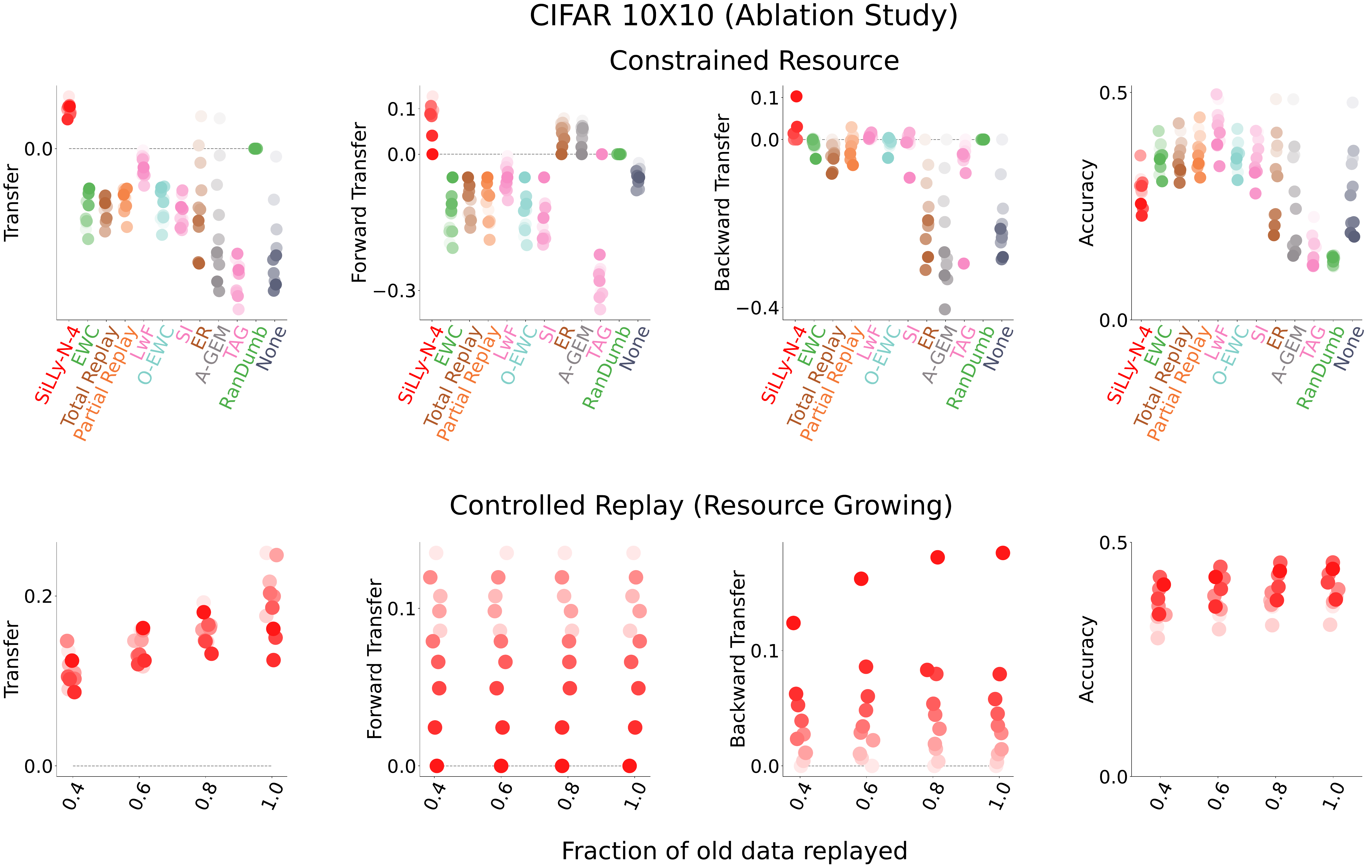}
    \caption{\textbf{Ablation experiments on \PLN\ using CIFAR 10X10.} \textit{Top row:} \PLN\ uniquely shows both positive forward and backward transfer while operating with the same number of parameters as other baseline constant parameter approaches.
    \textit{Bottom row:} Replaying old task data impacts backward transfer while keeping forward transfer unchanged for \PLN.
    }
    \label{fig:ablation}
\end{figure*}

\begin{figure*}[!ht]
    \centering
    \includegraphics[width=.8\linewidth]{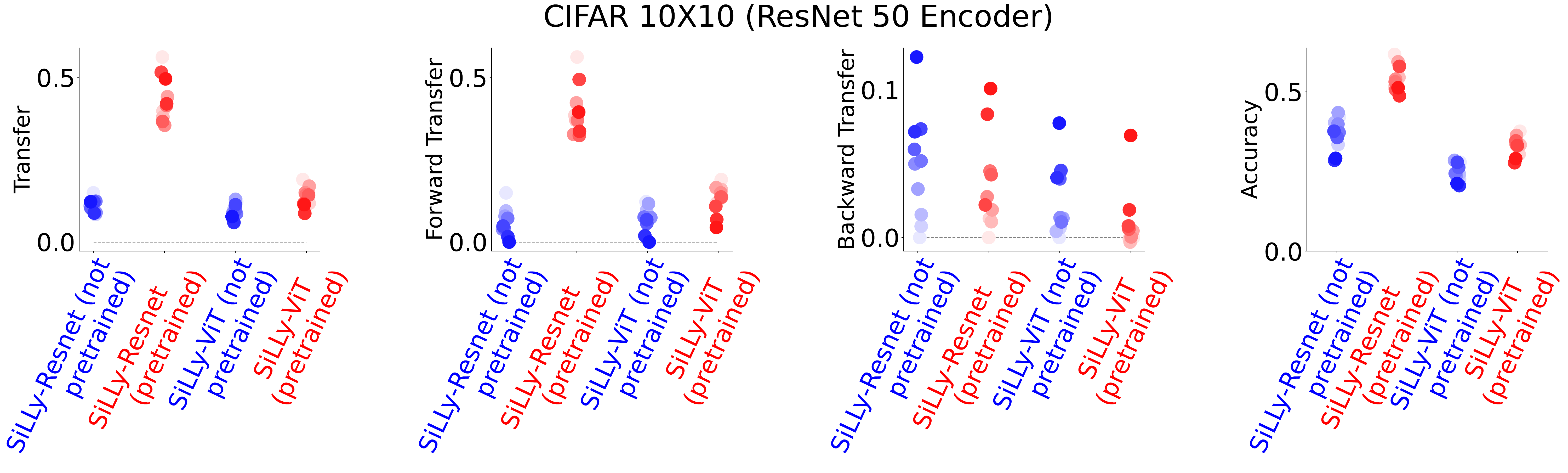}
    \caption{\textbf{Pretrained encoders on CIFAR 10X10.} Using pretrained encoders results in better forward transfer and accuracy for \PLN.}
    \label{fig:pretrain}
\end{figure*}

\subsection{Core benchmarks}

\subsubsection{CIFAR 10X10}\label{sec:cifar10X10}
The CIFAR 100 challenge~\citep{krizhevsky2009learning},  consists  of 50,000 training  and 10,000 test samples, each a 32x32 RGB image of a common object, from one of 100 possible classes, such as apples and bicycles. 
CIFAR 10x10 divides these data into 10 tasks, each with 10 classes~\citep{Lee2019-eg} (see Appendix~\ref{app:cifar} for details). 
% We compare  \PLF~and \PLN\ to the deep lifelong learning algorithms discussed above. 

% We first compare \PLF~and \PLN~to  several resource growing algorithms. . 
\PLN\ and \zoo~demonstrate positive forward and backward transfer for every task in CIFAR 10x10,
in contrast, other algorithms do not exhibit any positive backward transfer (Figure~\ref{fig:strip} first column). 
Moreover, they retained their accuracy while improving transfer (Figure~\ref{fig:strip} first column, first and fourth rows). \ProgNN~had a similar degree of forward transfer, but zero backward transfer, and requires quadratic space and time in sample size, unlike \PLN~which requires quasilinear space and time. 

\subsubsection{Spoken Digit}
In this experiment, we used the \textbf{Spoken Digit} dataset \citep{jackson2018jakobovski}.
% dataset provided in \url{https://github.com/Jakobovski/free-spoken-digit-dataset}. 
  As shown in Figure~\ref{fig:strip} second column, \PLN~shows positive backward and forward transfer between the spoken digit tasks, in contrast to other methods, some of which show only forward transfer, others show only backward transfer, with none showing both and some showing neither. See Appendix~\ref{app:cifar} for details of the experiment.

\subsubsection{FOOD1k 50X20 Dataset}
In this experiment, we use \textbf{Food1k} which is a large scale vision dataset consisting of $1000$ food categories from Food2k \cite{Weiqing-LSVFR-CoRR2021}. FOOD1k 50X20 splits these data into $50$ tasks with $20$ classes each. For each class, we randomly sampled $60$ samples per class for training the models and used rest of the data for testing purpose. Because on the CIFAR experiments \zoo\ performs the best among the reference resource growing models, and \sct{LwF} is the best performing resource constrained algorithm, we only use them  as the reference models for the large scale experiment to avoid heavy computational cost. As shown in Figure~\ref{fig:strip} third column, \PLN~performs the best among all the algorithms on this large dataset.

See Appendix \ref{app:cifar} for experiments with datasets having more samples per task. In lifelong learning, we are often primarily concerned with situations in which we have a small number of samples per task. If we have enough samples per task, the learning agent does not need to transfer knowledge from other tasks. However, below we also experiment with non-trivial lifelong learning setting where sample per task is high.

\textit{These experiments indicate that~\PLN\ has positive transfer in learning environments with various classes and sample sizes.}
% , indicating they are robust to distributional shift in ways that \ProgNN~and \sct{DF-CNN} are not (Figure~\ref{fig:strip} first column; see Appendix Figure~\ref{fig:cifar-10x10_resource_growing} for better view). 
% \PLN, \PLF~and \zoo \citep{ramesh2021model} demonstrate positive backward transfer, indicating that with each new task, performance on all prior tasks increases.  
% In contrast, other algorithms do not exhibit any positive backward transfer. 
% Transfer per task in Figure~\ref{fig:strip} is the overall transfer associated with that task having seen all the data.  \PLF~and \PLN~both demonstrate positive overall learning efficiency for all tasks, whereas other algorithms exhibit negative learning efficiency for at least one task (Figure~\ref{fig:strip} first column third row). 

\subsection{Ablation experiments} 
Our proposed algorithms can improve performance on all the tasks (past and future) by both growing additional resources and replaying data from the past tasks. Below we do two ablation experiments using CIFAR 10X10 to measure the relative contribution of resource growth and replay to the performance of our proposed algorithms.
% \begin{enumerate}
    % \item \textbf

\paragraph{Constrained resource experiment}
In this experiment, we ablate the capability of \PLN\ to grow additional resources after learning $4$ encoders. We also reduce the number of channels and nodes at each encoder layer by four times to keep the total number of parameters similar to the other constant-resource-algorithms. As shown in the top row of Figure~\ref{fig:ablation}, \PLN-4 still shows positive forward and backward transfer with constant resources. However, the accuracy for \PLN-4 gets reduced compared to that of resource growing \PLN\ in Figure~\ref{fig:strip}. Note that all the baseline algorithms have negative backward transfer.
%which implies there is a possibility that their accuracy on older tasks would reduce a lot if there were more tasks in the learning environment. 
This experiment indicates that constant resource mode operation for \PLN\ may be advantageous when we have a lot of tasks to learn and have a decent amount of storage budget available. We will elaborate the above point later with a large scale dataset (food1k).

\paragraph{Controlled replay experiment}
In this experiment, we train four different versions of \PLN\ sequentially on the $10$ tasks from CIFAR 10X10. The only difference between different versions of the algorithms is the amount of old task data replayed. In four different versions of each algorithm, we replay $40\%$, $60\%$, $80\%$ and $100\%$ of the old task data respectively. As apparent from Figure~\ref{fig:ablation} bottom, replaying old task data has no effect on forward transfer, but replaying more data improves backward transfer as the number of tasks increases.

\textit{These experiments  indicate that (i) constraining the resource growth results in lower accuracy while still achieves positive forward and backward transfer for the algorithm,  (ii) lowering the amount of replay lowers backward transfer without any effect on forward transfer.}

\paragraph{Experiment using pretrained encoders}
We explore the effect of using pretrained encoders on the performance of~\PLN. For this experiment only, we use \sct{ResNet 50} and vision transformer \sct{ViT\_B16} (provided in keras-vit package)~\cite{chollet2015keras}. We  freeze all the layers excluding the final two linear layers during training.  \textit{Pretraining the encoders results in better accuracy and forward transfer, but less backward transfer for \PLN\ (Figure~\ref{fig:pretrain}).} Vision transformers achieved lower accuracy, as expected with the small samples sizes used here~\citep{dosovitskiy2020image}. Pretrained network requires  less training epochs for each task (with early stopping).
In this experiment, one forward pass through \sct{ViT}\ takes $\sim 5$ seconds and training a \sct{ViT}\ encoder ($20$ epochs) takes $\sim 4$ minutes using an Apple M1 Max chip and $64$ GB of RAM. 
% However, as it is unclear how to use pretrained encoders for most of the baseline approaches, we do not use pretrained encoders for \PLN\ in other experiments for fair comparison.

\subsection{Adversarial analysis}

\begin{figure*}[!ht]
    \centering
    \includegraphics[width=.85\linewidth]{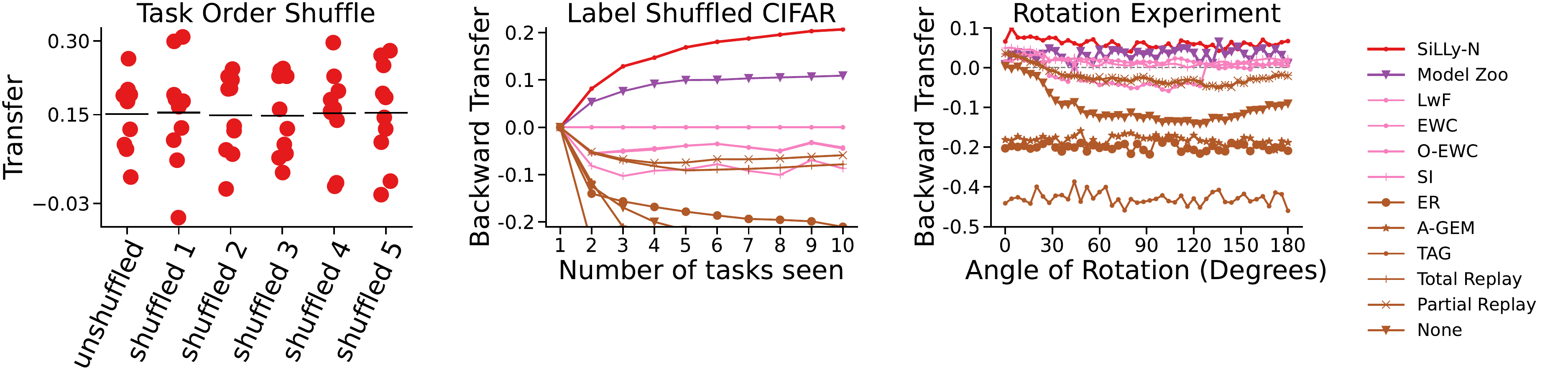}
    \caption{\textbf{Adversarial CIFAR 10x10 experiments.} 
    \textit{Left:} Permuting the task sequenc does not affect $\mathsf{Transfer}$ for \PLN. The mean position, indicated by the black dash, and the spread of the dots on both sides of the mean remain nearly the same.
    \textit{Middle:} Shuffling class labels within tasks two through nine with 500 samples each demonstrates \PLN~can still achieve positive backward transfer, and that the other algorithms still fail to transfer.
    \textit{Right:} \PLN~is nearly invariant to rotations, whereas other approaches are more sensitive to rotation.
    % , some failing to transfer even when the data from a second task is sampled from an identical distribution as the first task (0 angle).
    }
    \label{fig:cifar2}
\end{figure*}
Consider the same CIFAR 10x10 experiment setup in Section \ref{sec:cifar10X10}. In the following experiments, we modify the above setup in different adversarial settings.

\paragraph{Shuffled task order experiment}  We repeat the same experiment mentioned above $5$ times more by permuting the task order. Figure \ref{fig:cifar2} left column shows the mean as well as the spread of $\mathsf{Transfer}$ over all tasks remain the same for different shuffled task orders. Note that the encoders and thus all the channels remain the same regardless of the above task order permutation after all the tasks have been introduced, hence the distribution of $\mathsf{Transfer}$ for all the tasks after all the tasks have been introduced remain similar.

\paragraph{Label shuffle experiment} In this experiment, for Task 2 through 10, randomly permute the class labels within each task, rendering each of those tasks adversarial with respect to the first task (because the labels are uninformative). % 
Figure~\ref{fig:cifar2} middle column indicates that \PLN~show positive backward transfer even with such label shuffling (the other algorithms, except \sct{Model Zoo}, did not demonstrate positive backward transfer).
\paragraph{Rotation experiment} Consider  a Rotated CIFAR experiment, which uses only data from the first task, divided into two subsets of equal size (making two tasks), where the second subset is rotated by different amounts (Figure~\ref{fig:cifar2} right column). 
Backward transfer of \PLN~is nearly invariant to rotation angle, whereas the other approaches are much more sensitive to the rotation angle.
Note that the zero rotation angle corresponds to the two tasks \textit{having identical distributions}. 
The fact that other algorithms fail to transfer even in this setting suggests that they may never be able to achieve a positive backward transfer. See Appendix~\ref{app:cifar} for an additional experiment using CIFAR 10X10.

\textit{These adversarial experiments  indicate that~\PLN\ is robust to adversarial perturbations in the source tasks, while most of the other algorithms are not.}

% \begin{figure*}[!ht]
%     \centering
%     \includegraphics[width=.7\linewidth]{images/combined_all.pdf}
%     \caption{\textbf{Performance summary on vision and speech benchmark datasets with varying sample sizes per task.} Dataset on the right has more samples per task. Imagenet and five dataset have really high sample size per task which is a non-trivial lifelong learning setting. \PLN\ performs the best in the low sample size regime on the left two columns which is desirable in lifelong learning. See appendix Figure~\ref{fig:vision} and~\ref{fig:language} for the extended results.
%     % show better performance than \sct{Model Zoo} on CIFAR 10X10 (column 1), Speech (column 2), FOOD1k (column 3) which have relatively smaller sample size per task compared to those of Imagenet(column 4) and Five Dataset (column 5).
%     }
%     \label{fig:all} 
% \end{figure*}

% \subsection{Further experiments on additional datasets with more classes, tasks, and/or samples}

\subsection{Constant Resource Mode Operation}
\begin{figure*}[!ht]
    \centering
    \includegraphics[width=.7\linewidth]{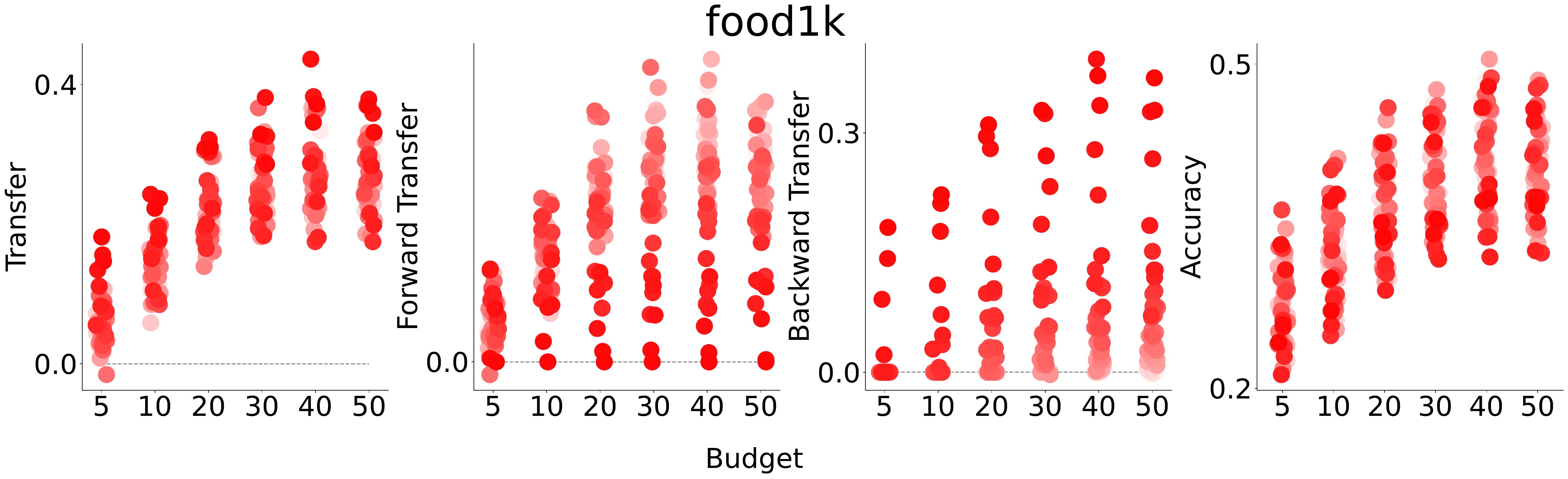}
    \caption{\textbf{\PLN\ in constant resource mode operation.} Improvement in Transfer between tasks becomes negligible at nearly $30$ encoders. The user can choose to operate in lower budget with less transfer. 
    }
    \label{fig:budget}
\end{figure*}

\subsubsection{FOOD1k 50X20}
The binary distinction we made above, algorithms either build resources or reallocate them, is a false dichotomy, and biologically unnatural. In biological learning, systems develop from building to fixed resources, as they grow from juveniles to adults.  To explore this continuum of amount of resources to grow, we experiment on FOOD1k 50X20 dataset using the constant resource mode operation of \PLN\ as described in Section~\ref{sec:algorithms}. We evaluate the performance of \PLN\ for different number of encoder budget. Performance of \PLN\ saturates after $30$ encoders, though with only $5$ encoders, still demonstrates forward and backward transfer (Figure~\ref{fig:budget}). 
% However, one can choose to operate under less storage and computation budget while sacrificing a certain amount of performance.

% \begin{figure}[!ht]
%     \centering
%     \includegraphics[width=.5\linewidth]{images/core50.pdf}
%     \caption{\textcolor{red}{\textbf{Performance of \PLN\ on CORe50 dataset ($110$ tasks in total) with a budget of $20$ encoders.} \PLN\ achieves high $\mathsf{Transfer}$ in a large scale dataset with accuracies close to $100\%$.
%     }}
%     \label{fig:core50}
% \end{figure}

\subsubsection{CORe50 110X5}
To further evaluate the effectiveness of constant resource mode operation, we utilize a large-scale dataset: CORe50 \citep{lomonaco2017core50}, partitioned into $110$ tasks with $5$ classes each and $100$ training samples per class. To simulate a resource-constrained environment, we conducted all experiments on an Apple M1 Max chip with 64 GB of RAM. We set a budget of 20 encoders, as this configuration maintains high accuracy (above $90\%$) across all tasks. The fourth row of Figure \ref{fig:strip} shows that \PLN\ achieves higher $\mathsf{Transfer}$ operating in constant resource mode on a large dataset, that is, CORe50 compared to the smaller datasets discussed above. In particular, $\mathsf{Backward Transfer}$ drops to zero after 20 tasks due to the encoder limit, leaving only $\mathsf{Forward Transfer}$ for subsequent tasks. Moreover, we cannot run resource-growing algorithms like \zoo\ in this resource-constrained environment. The constant resource algorithm, such as \sct{LwF}, completely forgets the first several tasks, and the corresponding task accuracies go as low as the chance level, $20\%$.

\subsection{\PLF\ on tabular data}

\begin{figure*}[!htbp]
    \centering
    \includegraphics[width=.8\linewidth]{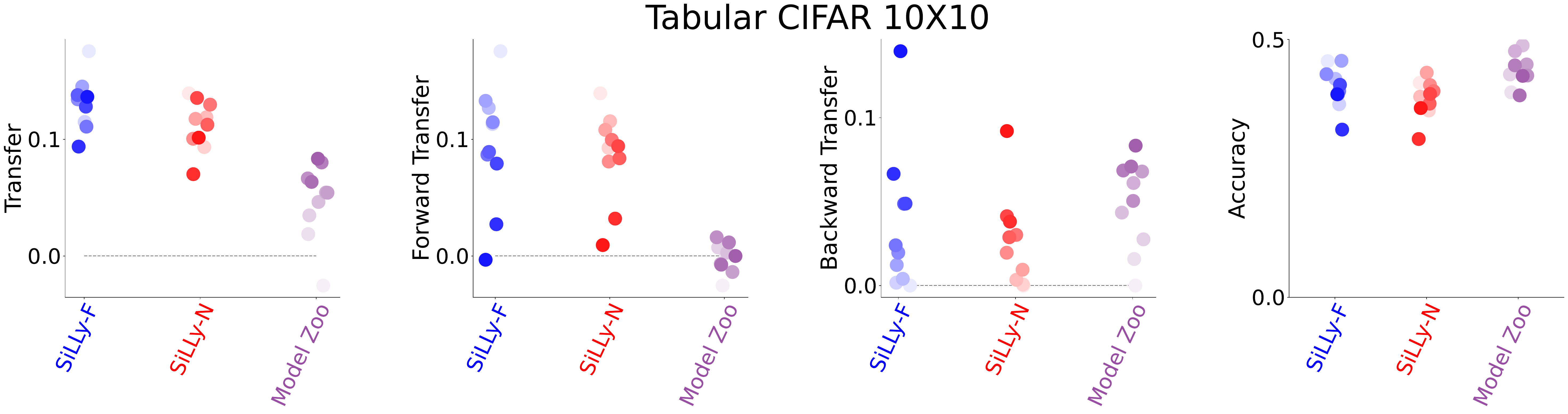}
    \caption{\textbf{Our proposed approach can be used with random forest as encoders on tabular data (\PLF)}. \PLF\ shows more positive forward and backward transfer while operating with less parameters compared to other baseline approaches on tabular data.
    }
    \label{fig:tabular}
\end{figure*}

In this experiment, we experiment with \PLF, an additional realization of our approach using random forests as encoders (described in Section~\ref{sec:algorithms}). We flatten the CIFAR 10X10 data and use them as tabular data. We train two other best performing baseline algorithms, \PLN\ and \zoo\ and use three fully connected hidden layers, each having $2000$ nodes, as encoders. As shown in Figure~\ref{fig:tabular}, \PLF\ performs the best among all the approaches. \textit{This experiment shows our approach can be used as a general structure to do lifelong learning using other machine learning models as encoder.}
\section{Discussion}
\label{sec:discussion}

We introduced representation ensembling  as a simple approach for lifelong learning.  Two specific algorithms, \PLN\ and \PLF, achieve both forward and backward transfer, by leveraging resources learned for other tasks without undue computational burdens. Our work is well suited for federated learning scenarios, where each data center independently trains a model on its private data and shares only the encoders with other centers \citep{dwork2008differential}. In this paper, we have mainly focused on task-aware setting, because it is simpler. Future work will extend our approach to more challenging task-unaware settings \textcolor{red}{\citep{caccia2020online}}. Our code, including the code for reproducing the experiments in this manuscript, is available from \url{http://proglearn.neurodata.io/}.
\section*{Acknowledgments}
The authors thank the support of the NSF-Simons Research Collaborations on the Mathematical and Scientific Foundations of Deep Learning (NSF grant 2031985). We also thank Raman Arora, Dinesh Jayaraman, Rene Vidal, Jeremias Sulam, Guillermo Sapiro, Michael Powell, and Weiwei Yang for helpful discussions. This work is graciously supported by the Defense Advanced Research Projects Agency (DARPA) Lifelong Learning Machines program through contracts FA8650-18-2-7834 and HR0011-18-2-0025. Research was partially supported by funding from Microsoft Research and the Kavli Neuroscience Discovery Institute.

\section*{Co-first Author Contributions}
J.D.: Conceptualization, Methodology, Software, Data Curation, Formal Analysis, Experimentation, Theoretical proof (see Appendix D), Writing $\&$ Editing; 
J.T.V.: Conceptualization, Methodology, Supervision, Writing—Review $\&$ Editing, and Funding Acquisition.

\bibliographystyle{IEEEtran}
\bibliography{biblifelonglearning.bib}

\begin{thebibliography}{10}
\providecommand{\url}[1]{#1}
\csname url@samestyle\endcsname
\providecommand{\newblock}{\relax}
\providecommand{\bibinfo}[2]{#2}
\providecommand{\BIBentrySTDinterwordspacing}{\spaceskip=0pt\relax}
\providecommand{\BIBentryALTinterwordstretchfactor}{4}
\providecommand{\BIBentryALTinterwordspacing}{\spaceskip=\fontdimen2\font plus
\BIBentryALTinterwordstretchfactor\fontdimen3\font minus \fontdimen4\font\relax}
\providecommand{\BIBforeignlanguage}[2]{{%
\expandafter\ifx\csname l@#1\endcsname\relax
\typeout{** WARNING: IEEEtran.bst: No hyphenation pattern has been}%
\typeout{** loaded for the language `#1'. Using the pattern for}%
\typeout{** the default language instead.}%
\else
\language=\csname l@#1\endcsname
\fi
#2}}
\providecommand{\BIBdecl}{\relax}
\BIBdecl

\bibitem{mitchell1999machine}
T.~M. Mitchell, ``Machine learning and data mining,'' \emph{Communications of the ACM}, vol.~42, no.~11, pp. 30--36, 1999.

\bibitem{Vapnik1971-um}
V.~Vapnik and A.~Chervonenkis, ``{On the Uniform Convergence of Relative Frequencies of Events to Their Probabilities},'' \emph{Theory Probab. Appl.}, vol.~16, no.~2, pp. 264--280, Jan. 1971.

\bibitem{Valiant1984-dx}
\BIBentryALTinterwordspacing
L.~G. Valiant, ``{A Theory of the Learnable},'' \emph{Commun. ACM}, vol.~27, no.~11, pp. 1134--1142, Nov. 1984. [Online]. Available: \url{http://doi.acm.org/10.1145/1968.1972}
\BIBentrySTDinterwordspacing

\bibitem{caruana1997multitask}
R.~Caruana, ``Multitask learning,'' \emph{Machine learning}, vol.~28, no.~1, pp. 41--75, 1997.

\bibitem{thrun1996learning}
S.~Thrun, ``Is learning the n-th thing any easier than learning the first?'' in \emph{Advances in neural information processing systems}, 1996, pp. 640--646.

\bibitem{Thrun2012-sj}
\BIBentryALTinterwordspacing
S.~Thrun and L.~Pratt, \emph{\BIBforeignlanguage{en}{{Learning to Learn}}}.\hskip 1em plus 0.5em minus 0.4em\relax Springer Science \& Business Media, Dec. 2012. [Online]. Available: \url{https://market.android.com/details?id=book-X_jpBwAAQBAJ}
\BIBentrySTDinterwordspacing

\bibitem{mccloskey1989catastrophic}
M.~McCloskey and N.~J. Cohen, ``Catastrophic interference in connectionist networks: The sequential learning problem,'' in \emph{Psychology of learning and motivation}.\hskip 1em plus 0.5em minus 0.4em\relax Elsevier, 1989, vol.~24, pp. 109--165.

\bibitem{mcclelland1995there}
J.~L. McClelland, B.~L. McNaughton, and R.~C. O'Reilly, ``Why there are complementary learning systems in the hippocampus and neocortex: insights from the successes and failures of connectionist models of learning and memory.'' \emph{Psychological review}, vol. 102, no.~3, p. 419, 1995.

\bibitem{doan2021theoretical}
T.~Doan, M.~A. Bennani, B.~Mazoure, G.~Rabusseau, and P.~Alquier, ``A theoretical analysis of catastrophic forgetting through the ntk overlap matrix,'' in \emph{International Conference on Artificial Intelligence and Statistics}.\hskip 1em plus 0.5em minus 0.4em\relax PMLR, 2021, pp. 1072--1080.

\bibitem{Zhao2016-vo}
J.~Zhao, B.~Quiroz, L.~Q. Dixon, and R.~M. Joshi, ``\BIBforeignlanguage{en}{{Comparing Bilingual to Monolingual Learners on English Spelling: A Meta-analytic Review}},'' \emph{\BIBforeignlanguage{en}{Dyslexia}}, vol.~22, no.~3, pp. 193--213, Aug. 2016.

\bibitem{kirkpatrick2017overcoming}
J.~Kirkpatrick, R.~Pascanu, N.~Rabinowitz, J.~Veness, G.~Desjardins, A.~A. Rusu, K.~Milan, J.~Quan, T.~Ramalho, A.~Grabska-Barwinska, D.~Hassabis, C.~Clopath, D.~Kumaran, and R.~Hadsell, ``Overcoming catastrophic forgetting in neural networks,'' \emph{Proceedings of the national academy of sciences}, vol. 114, no.~13, pp. 3521--3526, 2017.

\bibitem{schwarz2018progress}
J.~Schwarz, J.~Luketina, W.~M. Czarnecki, A.~Grabska-Barwinska, Y.~W. Teh, R.~Pascanu, and R.~Hadsell, ``Progress \& compress: A scalable framework for continual learning,'' \emph{arXiv preprint arXiv:1805.06370}, 2018.

\bibitem{chaudhry2018riemannian}
A.~Chaudhry, P.~K. Dokania, T.~Ajanthan, and P.~H. Torr, ``Riemannian walk for incremental learning: Understanding forgetting and intransigence,'' in \emph{Proceedings of the European conference on computer vision (ECCV)}, 2018, pp. 532--547.

\bibitem{zenke2017continual}
F.~Zenke, B.~Poole, and S.~Ganguli, ``Continual learning through synaptic intelligence,'' in \emph{Proceedings of the 34th International Conference on Machine Learning-Volume 70}.\hskip 1em plus 0.5em minus 0.4em\relax JMLR. org, 2017, pp. 3987--3995.

\bibitem{li2017learning}
Z.~Li and D.~Hoiem, ``Learning without forgetting,'' \emph{IEEE transactions on pattern analysis and machine intelligence}, vol.~40, no.~12, pp. 2935--2947, 2017.

\bibitem{wang2003mining}
H.~Wang, W.~Fan, P.~S. Yu, and J.~Han, ``Mining concept-drifting data streams using ensemble classifiers,'' in \emph{Proceedings of the ninth ACM SIGKDD international conference on Knowledge discovery and data mining}, 2003, pp. 226--235.

\bibitem{marouf2024weighted}
I.~E. Marouf, S.~Roy, E.~Tartaglione, and S.~Lathuili{\`e}re, ``Weighted ensemble models are strong continual learners,'' in \emph{European Conference on Computer Vision}.\hskip 1em plus 0.5em minus 0.4em\relax Springer, 2024, pp. 306--324.

\bibitem{dai2007boosting}
W.~Dai, Q.~Yang, G.-R. Xue, and Y.~Yu, ``Boosting for transfer learning.(2007), 193--200,'' in \emph{Proceedings of the 24th international conference on Machine learning}, 2007.

\bibitem{polikar2001learn++}
R.~Polikar, L.~Upda, S.~S. Upda, and V.~Honavar, ``Learn++: An incremental learning algorithm for supervised neural networks,'' \emph{IEEE transactions on systems, man, and cybernetics, part C (applications and reviews)}, vol.~31, no.~4, pp. 497--508, 2001.

\bibitem{van2022three}
G.~M. van~de Ven, T.~Tuytelaars, and A.~S. Tolias, ``Three types of incremental learning,'' \emph{Nature Machine Intelligence}, pp. 1--13, 2022.

\bibitem{ramesh2021model}
R.~Ramesh and P.~Chaudhari, ``Model zoo: A growing brain that learns continually,'' in \emph{International Conference on Learning Representations}, 2021.

\bibitem{rusu2016progressive}
A.~A. Rusu, N.~C. Rabinowitz, G.~Desjardins, H.~Soyer, J.~Kirkpatrick, K.~Kavukcuoglu, R.~Pascanu, and R.~Hadsell, ``Progressive neural networks,'' \emph{arXiv preprint arXiv:1606.04671}, 2016.

\bibitem{Lee2019-eg}
S.~Lee, J.~Stokes, and E.~Eaton, ``Learning shared knowledge for deep lifelong learning using deconvolutional networks,'' in \emph{Proceedings of the 28th International Joint Conference on Artificial Intelligence}, 2019, pp. 2837--2844.

\bibitem{mallya2018packnet}
A.~Mallya and S.~Lazebnik, ``Packnet: Adding multiple tasks to a single network by iterative pruning,'' in \emph{Proceedings of the IEEE conference on Computer Vision and Pattern Recognition}, 2018, pp. 7765--7773.

\bibitem{veniat2020efficient}
T.~Veniat, L.~Denoyer, and M.~Ranzato, ``Efficient continual learning with modular networks and task-driven priors,'' \emph{arXiv preprint arXiv:2012.12631}, 2020.

\bibitem{ostapenko2021continual}
O.~Ostapenko, P.~Rodriguez, M.~Caccia, and L.~Charlin, ``Continual learning via local module composition,'' \emph{Advances in Neural Information Processing Systems}, vol.~34, pp. 30\,298--30\,312, 2021.

\bibitem{mehta2021continual}
N.~Mehta, K.~Liang, V.~K. Verma, and L.~Carin, ``Continual learning using a bayesian nonparametric dictionary of weight factors,'' in \emph{International Conference on Artificial Intelligence and Statistics}.\hskip 1em plus 0.5em minus 0.4em\relax PMLR, 2021, pp. 100--108.

\bibitem{yan2021dynamically}
S.~Yan, J.~Xie, and X.~He, ``Der: Dynamically expandable representation for class incremental learning,'' in \emph{Proceedings of the IEEE/CVF Conference on Computer Vision and Pattern Recognition}, 2021, pp. 3014--3023.

\bibitem{kang2022forget}
H.~Kang, R.~J.~L. Mina, S.~R.~H. Madjid, J.~Yoon, M.~Hasegawa-Johnson, S.~J. Hwang, and C.~D. Yoo, ``Forget-free continual learning with winning subnetworks,'' in \emph{International Conference on Machine Learning}.\hskip 1em plus 0.5em minus 0.4em\relax PMLR, 2022, pp. 10\,734--10\,750.

\bibitem{frankle2018lottery}
J.~Frankle and M.~Carbin, ``The lottery ticket hypothesis: Finding sparse, trainable neural networks,'' \emph{arXiv preprint arXiv:1803.03635}, 2018.

\bibitem{tolias_architecture}
G.~M. van~de Ven, H.~T. Siegelmann, and A.~S. Tolias, ``Brain-inspired replay for continual learning with artificial neural networks,'' \emph{Nature communications}, vol.~11, p. 4069, 2020.

\bibitem{robins1995catastrophic}
A.~Robins, ``Catastrophic forgetting, rehearsal and pseudorehearsal,'' \emph{Connection Science}, vol.~7, no.~2, pp. 123--146, 1995.

\bibitem{shin2017continual}
H.~Shin, J.~K. Lee, J.~Kim, and J.~Kim, ``Continual learning with deep generative replay,'' in \emph{Advances in Neural Information Processing Systems}, 2017, pp. 2990--2999.

\bibitem{chaudhry2019tiny}
A.~Chaudhry, M.~Rohrbach, M.~Elhoseiny, T.~Ajanthan, P.~K. Dokania, P.~H. Torr, and M.~Ranzato, ``On tiny episodic memories in continual learning,'' \emph{arXiv preprint arXiv:1902.10486}, 2019.

\bibitem{buzzega2020dark}
P.~Buzzega, M.~Boschini, A.~Porrello, D.~Abati, and S.~Calderara, ``Dark experience for general continual learning: a strong, simple baseline,'' \emph{Advances in neural information processing systems}, vol.~33, pp. 15\,920--15\,930, 2020.

\bibitem{channappayya2023augmented}
S.~Channappayya, B.~R. Tamma \emph{et~al.}, ``Augmented memory replay-based continual learning approaches for network intrusion detection,'' \emph{Advances in Neural Information Processing Systems}, vol.~36, pp. 17\,156--17\,169, 2023.

\bibitem{chen2021human}
C.~Chen, K.~Ota, M.~Dong, C.~Yu, and H.~Jin, ``Human activity recognition-oriented incremental learning with knowledge distillation,'' \emph{Journal of Circuits, Systems and Computers}, vol.~30, no.~06, p. 2150096, 2021.

\bibitem{von2019continual}
J.~Von~Oswald, C.~Henning, B.~F. Grewe, and J.~Sacramento, ``Continual learning with hypernetworks,'' \emph{arXiv preprint arXiv:1906.00695}, 2019.

\bibitem{riemer2018learning}
M.~Riemer, I.~Cases, R.~Ajemian, M.~Liu, I.~Rish, Y.~Tu, and G.~Tesauro, ``Learning to learn without forgetting by maximizing transfer and minimizing interference,'' \emph{arXiv preprint arXiv:1810.11910}, 2018.

\bibitem{dey2021towards}
J.~Dey, A.~Geisa, R.~Mehta, T.~M. Tomita, H.~S. Helm, H.~Xu, E.~Eaton, J.~Dick, C.~E. Priebe, and J.~T. Vogelstein, ``Towards a theory of out-of-distribution learning,'' \emph{arXiv preprint arXiv:2109.14501}, 2021.

\bibitem{LopezPaz2017GradientEM}
D.~Lopez-Paz and M.~Ranzato, ``Gradient episodic memory for continual learning,'' in \emph{NIPS}, 2017.

\bibitem{Benavides-Prado2018-nv}
D.~Benavides-Prado, Y.~S. Koh, and P.~Riddle, ``{Measuring Cumulative Gain of Knowledgeable Lifelong Learners},'' in \emph{{NeurIPS Continual Learning Workshop}}, 2018, pp. 1--8.

\bibitem{diaz2018don}
N.~D{\'\i}az-Rodr{\'\i}guez, V.~Lomonaco, D.~Filliat, and D.~Maltoni, ``Don't forget, there is more than forgetting: new metrics for continual learning,'' \emph{arXiv preprint arXiv:1810.13166}, 2018.

\bibitem{judea2018gained}
P.~Judea, ``What is gained from past learning,'' \emph{Journal of Causal Inference}, vol.~6, no.~1, 2018.

\bibitem{bickel2015mathematical}
P.~J. Bickel and K.~A. Doksum, \emph{Mathematical statistics: basic ideas and selected topics, volumes I-II package}.\hskip 1em plus 0.5em minus 0.4em\relax Chapman and Hall/CRC, 2015.

\bibitem{chakraborty2022efficient}
S.~Chakraborty, B.~Uzkent, K.~Ayush, K.~Tanmay, E.~Sheehan, and S.~Ermon, ``Efficient conditional pre-training for transfer learning,'' in \emph{Proceedings of the IEEE/CVF Conference on Computer Vision and Pattern Recognition}, 2022, pp. 4241--4250.

\bibitem{de2021continual}
M.~De~Lange, R.~Aljundi, M.~Masana, S.~Parisot, X.~Jia, A.~Leonardis, G.~Slabaugh, and T.~Tuytelaars, ``A continual learning survey: Defying forgetting in classification tasks,'' \emph{IEEE transactions on pattern analysis and machine intelligence}, vol.~44, no.~7, pp. 3366--3385, 2021.

\bibitem{yuan2023survey}
B.~Yuan and D.~Zhao, ``A survey on continual semantic segmentation: Theory,'' \emph{Challenge, Method and Application}, 2023.

\bibitem{kuo2003optimal}
W.~Kuo and M.~J. Zuo, \emph{Optimal reliability modeling: principles and applications}.\hskip 1em plus 0.5em minus 0.4em\relax John Wiley \& Sons, 2003.

\bibitem{zhang2021understanding}
C.~Zhang, S.~Bengio, M.~Hardt, B.~Recht, and O.~Vinyals, ``Understanding deep learning (still) requires rethinking generalization,'' \emph{Communications of the ACM}, vol.~64, no.~3, pp. 107--115, 2021.

\bibitem{Cover2012-sl}
T.~M. Cover and J.~A. Thomas, \emph{\BIBforeignlanguage{en}{{Elements of Information Theory}}}.\hskip 1em plus 0.5em minus 0.4em\relax New York: John Wiley \& Sons, Nov. 2012.

\bibitem{Cho2014-ew}
K.~Cho, B.~{van Merrienboer}, C.~Gulcehre, F.~Bougares, H.~Schwenk, and Y.~Bengio, ``\BIBforeignlanguage{English (US)}{Learning phrase representations using rnn encoder-decoder for statistical machine translation},'' in \emph{\BIBforeignlanguage{English (US)}{Conference on Empirical Methods in Natural Language Processing (EMNLP 2014)}}, 2014.

\bibitem{Vaswani2017-lq}
A.~Vaswani, N.~Shazeer, N.~Parmar, J.~Uszkoreit, L.~Jones, A.~N. Gomez, {\L}.~U. Kaiser, and I.~Polosukhin, ``{Attention is All you Need},'' in \emph{{Advances in Neural Information Processing Systems 30}}, I.~Guyon, U.~V. Luxburg, S.~Bengio, H.~Wallach, R.~Fergus, S.~Vishwanathan, and R.~Garnett, Eds.\hskip 1em plus 0.5em minus 0.4em\relax Curran Associates, Inc., 2017, pp. 5998--6008.

\bibitem{Devlin2018-lk}
\BIBentryALTinterwordspacing
J.~Devlin, M.~Chang, K.~Lee, and K.~Toutanova, ``{BERT:} pre-training of deep bidirectional transformers for language understanding,'' \emph{CoRR}, vol. abs/1810.04805, 2018. [Online]. Available: \url{http://arxiv.org/abs/1810.04805}
\BIBentrySTDinterwordspacing

\bibitem{breiman1984classification}
L.~Breiman, J.~Friedman, C.~J. Stone, and R.~A. Olshen, \emph{Classification and regression trees}.\hskip 1em plus 0.5em minus 0.4em\relax CRC press, 1984.

\bibitem{denil14}
M.~Denil, D.~Matheson, and N.~D. Freitas, ``Narrowing the gap: Random forests in theory and in practice,'' in \emph{Proceedings of the 31st International Conference on Machine Learning}, ser. Proceedings of Machine Learning Research, E.~P. Xing and T.~Jebara, Eds., vol.~32, 6 2014, pp. 665--673.

\bibitem{Athey19}
S.~Athey, J.~Tibshirani, and S.~Wager, ``Generalized random forests,'' \emph{Annals of Statistics}, vol.~47, no.~2, pp. 1148--1178, 2019.

\bibitem{Breiman1996-yz}
L.~Breiman, ``{Bagging predictors},'' \emph{Mach. Learn.}, vol.~24, no.~2, pp. 123--140, Aug. 1996.

\bibitem{Freund1995-md}
Y.~Freund, ``{Boosting a Weak Learning Algorithm by Majority},'' \emph{Inform. and Comput.}, vol. 121, no.~2, pp. 256--285, Sep. 1995.

\bibitem{breiman2001random}
L.~Breiman, ``Random forests,'' \emph{Machine learning}, vol.~45, no.~1, pp. 5--32, 2001.

\bibitem{Amit1997-nd}
Y.~Amit and D.~Geman, ``{Shape Quantization and Recognition with Randomized Trees},'' \emph{Neural Comput.}, vol.~9, no.~7, pp. 1545--1588, Oct. 1997.

\bibitem{alijundi2018memory}
R.~Aljundi, F.~Babiloni, M.~Elhoseiny, M.~Rohrbach, and T.~Tuytelaars, ``Memory aware synapses: Learning what (not) to forget,'' in \emph{Computer Vision -- ECCV 2018}, V.~Ferrari, M.~Hebert, C.~Sminchisescu, and Y.~Weiss, Eds.\hskip 1em plus 0.5em minus 0.4em\relax Cham: Springer International Publishing, 2018, pp. 144--161.

\bibitem{wang2022coscl}
L.~Wang, X.~Zhang, Q.~Li, J.~Zhu, and Y.~Zhong, ``Coscl: Cooperation of small continual learners is stronger than a big one,'' in \emph{European Conference on Computer Vision}.\hskip 1em plus 0.5em minus 0.4em\relax Springer, 2022, pp. 254--271.

\bibitem{prabhu2024randumb}
A.~Prabhu, S.~Sinha, P.~Kumaraguru, P.~Torr, O.~Sener, and P.~Dokania, ``Randumb: Random representations outperform online continually learned representations,'' \emph{Advances in Neural Information Processing Systems}, vol.~37, pp. 37\,988--38\,006, 2024.

\bibitem{prabhu2020gdumb}
A.~Prabhu, P.~H. Torr, and P.~K. Dokania, ``Gdumb: A simple approach that questions our progress in continual learning,'' in \emph{Computer Vision--ECCV 2020: 16th European Conference, Glasgow, UK, August 23--28, 2020, Proceedings, Part II 16}.\hskip 1em plus 0.5em minus 0.4em\relax Springer, 2020, pp. 524--540.

\bibitem{chaudhry2018efficient}
A.~Chaudhry, M.~Ranzato, M.~Rohrbach, and M.~Elhoseiny, ``Efficient lifelong learning with a-gem,'' \emph{arXiv preprint arXiv:1812.00420}, 2018.

\bibitem{malviya2021tag}
P.~Malviya, S.~Chandar, and B.~Ravindran, ``Tag: Task-based accumulated gradients for lifelong learning,'' \emph{arXiv preprint arXiv:2105.05155}, 2021.

\bibitem{Van_de_Ven2019-wy}
\BIBentryALTinterwordspacing
G.~M. van~de Ven and A.~S. Tolias, ``Three scenarios for continual learning,'' \emph{CoRR}, vol. abs/1904.07734, 2019. [Online]. Available: \url{http://arxiv.org/abs/1904.07734}
\BIBentrySTDinterwordspacing

\bibitem{krizhevsky2009learning}
A.~Krizhevsky, ``Learning multiple layers of features from tiny images,'' \emph{University of Toronto}, 05 2012.

\bibitem{jackson2018jakobovski}
Z.~Jackson, C.~Souza, J.~Flaks, Y.~Pan, H.~Nicolas, and A.~Thite, ``Jakobovski/free-spoken-digit-dataset: v1. 0.8,'' \emph{Zenodo, August}, 2018.

\bibitem{Weiqing-LSVFR-CoRR2021}
W.~Min, Z.~Wang, Y.~Liu, M.~Luo, L.~Kang, X.~Wei, X.~Wei, and S.~Jiang, ``Large scale visual food recognition,'' \emph{CoRR}, vol. abs/2103.16107, 2021.

\bibitem{chollet2015keras}
F.~Chollet \emph{et~al.}, ``Keras,'' \url{https://github.com/fchollet/keras}, 2015.

\bibitem{dosovitskiy2020image}
A.~Dosovitskiy, ``An image is worth 16x16 words: Transformers for image recognition at scale,'' \emph{arXiv preprint arXiv:2010.11929}, 2020.

\bibitem{lomonaco2017core50}
V.~Lomonaco and D.~Maltoni, ``Core50: a new dataset and benchmark for continuous object recognition,'' in \emph{Conference on Robot Learning}.\hskip 1em plus 0.5em minus 0.4em\relax PMLR, 2017, pp. 17--26.

\bibitem{caccia2020online}
M.~Caccia, P.~Rodriguez, O.~Ostapenko, F.~Normandin, M.~Lin, L.~Page-Caccia, I.~H. Laradji, I.~Rish, A.~Lacoste, D.~V{\'a}zquez \emph{et~al.}, ``Online fast adaptation and knowledge accumulation (osaka): a new approach to continual learning,'' \emph{Advances in Neural Information Processing Systems}, vol.~33, pp. 16\,532--16\,545, 2020.

\bibitem{dwork2008differential}
C.~Dwork, ``Differential privacy: A survey of results,'' in \emph{International conference on theory and applications of models of computation}.\hskip 1em plus 0.5em minus 0.4em\relax Springer, 2008, pp. 1--19.

\bibitem{Van_Rooij2019-hu}
I.~van Rooij, M.~Blokpoel, J.~Kwisthout, and T.~Wareham, \emph{\BIBforeignlanguage{en}{{Cognition and Intractability: A Guide to Classical and Parameterized Complexity Analysis}}}.\hskip 1em plus 0.5em minus 0.4em\relax Cambridge University Press, Apr. 2019.

\bibitem{netzer2011reading}
Y.~Netzer, T.~Wang, A.~Coates, A.~Bissacco, B.~Wu, and A.~Y. Ng, ``Reading digits in natural images with unsupervised feature learning,'' 2011.

\bibitem{notMNIST}
Y.~Bulatov, ``http://yaroslavvb.blogspot.com/2011/09/notmnist-dataset.html,'' 2011.

\bibitem{xiao2017fashion}
H.~Xiao, K.~Rasul, and R.~Vollgraf, ``Fashion-mnist: a novel image dataset for benchmarking machine learning algorithms,'' \emph{arXiv preprint arXiv:1708.07747}, 2017.

\bibitem{wyner2017explaining}
A.~J. Wyner, M.~Olson, J.~Bleich, and D.~Mease, ``Explaining the success of adaboost and random forests as interpolating classifiers,'' \emph{The Journal of Machine Learning Research}, vol.~18, no.~1, pp. 1558--1590, 2017.

\bibitem{Caruana2006-wp}
R.~Caruana and A.~Niculescu-Mizil, ``{An Empirical Comparison of Supervised Learning Algorithms},'' in \emph{{Proceedings of the 23rd International Conference on Machine Learning}}, ser. ICML '06.\hskip 1em plus 0.5em minus 0.4em\relax New York, NY, USA: ACM, 2006, pp. 161--168.

\bibitem{caruana2004ensemble}
R.~Caruana, A.~Niculescu-Mizil, G.~Crew, and A.~Ksikes, ``Ensemble selection from libraries of models,'' in \emph{Proceedings of the twenty-first international conference on Machine learning}, 2004, p.~18.

\end{thebibliography}

%\newpage

\section{Biography Section}
\begin{IEEEbiography}[{\includegraphics[width=1in,height=1.25in,clip,keepaspectratio]{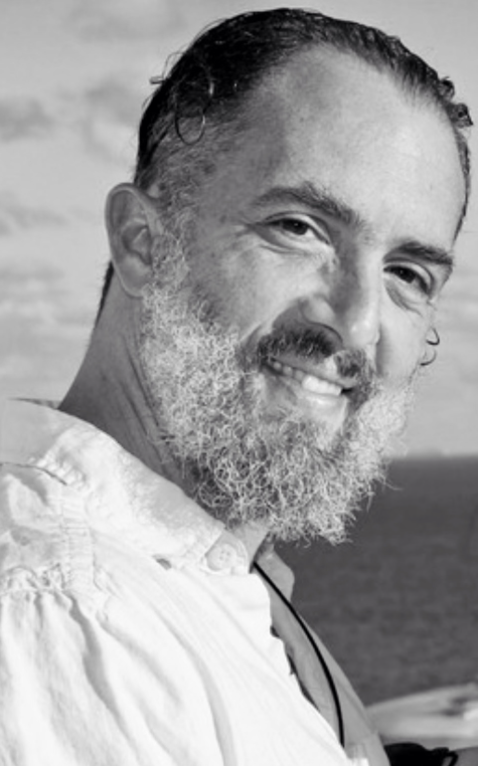}}]{Joshua T. Vogelstein}
is an Associate Professor in the Department of Biomedical Engineering at Johns Hopkins University, with joint appointments in Applied Mathematics and Statistics, Computer Science, Electrical and Computer Engineering, Neuroscience, and Biostatistics. His research focuses primarily on the intersection of natural and artificial intelligence. His lab develops and applies high-dimensional nonlinear machine learning methods to biomedical big data science challenges. He has published about 200 papers in prominent scientific and engineering venues, with $>12,000$ citations and an h-index $>45$. His group is one of the few in the world that regularly publishes in both top scientific (e.g., Nature, Science, Cell, PNAS, eLife) and top artificial intelligence (e.g., JMLR, Neurips, ICML) venues. His group has received funding from the Transformative Research Award from NIH, the NSF CAREER award, Microsoft Research, and many other government, for-profit and nonprofit organizations. He has advised over 60 trainees, and taught about 200 students in his eight years as faculty. In addition to his academic work, he co-founded Global Domain Partners, a quantitative hedge fund that was acquired by Mosaic Investment Partners in 2012, and software startup Gigantum, which was acquired by nVidia in early 2022. He lives in the Chesapeake Bay Watershed with his beloved eternal wife and our three children.
\end{IEEEbiography}

\begin{IEEEbiography}[{\includegraphics[width=1in,height=1.25in,clip,keepaspectratio]{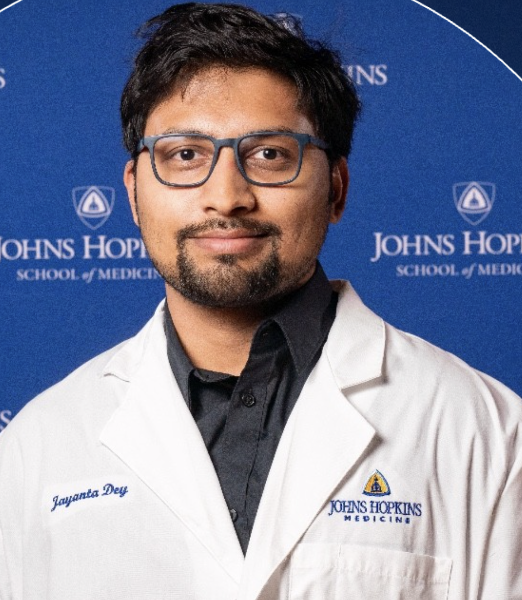}}]{Jayanta Dey} received his B.Sc. and M. Sc. degrees in electrical and electronic engineering from Bangladesh University of Engineering and Technology (BUET), Dhaka, Bangladesh, in 2017 and 2019, respectively with a focus on ultrasound imaging system and signal processing. Subsequently, he earned his M.Sc. and PhD in biomedical engineering from Johns Hopkins University under the supervision of Dr. Joshua T. Vogelstein in 2024 with a focus on out-of-distribution learning. This work is a major part of his PhD thesis. Currently, he is working as a postdoctoral fellow under the supervision of Dr. Dhireesha Kudithipudi in NuAI lab at UTSA with a focus on the temporal aspects of out-of-distribution learning. His research interest includes making artificial intelligence less artificial and more natural.
\end{IEEEbiography}

\begin{IEEEbiography}[{\includegraphics[width=1in,height=1.25in,clip,keepaspectratio]{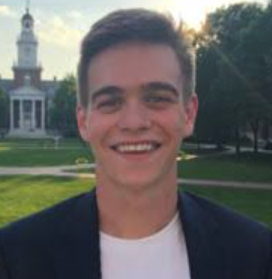}}]{Hayden S. Helm} is a researcher specializing in statistical pattern recognition and machine learning. He is the founder and principal researcher at Helivan Research, based in San Francisco, California. He also holds a position as a researcher at Nomic AI.
\end{IEEEbiography}

\begin{IEEEbiography}[{\includegraphics[width=1in,height=1.25in,clip,keepaspectratio]{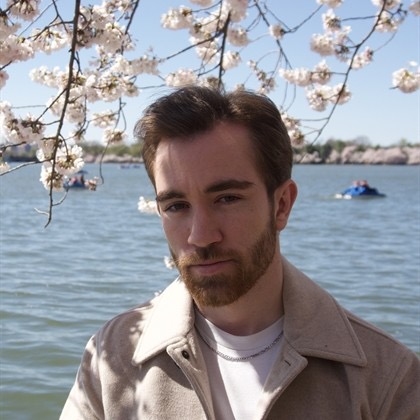}}]{Will LeVine} is a researcher at Microsoft’s Mixed Reality division. He did his Master’s at Johns Hopkins but dropped out of college. His current interests lie at the intersection of interpretability, reliability, and multi-modal learning. 
\end{IEEEbiography}

\begin{IEEEbiography}[{\includegraphics[width=1in,height=1.25in,clip,keepaspectratio]{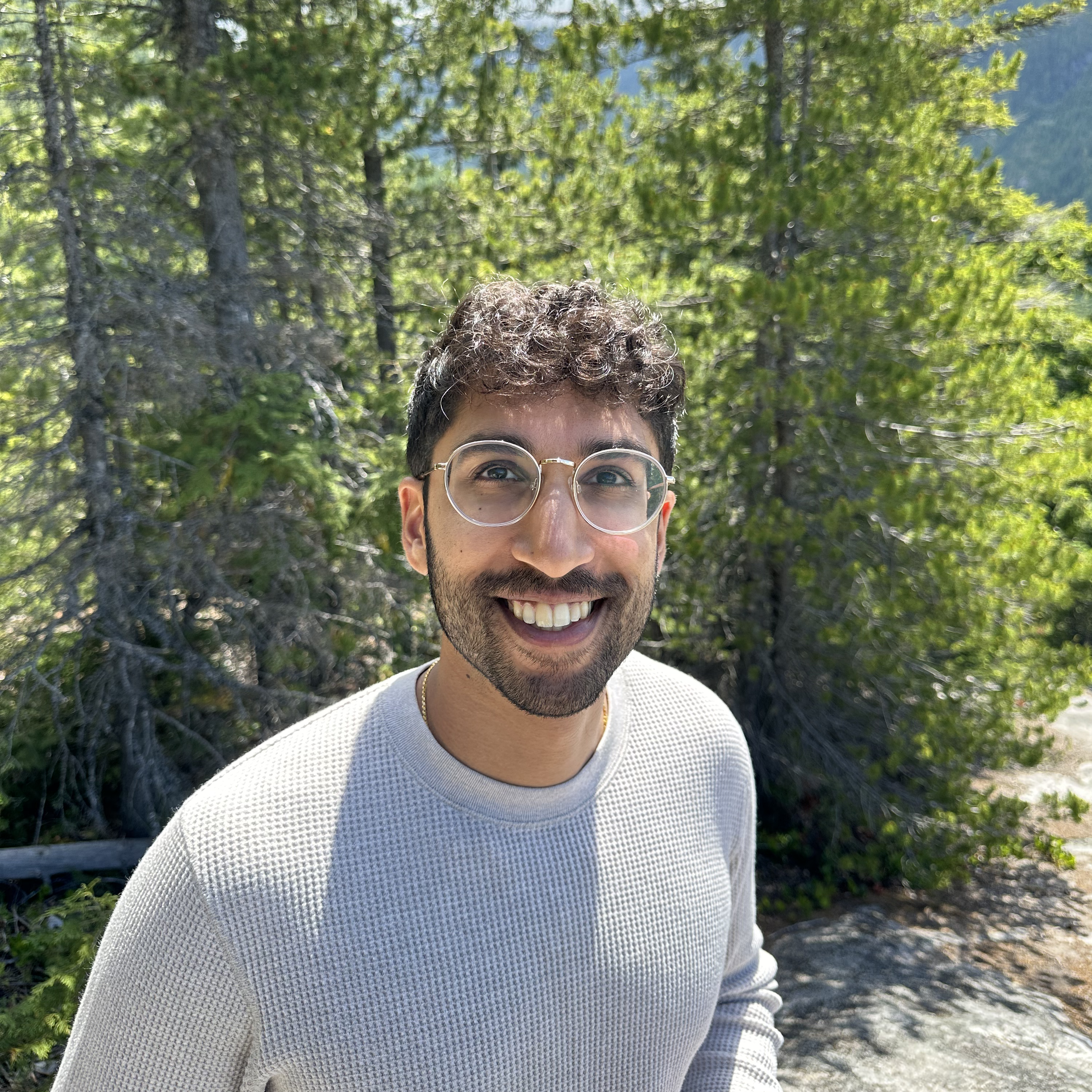}}]{Ronak D. Mehta} is a Ph.D. candidate at the Department of Statistics, University of Washington, advised by Zaid Harchoui. He completed his undergraduate and Master's at Johns Hopkins University under the supervision of Joshua T. Vogelstein and Carey Priebe. His research interests lie broadly in mathematical optimization, distributional shift, and distributional robustness. 
\end{IEEEbiography}

\begin{IEEEbiography}[{\includegraphics[width=1in,height=1.25in,clip,keepaspectratio]{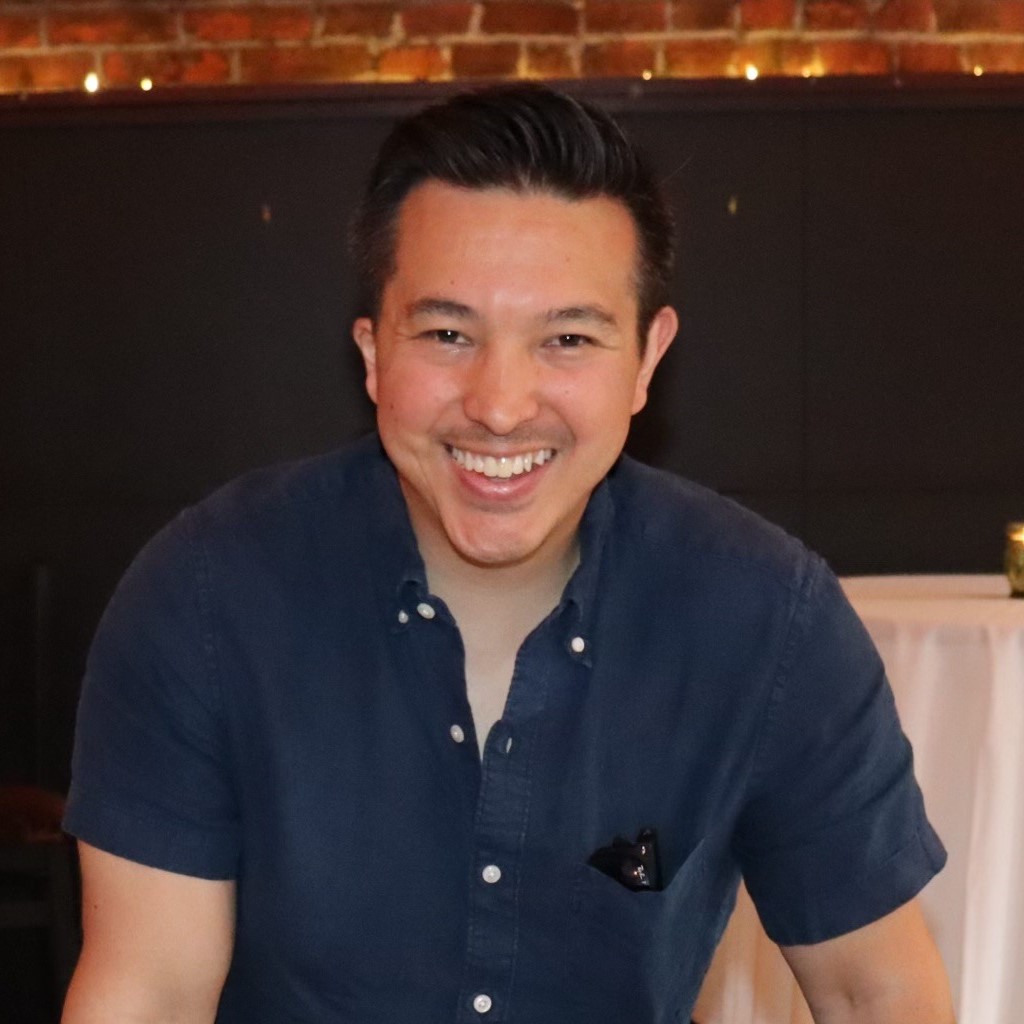}}]{Tyler M. Tomita} is a data scientist and machine learning researcher at the Johns Hopkins University Applied Physics Laboratory. He received his B.S in Biomedical Engineering from the University of California, Davis and his Ph.D in Biomedical Engineering at Johns Hopkins University under the supervision of Dr. Joshua Vogelstein. His interests are in flexible, robust, and data-efficient machine learning solutions. 
\end{IEEEbiography}

\begin{IEEEbiography}[{\includegraphics[width=1in,height=1.25in,clip,keepaspectratio]{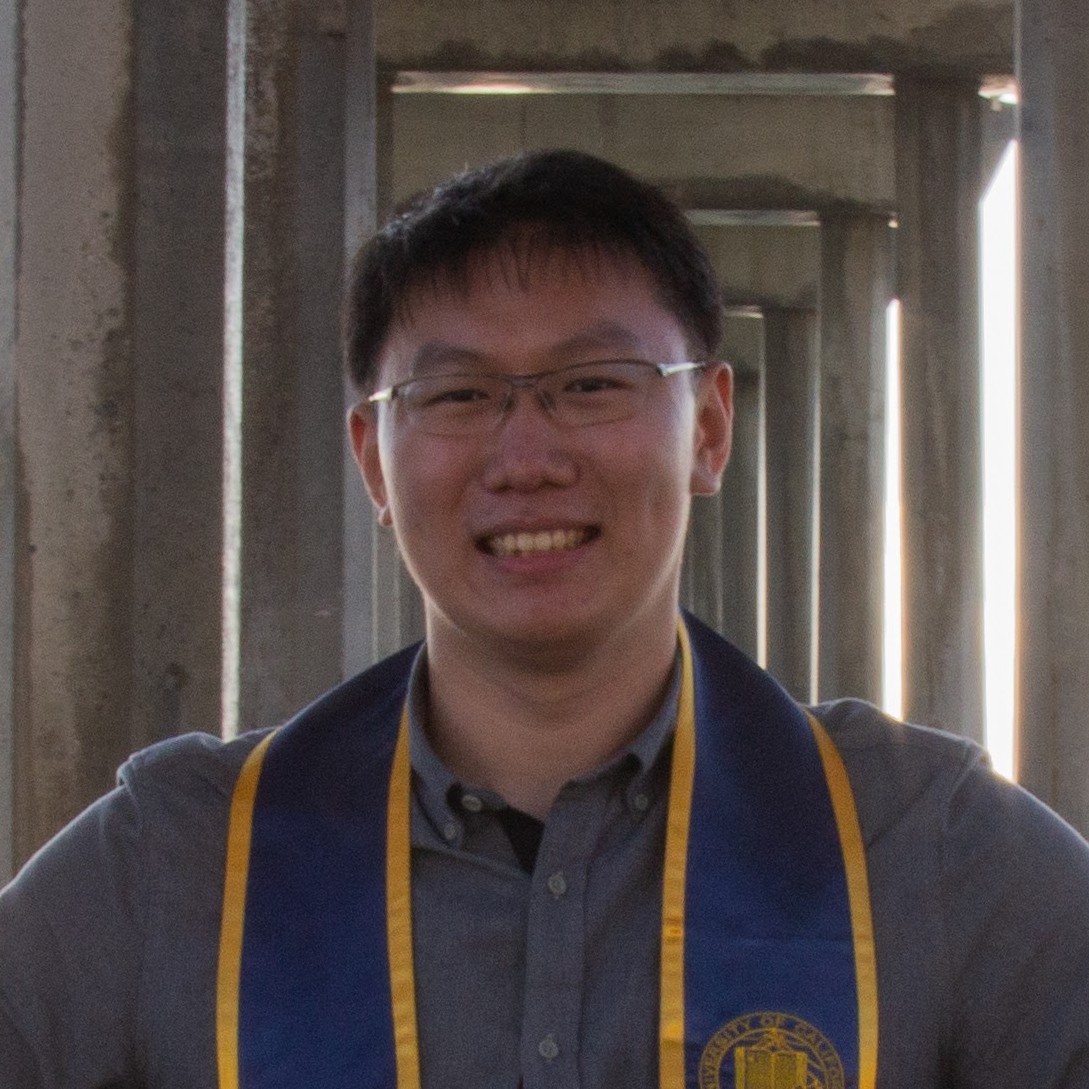}}]{Haoyin Xu} is a PhD student at The Johns Hopkins University, Department of Biomedical Engineering. His research fields include biomedical data science and machine learning. He is currently working on random forests based hypothesis testing and online learning methods.
\end{IEEEbiography}

\begin{IEEEbiography}[{\includegraphics[width=1in,height=1.25in,clip,keepaspectratio]{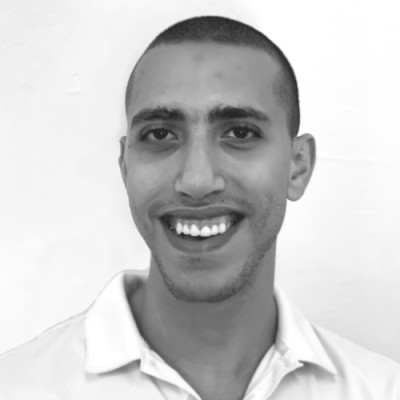}}]{Ali Geisa} is a software engineer at Vendelux. He completed his undergraduate and Master's at Johns Hopkins University, and was part of the Neurodata lab with Dr. Joshua T. Vogelstein. His research interests lie broadly in machine learning and algorithm development.
\end{IEEEbiography}

\begin{IEEEbiography}[{\includegraphics[width=1in,height=1.25in,clip,keepaspectratio]{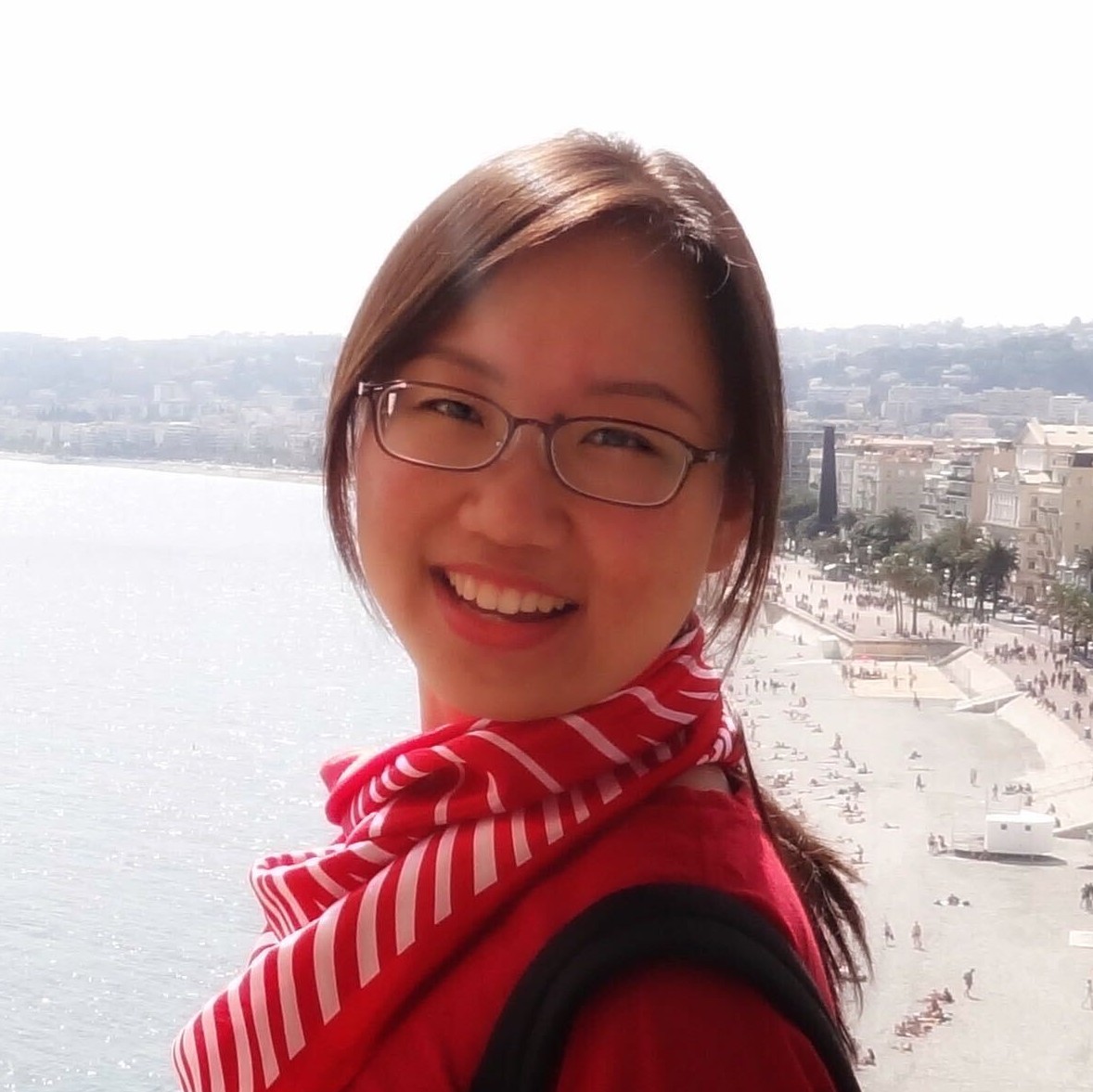}}]{Qingyang (Alice) Wang} is a Ph.D. candidate in the Department of Neuroscience at Johns Hopkins University, where she is advised by Dr. Joshua Vogelstein and Dr. Carey Priebe. She earned her Bachelor of Science degree from the Hong Kong University of Science and Technology, with a double major in Biochemistry and Cell Biology, and Computer Science. Her research focuses on the intersection of neuroscience and AI, exploring how insights from the brain can advance AI and how AI models can deepen our understanding of the brains.
\end{IEEEbiography}

\begin{IEEEbiography}[{\includegraphics[width=1in,height=1.25in,clip,keepaspectratio]{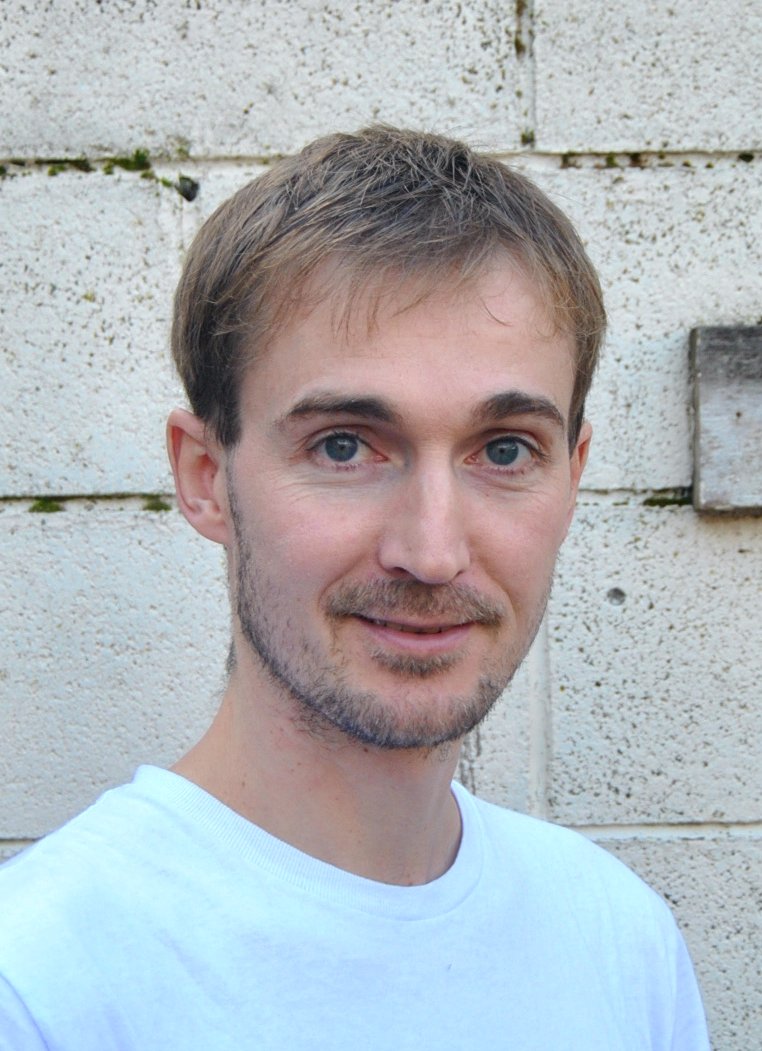}}]{Gido M. van de Ven} is currently a postdoctoral researcher at the KU Leuven in Belgium, where he studies continual learning from both a deep learning and a cognitive science perspective. At the time of writing this paper, he was a postdoctoral researcher at the Baylor College of Medicine in Houston and a visiting researcher at the University of Cambridge.
\end{IEEEbiography}

\begin{IEEEbiography}[{\includegraphics[width=1in,height=1.25in,clip,keepaspectratio]{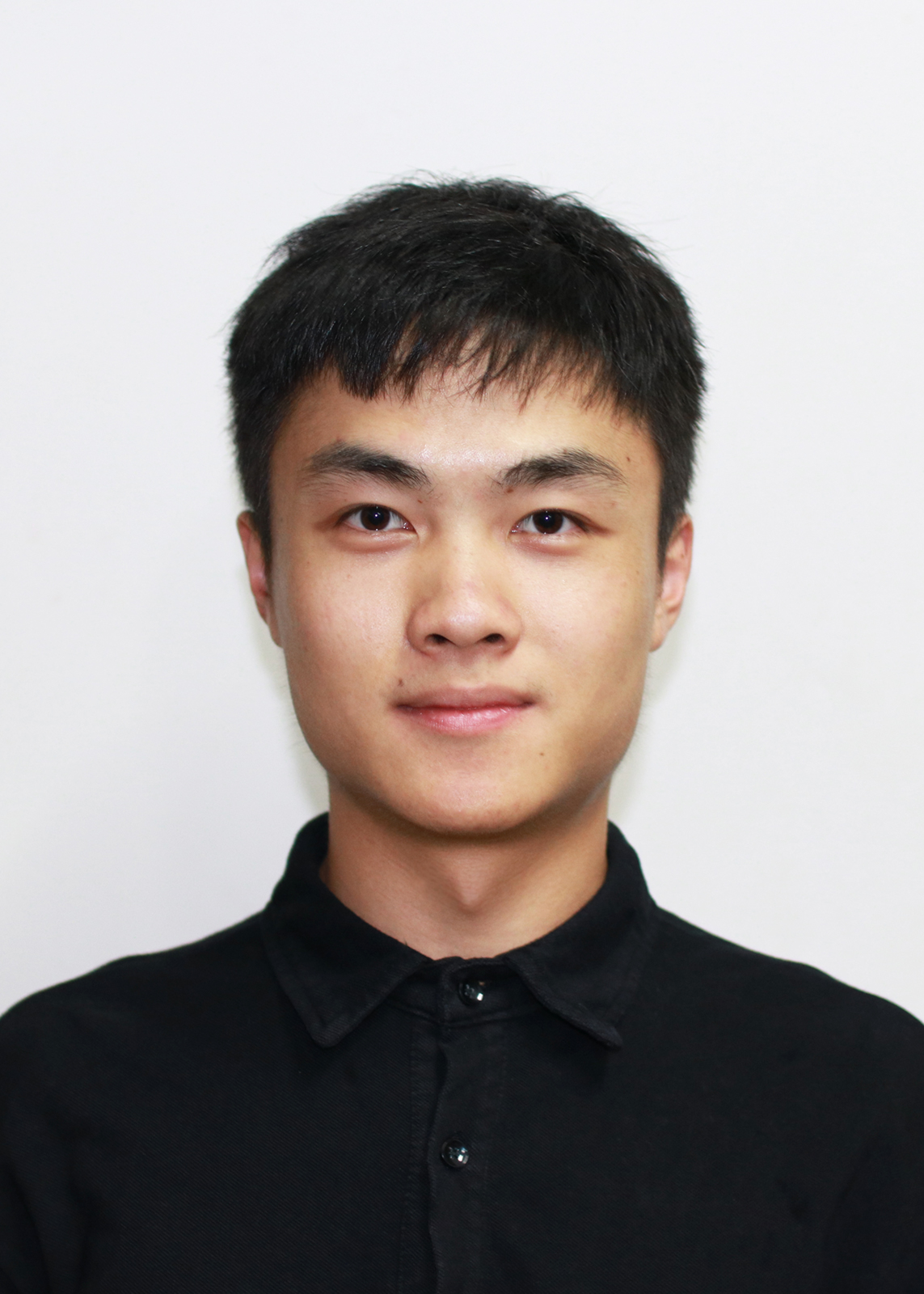}}]{Chenyu Gao} is a PhD student in Electrical and Computer Engineering at Vanderbilt University. His research focuses on image processing and computer vision for medical image analysis. He obtained a master's degree in Biomedical Engineering from Johns Hopkins University.
\end{IEEEbiography}

\begin{IEEEbiography}[{\includegraphics[width=1in,height=1.25in,clip,keepaspectratio]{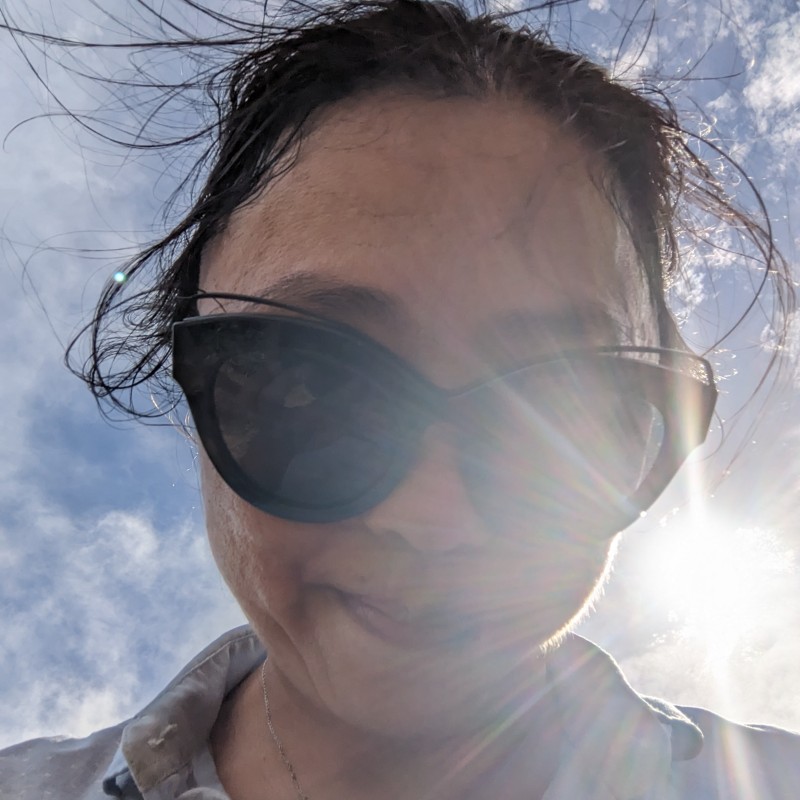}}]{Weiwei Yang} works in resource efficient alt-SGD ML methods inspired by biological learning. The applied research group, she leads, aims to democratize AI by addressing issues of sustainability, robustness, scalability, and efficiency in ML. Her group has applied ML to address social issues such as countering human trafficing and to energy grid stabilizations. 
\end{IEEEbiography}

\begin{IEEEbiography}[{\includegraphics[width=1.25in,height=1in,clip,keepaspectratio]{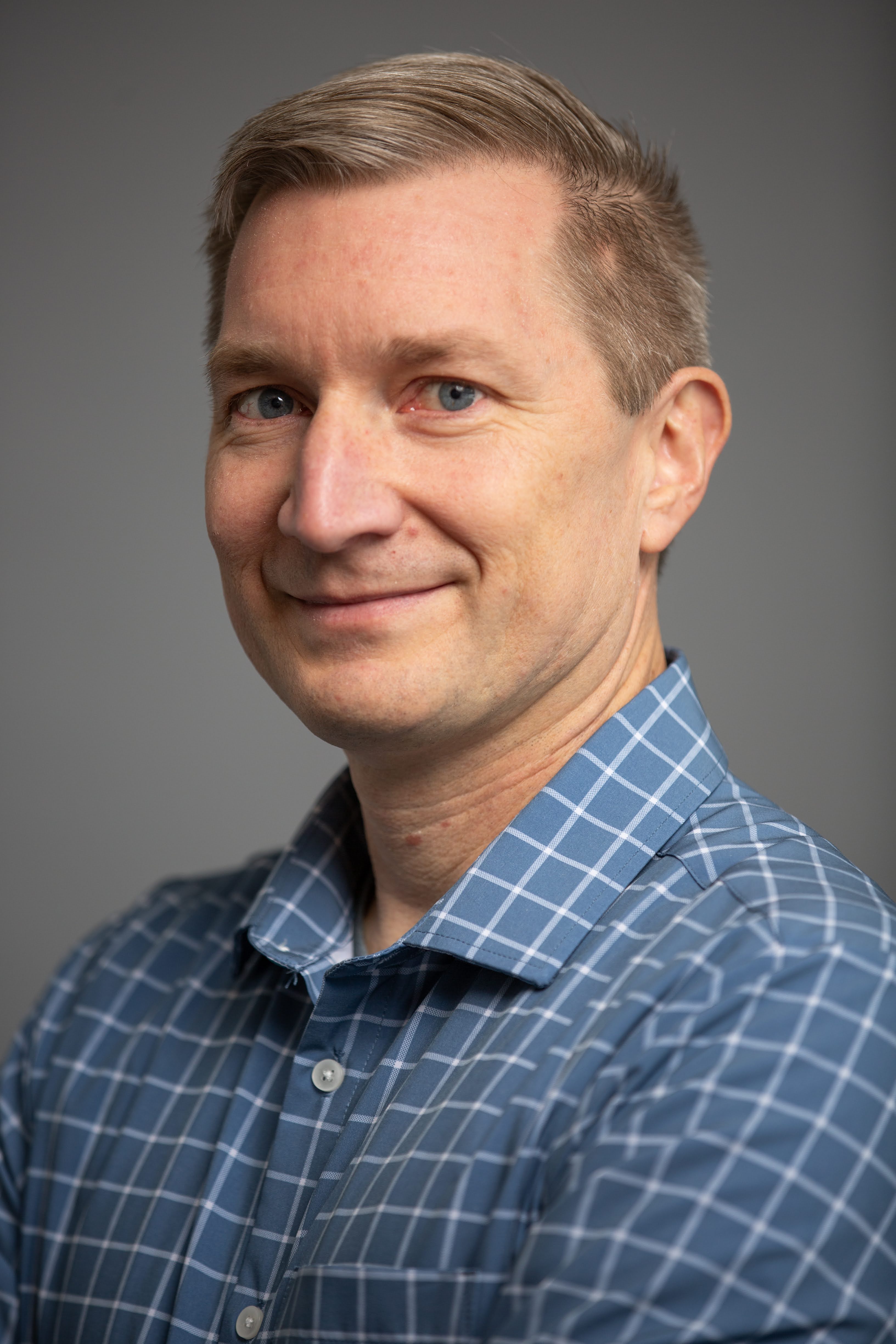}}]{Bryan Tower} has been working with machine learning, information retrieval, and graph learning for over twenty years. Most recently he has been working for Microsoft Research on Large Foundational Models. 
\end{IEEEbiography}

\begin{IEEEbiography}[{\includegraphics[width=1in,height=1in,clip,keepaspectratio]{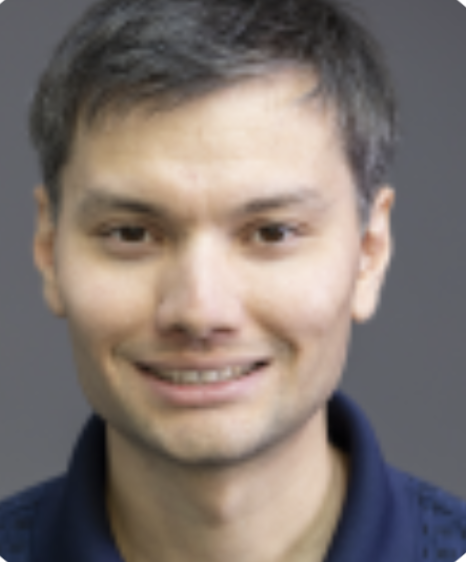}}]{Jonathan Larson} is a Senior Principal Data Architect at Microsoft Research working in Special Projects. He currently leads a research team focused on the intersection of graph machine learning, LLM memory representations, and LLM orchestration.
\end{IEEEbiography}

\begin{IEEEbiography}[{\includegraphics[width=1in,height=1.25in,clip,keepaspectratio]{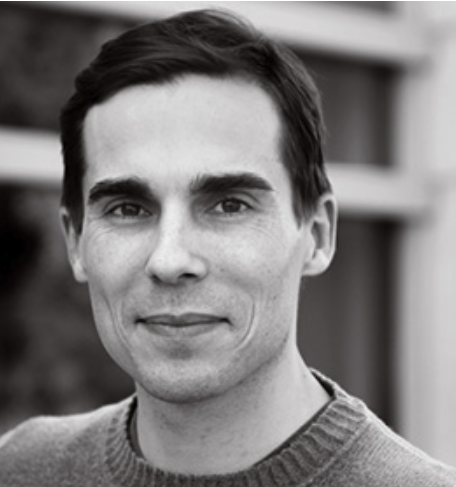}}]{Christopher M. White} is partner and managing director of special projects at Microsoft. He leads research teams with world-class specialists solving highly uncertain, complex problems. His group builds technologies to benefit society, including tools for digital safety, plurality, and evidence-based policy.

\end{IEEEbiography}

\begin{IEEEbiography}[{\includegraphics[width=1in,height=1.1in,clip,keepaspectratio]{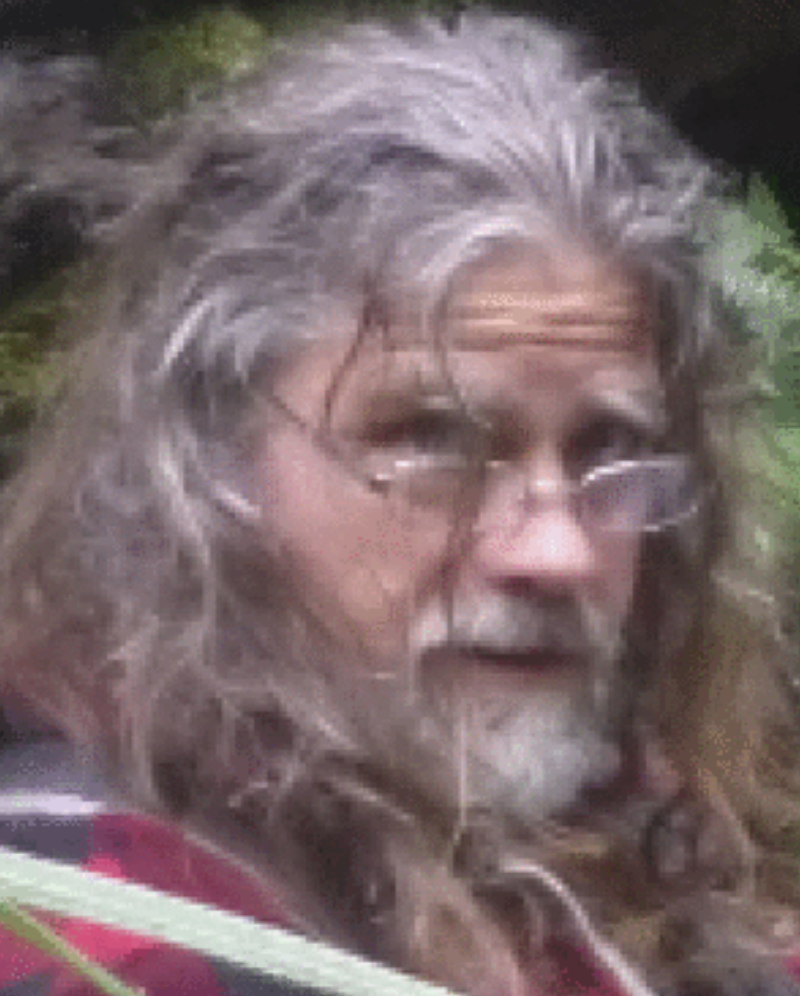}}]{Carey E. Priebe} (Senior Member, IEEE) received the B.S. degree in mathematics from Purdue University, West Lafayette, IN, USA, in 1984, the M.S. degree in computer science from San Diego State University, San Diego, CA, USA, in 1988, and the Ph.D. degree in information technology (computational statistics) from George Mason University, Fairfax, VA, USA, in 1993. From 1985 to 1994, he was a mathematician and scientist with the US Navy research and development laboratory system. Since 1994, he has been a Professor with the Department of Applied Mathematics and Statistics, Johns Hopkins University. His research interests include computational statistics, kernel and mixture estimates, statistical pattern recognition, model selection, and statistical inference for high-dimensional and graph data. He is an Elected Member of the International Statistical Institute, a Fellow of the Institute of Mathematical Statistics, and a Fellow of the American Statistical Association.
\end{IEEEbiography}

\clearpage
\newpage
\smallskip
\appendix
\label{appendix}

\setcounter{figure}{0}    
\setcounter{table}{0}    

\subsection{Decomposition of $\mathsf{Transfer}$}
\label{app:decomposition_tx}
$\mathsf{Transfer}$ can be decomposed into $\mathsf{Forward~Transfer}$ and $\mathsf{Backward~Transfer}$:
\begin{align} \label{eq:dcompose}
\mathsf{Transfer}^t(f) &=  \log \frac{\mc{E}^t_f(\Dn^t)}{\mc{E}^t_f(\bigcup_{t'=1}^{T}\Dn^{t'})}\\
&= \log \frac{\mc{E}^t_f(\Dn^t)}  {\mc{E}^t_f(\bigcup_{t'=1}^t\Dn^{t'})} + \log \frac{\mc{E}^t_f(\bigcup_{t'=1}^t\Dn^{t'})}{\mc{E}^t_f(\bigcup_{t'=1}^{T}\Dn^{t'})}\\
=& \mathsf{Forward~Transfer}^t(f) + \mathsf{Backward~Transfer}^t(f).
\end{align}
 We say that an algorithm $ f $ has  transferred to task $t$ from all the tasks up to $T$
if and only if 
\begin{equation}
    \mathsf{Transfer}^t(f) > 0.
\end{equation}
\section{Evaluation Criteria}
\label{app:evaluation-ceiterion}

\subsection{Representation Ensembling Algorithms}
\label{app:progl_algs}

\subsubsection{Model Architecture}
In this paper, we  proposed two representation ensembling algorithms, Simple Lifelong Learning Networks (\PLN) and Simple Lifelong Learning Forests (\PLF).  The two algorithms differ in their details of how to update encoders and channels, but abstracting a level up they are both special cases of the same procedure.  Let \sct{SiLLy-X} refer to any possible representation algorithm.  Algorithms ~\ref{alg:odxtrain},~\ref{alg:synx_add_voter},~\ref{alg:synx_update_voter}, and ~\ref{alg:synx_predict} provide pseudocode for adding encoders, updating channels, and doing inference for any \sct{SiLLy-X} algorithm. 

\subsubsection{Data Preparation}
Whenever the learner gets access to a new task data, we use Algorithm \ref{alg:odxtrain} to train a new encoder for the corresponding task. We split the data into two portions --- in-task set and held out or out-of-bag set.

\subsubsection{Training Procedures}
In-task set is used to learn the encoder and the indices of the out-of-bag (OOB) data which is returned by Algorithm \ref{alg:odxtrain} to be used by Algorithm \ref{alg:synx_add_voter} to learn the channel for the corresponding task. 
Note that we push the OOB data through the in-task encoder and the whole dataset through the cross-task encoders to update the channel, i.e, learn the posteriors according to the new encoder (see Algorithm \ref{alg:synx_update_voter}). 
Finally,  Algorithm \ref{alg:synx_predict} does inference on a new sample. Given the task identity, we use the corresponding channel to get the average estimated posterior and predict the class label as the $\argmax$ of the estimated posteriors.

%\subsection{Omnidirectional Forests}

\begin{algorithm}[!htbp]
  \caption{Add a new \sct{SiLLy-X}~encoder for a task. OOB = out-of-bag.
}
  \label{alg:odxtrain}
\begin{algorithmic}[1]
  \Require 
  \Statex (1) $t$ \Comment{current task number}
  \Statex (2) $\mathcal{D}_n^t = (\mathbf{x}^t,\mathbf{y}^t) \in \Real^{n \times p} \times \{1,\ldots, K\}^n$ \Comment{training data for task $t$}
\Ensure 
\Statex (1) $u_t$  \Comment{an encoder trained on task $t$}
\Statex (2) $\mc{I}_{OOB}^t$  \Comment{a set of the indices of OOB data}

\Function{\sct{SiLLy-X}.fit}{$t, (\mathbf{x}^t,\mathbf{y}^t)$}
\State $u_t, \mc{I}_{OOB}^t \leftarrow$ encoder.fit($\mathbf{x}^t$, $\mathbf{y}^t$) 
\Comment train an encoder on training data partitioned into in-bag and OOB samples
    \State \Return $u_t, \mc{I}_{OOB}^t$ 
\EndFunction 
\end{algorithmic}
\end{algorithm}

\begin{algorithm}[!htbp]
  \caption{Add a new \sct{SiLLy-X}~channel for the current task. 
}
  \label{alg:synx_add_voter}
\begin{algorithmic}[1]
  \Require 
  \Statex (1) $t$ \Comment{current task number}
  \Statex (2) $\mc{U} = \{u_{t'}\}_{t'=1}^t$ \Comment{the set of encoders}
  \Statex (3) $\mathcal{D}_n^t = (\mathbf{x}_t,\mathbf{y}_t) \in \Real^{n \times p} \times \{1,\ldots, K\}^n$  \Comment{training data for task $t$}
  \Statex (4) $\mc{I}_{OOB}^t$  \Comment{a set of the indices of OOB data for the current task}
\Ensure $v_t$  \Comment channel for task $t$

\Function{\sct{SiLLy-X}.add\_channel}{$t, u_t, (\mathbf{x}_t, \mathbf{y}_t), \mc{I}_{OOB}^t$}

\State $v_{t} \leftarrow u_{t}$.add\_channel($(\mathbf{x}_t, \mathbf{y}_t), \mc{I}_{OOB}^t$) 
\Comment add the new in-task channel using OOB data

\For{$t' = 1, \ldots, t-1$}
    \State $v_{t} \leftarrow u_{t'}$.update\_channel($\mathbf{x}_t, \mathbf{y}_t, v_{t}$)
    \Comment update the channel for task $t$ using the old encoders
\EndFor
    \State \Return $v_t$ 
\EndFunction 
\end{algorithmic}
\end{algorithm}

\begin{algorithm}[H]
  \caption{Update \sct{SiLLy-X}~channel for the previous tasks. 
}
  \label{alg:synx_update_voter}
\begin{algorithmic}[1]
  \Require 
  \Statex (1) $t$ \Comment current task number 
  \Statex (2) $u_t$ \Comment encoder for the current task
  \Statex (3) $\mathcal{D} =\{\mathcal{D}^{t'} \}_{t'=1}^{t-1}$ \Comment training data for old tasks
  \Statex (4) $\mc{V} = \{\mb{v}_{t'}\}_{t'=1}^{t-1}$ \Comment set of all  previous task voters
\Ensure $\mc{V} = \{\mb{v}_{t'}\}_{t'=1}^{t-1}$

\Function{\sct{SiLLy-X}.update\_old\_channel}{$t, u_t, \mathcal{D}, \mc{V}$}
% \State $t' \leftarrow 1$
% \Comment initialize task index to iterate over the previous tasks

\For{$t'= 1, \ldots, t-1$}
    \State $v_{t'} \leftarrow u_{t}$.update\_channel($\mathcal{D}^{t'}, v_{t'}$)
    \Comment update the old task channels
    % \State $t' \leftarrow t'+1$
\EndFor
    \State \Return $\mc{V}$ 
\EndFunction 
\end{algorithmic}
\end{algorithm}

\begin{algorithm}[H]
  \caption{Predicting a class label  using \sct{SynX}. 
}
  \label{alg:synx_predict}
\begin{algorithmic}[1]
  \Require
  \Statex (1) $\mb{x} \in \Real^{p}$ \Comment test datum 
  \Statex (2) $t$ \Comment task identity associated with $\mb{x}$
  \Statex (3) $\mc{U}$  \Comment set of all $t$ encoders
  \Statex (4) $v_t$ \Comment channel for task $t$
\Ensure $\hat{y}$ \Comment a predicted class label 

\Function{$\hat{y} =$ \sct{SiLLy-X}.predict}{$t, \mb{x}, v_t$} 
    % \State $T \leftarrow$ \sct{SynX}.get\_task\_number() 
    % \Comment get the total number of tasks
    % \State $t' = 1$ 
% \Comment initialize task index to iterate over the tasks
%     \State $\hat{\mathbf{p}}_t = \mathbf{0}$ \Comment $\hat{\mathbf{p}}_t$ is a $K$-dimensional posterior vector
%     \For{$t' = 1,\ldots, T$} 
%     \Comment aggregate the posteriors calculated from $T$-th task channel
%     \State $\hat{\mathbf{p}}_t \leftarrow \hat{\mathbf{p}}_t + v_{tt'}$.predict\_proba($u_{t'}(x)$) 
% % \State $t' \leftarrow t' + 1$
%     \EndFor 
%     \State $\hat{\mathbf{p}}_t \leftarrow \hat{\mathbf{p}}_t / T$
\For{$t' = 1,\ldots, t$} 
\Comment get the output $\tilde{\mb{x}} = \{\tilde{\mb{x}}_{t'}\}_{t'=1}^t$ from all the encoders
    \State $\tilde{\mb{x}}_{t'} \leftarrow u_{t'}.encode(\mb{x})$
\EndFor

\State $\hat{\mathbf{p}} \leftarrow v_t.predict\_proba(\tilde{\mb{x}})$
\Comment $\hat{\mathbf{p}}$ is a $K_t$-dimensional posterior vector
\State $\hat{y} = \argmax(\hat{\mathbf{p}})$
\Comment find the index of the elements in the vector $\hat{\mathbf{p}}$ with maximum value
\State \Return $\hat{y}$
\EndFunction 
\end{algorithmic}
\end{algorithm}

\begin{figure*}[!ht]
    \centering
    \includegraphics[width=.7\linewidth]{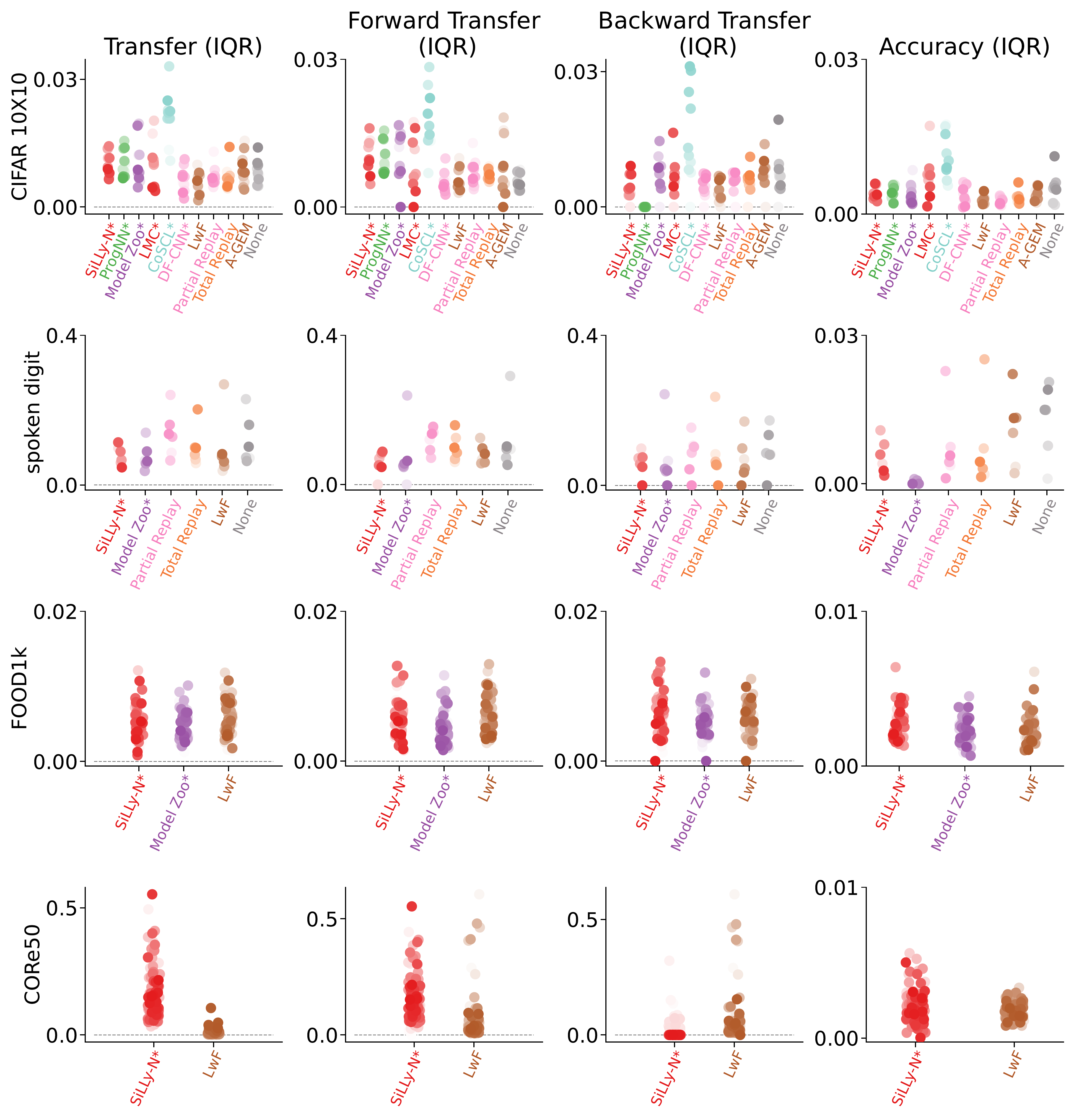}
    \caption{\textbf{Error bars (interquartile range, i.e., IQR) for each dot in Figure \ref{fig:strip} on different vision and speech benchmark datasets.} Error bars for different performance statistics is negligible in comparison with the median performance shown in the main text Figure \ref{fig:strip}.
    }
    \label{fig:strip_IQR}
\end{figure*}

\begin{figure*}[!htbp]
    \centering
    \includegraphics[width=.7\linewidth]{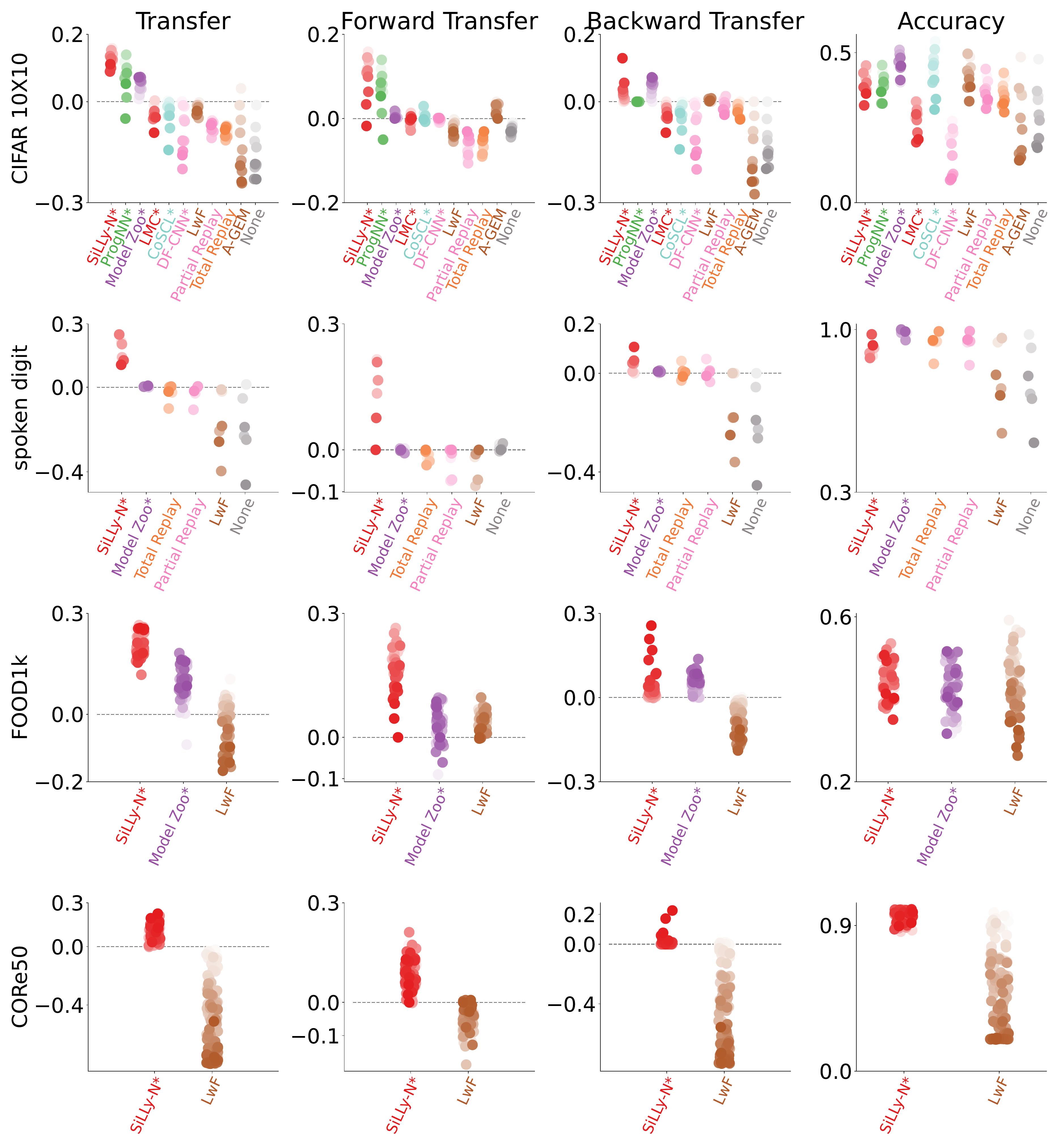}
    \caption{\textbf{Performance summary on vision and audition benchmark datasets using Veniat's \cite{veniat2020efficient}'s statistics.} See Figure~\ref{fig:strip} for caption details. Note that the results here look nearly identical other than the y-axis labels. 
    }
    \label{fig:strip_overall}
\end{figure*}

\begin{figure*}[!ht]
    \centering
    \includegraphics[width=.7\linewidth]{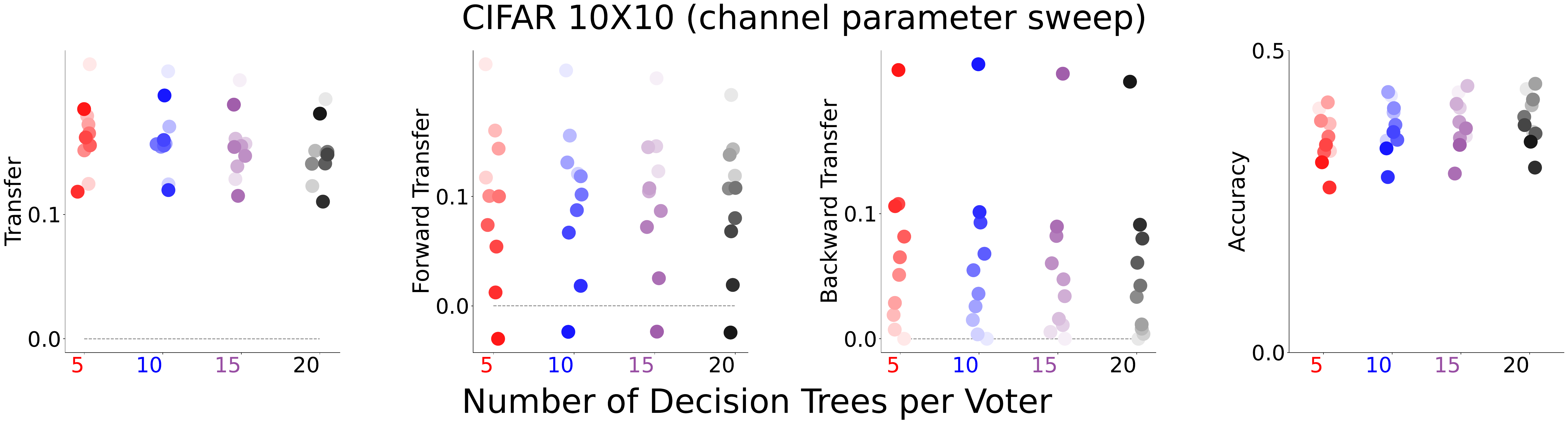}
    \caption{\textbf{Performance of \PLN\ on CIFAR 10X10 remains nearly unchanged for different number of decision trees per channel.}
    }
    \label{fig:forest_voter}
\end{figure*}

% \begin{table}[H]
% % \centering
% \caption{Hyperparameters for \PLF\ in CIFAR-10X10 experiments. }
%     \label{tab:hyperparameter_table}
% \begin{center}
% \begin{tabular}{|l|l|}
%     \hline
%         \textbf{Hyperparameters} & \textbf{Value} \\
%     \hline
%         n\_estimators ($500$ training samples per task) & $10$\\
%     \hline
%         n\_estimators ($5000$ training samples per task) & $40$\\
%     \hline
%         max\_depth & $30$\\
%     \hline
%         max\_samples (OOB split) & $0.67$\\
%     \hline
%         min\_samples\_leaf & $1$\\
%     \hline
% \end{tabular}
% \end{center}

% \end{table}

\begin{table}[H]
% \centering
\caption{Hyperparameters for \PLN\ in CIFAR 10X10, Five Datasets, Split Mini-Imagenet, FOOD1k experiments. Note that we use the same hyperparameters for all the experiments.}
    \label{tab:hyperparameter_table_synn}
\begin{center}
\begin{tabular}{|l|l|}
    \hline
        \textbf{Hyperparameters} & \textbf{Value} \\
    \hline
        optimizer  & Adam\\
    \hline
        learning rate & $3 \times 10^{-4}$\\
    \hline
        max\_samples (OOB split) & $0.67$\\
    \hline
      n\_estimators (decision forest channel) & $20$ \\ \hline
\end{tabular}
\end{center}

\end{table}

\begin{table}[ht]
% \centering
\caption{Hyperparameters for \PLF\ in tabular CIFAR 10X10 experiments.}
    \label{tab:hyperparameter_table_5_dataset}
\begin{center}
\begin{tabular}{|l|l|}
    \hline
        \textbf{Hyperparameters} & \textbf{Value} \\
    \hline
        n\_estimators & $10$\\
    \hline
        max\_depth & $30$\\
    \hline
        max\_samples (OOB split) & $0.67$\\
    \hline
        min\_samples\_leaf & $1$\\
    \hline
\end{tabular}
\end{center}

\end{table}

%\subsection{Omnidirectional Networks}

% This method of using the estimated posteriors from a transfer classifier trained on the penultimate activations of an DN is inspired by~\cite{}.

% \clearpage

\subsection{Reference Algorithm Implementation Details}\label{app:archs}
The same network architecture was used for all baseline deep learning methods. Following the work in ~\cite{tolias_architecture}, the `base network architecture' consisted of five convolutional layers followed by two-fully connected layers each containing 2000 nodes with ReLU non-linearities and a softmax output layer. The convolutional layers had $16$, $32$, $64$, $128$ and $254$ channels, they used batch-norm and a ReLU non-linearity, they had a 3x3 kernel, a padding of 1 and a stride of 2 (except the first layer, which had a stride of 1). This architecture was used with a multi-headed output layer (i.e., a different output layer for each task) for all algorithms using a fixed-size network. For ProgNN and DF-CNN the same architecture was used for each column introduced for each new task, and in our \PLN\ this architecture was used for the transformers $u_t$ (see above). 
In these implementations,  ProgNN and DF-CNN have the same architecture for each column introduced for each task. Among the reference algorithms, \sct{EWC}, \sct{O-EWC}, \sct{LwF}, \sct{SI}, \sct{Total Replay} and \sct{Partial Replay} results were produced using the repository \url{https://github.com/GMvandeVen/progressive-learning-pytorch}. For \ProgNN\ and \DFCNN\, we used the code provided in \url{https://github.com/Lifelong-ML/DF-CNN}. For all other reference algorithms, we modified the code provided by the authors to match the deep net architecture as mentioned above and used the default hyperparameters provided in the code. 
\subsection{Training Time Complexity Analysis}
\label{app:complexity}
We use the soft-O notation $\tilde{\mc{O}}$ to quantify complexity~\citep{Van_Rooij2019-hu}. Letting $n$ be the sample size and $T$ be the number of tasks, we write that the capacity, space or time complexity of a lifelong learning algorithm is $f(n,t) = \tilde{\mc{O}}(g(n,T))$ when $|f|$ is bounded above asymptotically by a function $g$ of $n$ and $T$ up to a constant factor and  polylogarithmic terms. For simplifying the calculation, we make the following assumptions:
\begin{enumerate}
    \item Each task has the same number of training samples.
    \item Capacity grows linearly with the number of trainable parameters in the model.
    \item The number of epochs is fixed for each task.
    \item For the algorithms with dynamically expanding capacity, we assume the worst case scenario where an equal amount of capacity is added to the hypothesis with an additional task.
\end{enumerate}
Assumption $3$ enables us to write time complexity as a function of the sample size. Table~\ref{tab:tax} summarizes the capacity, space and time complexity of several reference algorithms, as well as our \PLN~and \PLF.

Lifelong learning methods are parametric if they have a representational capacity which is invariant to sample size and task number. Although the space complexity of some of these algorithms grow (because the number of the constraints stored by the algorithms grows, or they continue to store more data), their capacity is fixed. Thus, given a sufficiently large number of tasks with increasing complexity, in general, eventually all parametric methods will catastrophically forget. \sct{EWC} \citep{kirkpatrick2017overcoming}, \sct{Online EWC} \citep{schwarz2018progress}, \sct{SI} \citep{zenke2017continual}, and  \sct{LwF} \citep{li2017learning} are all examples of parametric lifelong learning algorithms. Our fixed resource algorithms are also parametric.
%For comparison, we implement another baseline algorithm and refer to it as \sct{Total Replay}, which is also parametric. \sct{Total Replay} replays both old and current task data while learning a new task.}

Lifelong learning methods are semi-parametric  if they have a representational capacity which grows slower than sample size.  For example, if $T$ is increasing slower than $n$ (e.g., $T \propto \log n$), then algorithms whose capacity is proportional to $T$ are semi-parametric.  
\ProgNN\ \citep{rusu2016progressive} is semi-parametric, nonetheless, its space complexity is $\tilde{\mc{O}}(T^2)$ due to the lateral connections.    
Moreover, the time complexity for \ProgNN\ also scales quadratically with $n$ when $n \propto T$. Thus, an algorithm that literally stores all the data it has ever seen, and retrains a fixed size network on all those data with the arrival of each new task, would have smaller space complexity and the same time complexity as \ProgNN.    
\sct{DF-CNN}~\citep{Lee2019-eg} improves upon \ProgNN\ by introducing a ``knowledge base'' with lateral connections to each new column, thereby avoiding all pairwise connections.  
Because these semi-parametric methods have a fixed representational capacity per task, they  will either lack the representation capacity to perform well given sufficiently complex tasks, and/or will waste resources for very simple tasks.   

Lifelong learning methods are non-parametric  if they have a representational capacity which \emph{grow in proportion} to the number of tasks or data samples. Table \ref{tab:tax} shows the Indian Buffet Process for Weight Factors (\sct{IBP-WF}) is a notable non-parametric approach alongside \PLF.

Our proposed approaches, \PLN\ and \PLF, as we will discuss in details in Section \ref{sec:algorithms}, eliminate lateral connections between the columns of the network, thus reducing the complexity of the space to $\tmc{O}(T)$. Moreover, our proposed approaches can adapt flexibly to any of the three categories based on the constraints of the application environment, as illustrated in Table \ref{tab:tax}.
%(Figure~\ref{fig:size} top left). Memory consumed by new channels is negligible compared to that of memory required for storing the encoders (Figure~\ref{fig:size} top right). The time required for inference on $1000$ testing points is order of magnitude lower in comparison with the time required for training a new encoder with $500$ samples (Figure~\ref{fig:size} bottom).
%Indian Buffet Process for Weight Factors (\sct{IBP-WF}) %\citep{mehta2021continual} proposed the only is another non-parametric lifelong learning algorithm. 

\subsubsection{Complexity analysis}
Consider a lifelong learning environment with $T$ tasks each with $n'$ samples, i.e., total training samples, $n = n'T$. For all the algorithm with time complexity $\tmc{O}(n)$, the training time grows linearly with more training samples. We discuss all other algorithms with non-linear time complexity below.

\paragraph{EWC}
Consider the time required to train the weights for each task in \sct{EWC}~is $k_c n'$ and each task adds additional $k_l n'$ time from the regularization term. Here, $k_c$ and $k_l$ are both constants. Therefore, time required to learn all the $T$ tasks can be written as:

\begin{align}
\label{eq:ewc_complexity}
      & k_cn' + (k_c n' + k_l n') + \cdots + ( k_c n' + (T-1)k_l n') \nonumber\\
       &= k_c n'T + k_l n' \sum_{t=1}^{T-1} t  \nonumber\\
       &= k_c n' T + k_l n' \frac{T(T-1)}{2} \nonumber \\
       &= k_c n + 0.5 k_l n T - 0.5 k_l n \nonumber \\
       &= \tmc{O}(nT).
\end{align}

\paragraph{Total Replay}
Consider the time to train the model on $n'$ samples is $k_c n'$. Therefore, time required to learn all the $T$ tasks can be written as:

\begin{align}
    \label{eq:replay_complexity}
     & k_c n' + k_c (n'+n') + \cdots + k_cn'T \nonumber\\
      &= k_c n' \sum_{t=1}^T t \nonumber\\
      &=  k_c n' \frac{T(T+1)}{2} \nonumber\\
      &= 0.5 k_c nT + 0.5k_cn \nonumber\\
      &= \tmc{O}(nT)
\end{align}

\paragraph{\ProgNN}
Consider the time required to train each column in \ProgNN~is $k_c n'$ and each lateral connection can be learned with time $k_l n'$. Therefore, time required to learn all the $T$ tasks can be written as:

\begin{align}
\label{eq:prognn_complexity}
      & k_cn' + (k_c n' + k_l n') + \cdots + ( k_c n' + (T-1)k_l n') \nonumber\\
       &= k_c n'T + k_l n' \sum_{t=1}^{T-1} t  \nonumber\\
       &= k_c n' T + k_l n' \frac{T(T-1)}{2} \nonumber \\
       &= k_c n + 0.5 k_l n T - 0.5 k_l n \nonumber \\
       &= \tmc{O}(nT)
\end{align}

\subsection{Simulation Experiment Details}
\label{app:sim}
In each simulation, we constructed an environment with two tasks.  
For each, we sample 750 times from the first task, followed by 750 times from the second task. 
These 1,500 samples comprise the training data.
We sample another 1,000 hold out samples to evaluate the algorithms.  
% 
%We fit a random forest (\RF) (technically, an uncertainty forest which is an honest forest with a finite-sample correction~\citep{Guo2019-xe}) and a \PLF. 
For \PLN, we have used a deep network (\DN) architecture with two hidden layers each having $10$ nodes. Similarly, for \PLN\ experiments we did $100$ repetitions and reported the results after smoothing it using moving average with a window size of $5$. 
%For the \PLF\ experiments we used $1000$ repetitions and reported the mean of these repetitions. 
%We repeat this process 30 times to obtain errorbars. Error bars in all cases were negligible.

Gaussian XOR is two class classification problem with equal class priors. Conditioned on being in class 0, a sample is drawn from a mixture of two Gaussians with means $\pm \begin{bmatrix}  0.5, &  0.5\end{bmatrix}\T $, and variances proportional to the identity matrix. Conditioned on being in class 1, a sample is drawn from a mixture of two Gaussians with means $ \pm \begin{bmatrix}  0.5, & - 0.5\end{bmatrix}\T $, and variances proportional to the identity matrix. Gaussian XNOR is the same distribution as Gaussian XOR with the class labels flipped.
Rotated XOR (R-XOR) rotates XOR by $\theta^\circ$ degrees.

This simulation setup facilitates the manipulation of task overlap, allowing for an examination of the transfer properties of our proposed approach under different levels of task similarity (see main text).

\subsection{Real Data Extended Experiments and Details} 
\label{app:cifar}
 This section contains extended results on algorithms not shown in the main text (see Appendix Figure \ref{fig:vision}). FOOD1k and Mini-Imagenet datasets were obtained from \url{https://www.kaggle.com/datasets/whitemoon/miniimagenet} and \url{https://github.com/pranshu28/TAG}, respectively.

\begin{table*}[!htb]
\centering
\caption{Benchmark dataset details.}
\label{tab:dataset-table}
\resizebox{\textwidth}{!}{
\begin{tabular}{lllll}\hline
 Experiment & Dataset & Training samples & Testing samples & Dimension\\ \hline
 CIFAR 10X10 & CIFAR 100 & $5000$ & $10000$ & $3\times32\times32$ \\ \hline 
\multirow{5}{*}{5-dataset} & CIFAR-10 & $50000$ & $10000$ & \multirow{5}{*}{$3\times32\times32$ (resized)}\\
 & MNIST & $60000$ & $10000$ \\
 & SVHN & $73257$ & $26032$ \\
 & notMNSIT & $16853$ & $1873$ \\
 & Fashion-MNIST & $60000$ & $10000$\\ \hline 
 Split Mini-Imagenet & Mini-Imagenet & $48000$ & $12000$ & $3\times84\times84$ \\ \hline

 FOOD1k 50X20 & Food1k & $60000$ & $99682$ & $3\times50\times50$ (resized)\\ \hline
 Spoken Digit & Spoken Digit & $1650$ & $1350$ & $28 \times 28$ (processed and resized) \\ \hline
\end{tabular}
}
\end{table*}

\paragraph{Split Mini-Imagenet}
In this experiment, we have used the \textbf{Mini-Imagenet} dataset \citep{malviya2021tag}.
% provided in \url{https://www.kaggle.com/datasets/whitemoon/miniimagenet}. 
The dataset was split into $20$ tasks with $5$ classes each. Each task has $2400$ training samples and $600$ testing samples. As shown in Figure~\ref{fig:vision} right column, we get positive forward and backward transfer for \PLN. However, although samples per task is lower compared to that of 5-dataset, it is still quite high. Hence, \zoo\ outperforms all the algorithms in this experiment.

\paragraph{5-dataset}

In this experiment, we have used \textbf{5-dataset} \citep{malviya2021tag}.
% provided in \url{https://github.com/pranshu28/TAG}. 
It consists of $5$ tasks from five different datasets: CIFAR-10 \citep{krizhevsky2009learning}, MNIST, SVHN \citep{netzer2011reading}, notMNIST \citep{notMNIST}, Fashion-MNIST \citep{xiao2017fashion}. All the monochromatic images were converted to RGB format, and then resized to $3 \times 32 \times 32$. As shown in Appendix Table~\ref{tab:dataset-table}, training samples per task in $5$-dataset is relatively higher than that of low data regime typically considered in lifelong learning setting. However, as shown in Figure~\ref{fig:vision} left column, \PLN~show less forgetting than most of the reference algorithms. On the other hand, \zoo\ shows comparatively better performance in relatively high task data size setup. Recall that \PLN\ is based on bagging, and \zoo\ is based on boosting. It is well known that boosting often outperforms bagging when sample sizes are large~\footnote{Authors in \cite{wyner2017explaining} shows that both bagging and boosting asymptotically converge to the Bayes optimal solution. However, for finite sample size and similar model complexity, we empirically find bagging approach to lifelong learning performs better than that of boosting when the training sample size is low whereas boosting performs better on large training sample size (See Figure~\ref{fig:strip}). This is consistent with similar results in single task learning~\cite{Caruana2006-wp, caruana2004ensemble,diaz2018don}}.

\begin{figure}%[h]
    \centering
    \includegraphics[width=0.7\linewidth]{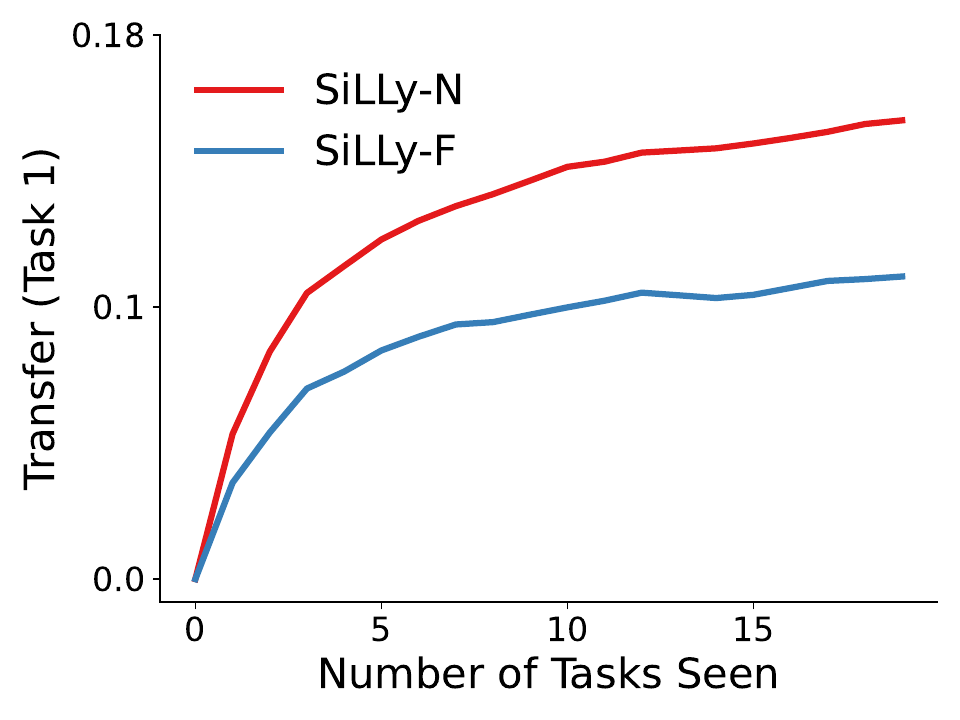}
    \caption{\PLN\ and \PLF\ transfer knowledge effectively when tasks share common classes. Each task is a random selection of 10 out of the 100 CIFAR-100 classes. Both \PLF\ and \PLN\ demonstrate monotonically increasing transfer efficiency for up to 20 tasks. }
    \label{fig:overlapping}
\end{figure}

\paragraph{Overlapping Task Experiment}
\label{app:cifar_repeated}
We considered the setting where each task is defined by a random sampling of 10 out of 100 classes with replacement in CIFAR 10x10. 
This environment is designed to demonstrate the effect of having overlapping tasks, which is a common property of real world lifelong learning tasks. %This setting is related to the previously proposed ``Class-Incremental'' and ``Task-Incremental'' distinction~\cite{Van_de_Ven2019-wy}.
Appendix Figure \ref{fig:overlapping} shows positive transfer from other tasks to Task 1  for \PLF\ and \PLN.

\begin{figure}[!htb]
    \centering
    \includegraphics[width=\linewidth]{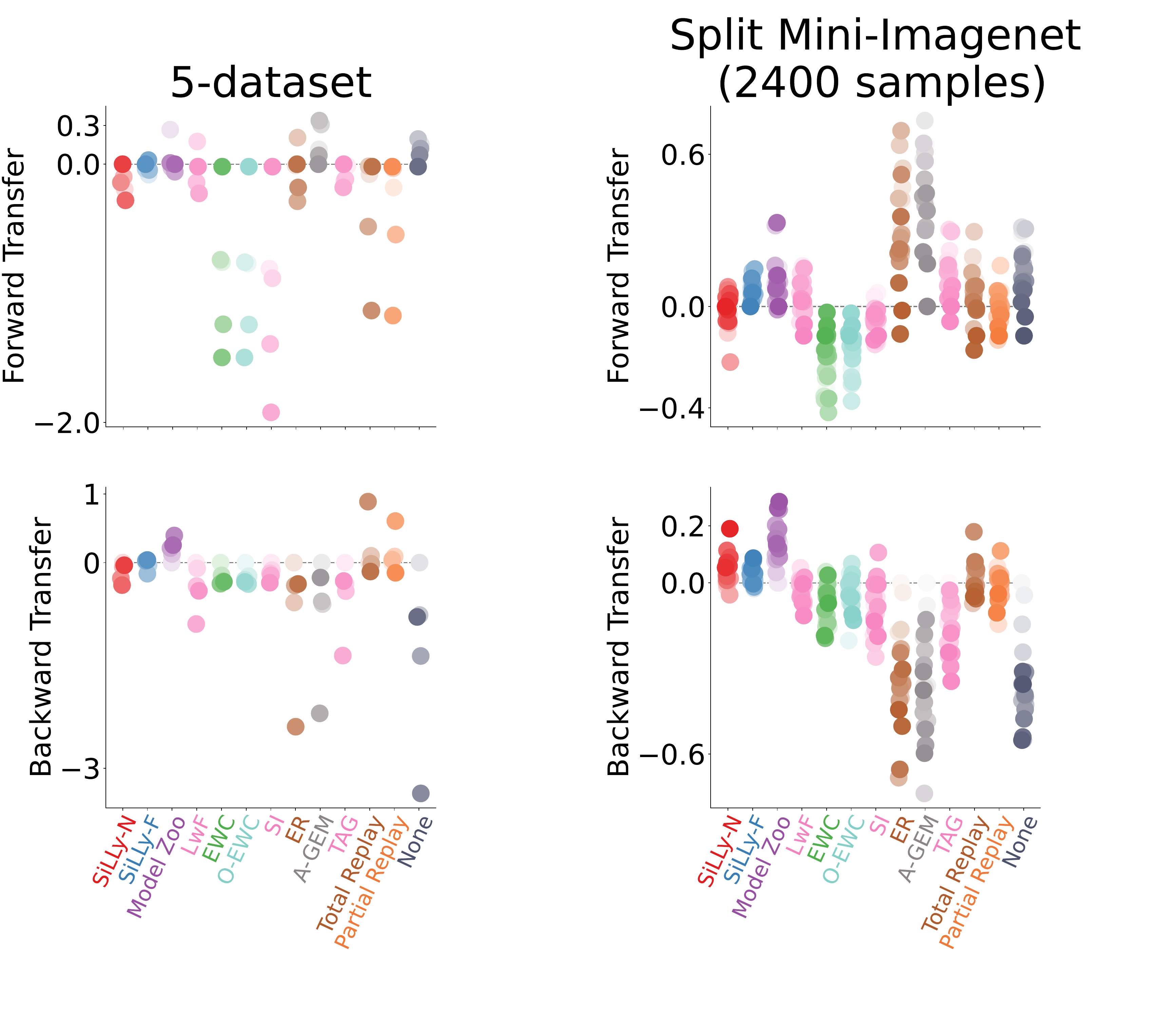}
    \caption{\textbf{Performance of different lifelong learners on two vision datasets.}. 
    }
    \label{fig:vision}
\end{figure}

\begin{figure}%[h]
    \centering
    \includegraphics[width=\linewidth]{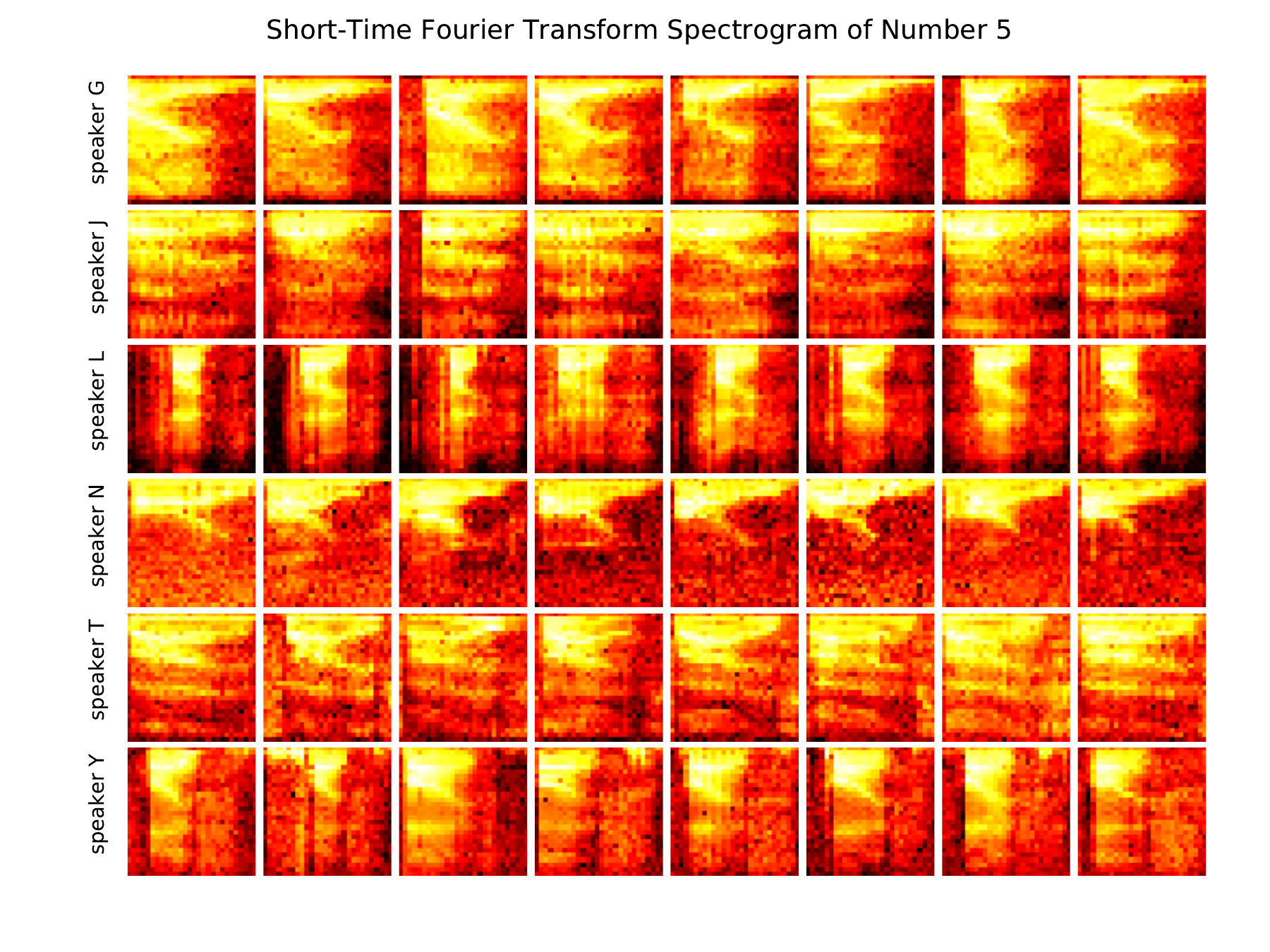}
    \caption{Spectrogram extracted from eight different recordings of six speakers uttering the digit `five'. }
    \label{fig:spectrogram}
\end{figure}

\begin{figure*}[!ht]
    \centering
    \includegraphics[width=.7\linewidth]{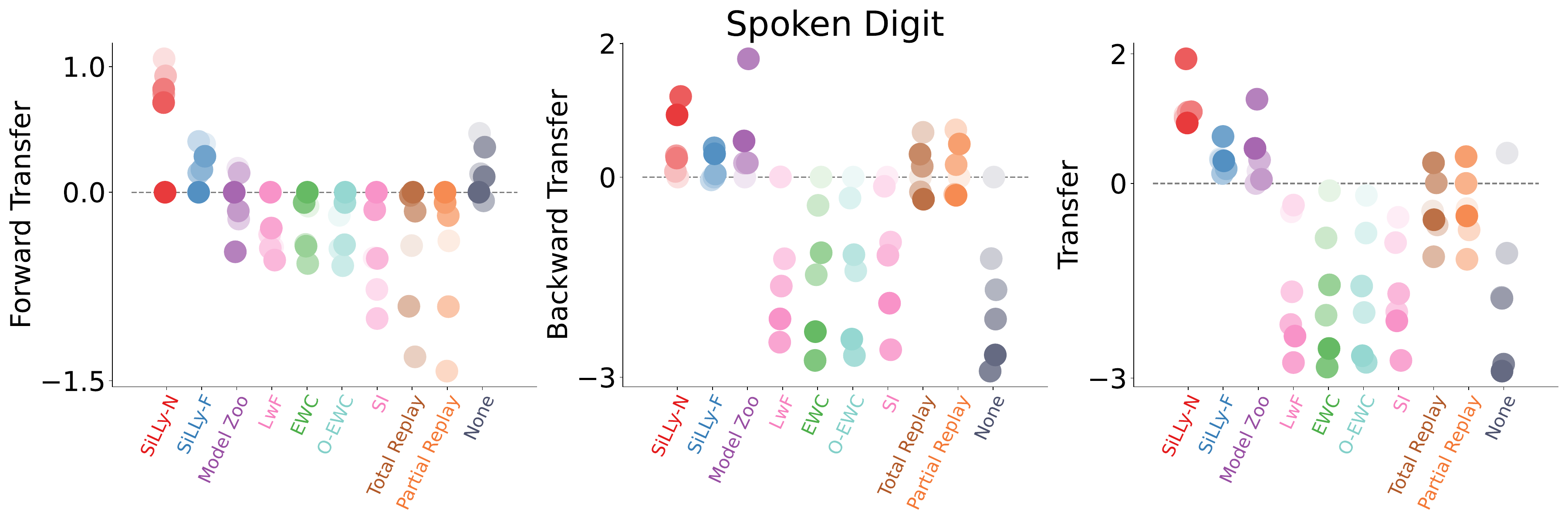}
    \caption{\textbf{Extended results on the Spoken Digit experiments.} This plot contains algorithms not shown in main text Figure \ref{fig:strip}.}
    \label{fig:language}
\end{figure*}

\paragraph{Spoken Digit Experiment Details}
\label{app:spoken}
In this experiment, we used the \textbf{Spoken Digit} dataset provided in \url{https://github.com/Jakobovski/free-spoken-digit-dataset}. 
The dataset contains audio recordings from six different speakers with $50$ recordings for each digit per speaker ($3000$ recordings in total). The experiment was set up with six tasks where each task contains recordings from only one speaker. For each recording, a spectrogram was extracted using Hanning windows of duration $16$ ms with an overlap of $4$ ms between the adjacent windows. The spectrograms were resized down to $28 \times 28$. The extracted spectrograms from eight random recordings of `$5$' for six speakers are shown in Figure \ref{fig:spectrogram}. For each Monte Carlo repetition of the experiment, spectrograms extracted for each task were randomly divided into $55\%$ train and $45\%$ test set. The experiment is summarized in Figure \ref{fig:language}. Note that we could not run the experiment on other $5$ reference algorithms using the code provided by their authors.

\end{document}